\documentclass[onecolumn]{fairmeta}
\usepackage{wrapfig}
\usepackage{tabularx}
\usepackage{textcomp}
\usepackage{stfloats}
\usepackage{url}
\usepackage{verbatim}
\usepackage{graphicx}
\usepackage{titlesec}
\usepackage{tocloft}
\usepackage{adjustbox}
\usepackage{enumitem}
\usepackage{multirow}
\usepackage{pifont}
\usepackage{tikz}
\usepackage{comment}
\usepackage{amsmath,amssymb} % define this before the line numbering.
\usepackage{colortbl}  %彩色表格需要加载的宏包
\usepackage{color}
\usepackage{booktabs} 
\usepackage{hyperref}
\usepackage{graphicx}    % 用于插入图片
\usepackage{subcaption} 
\usepackage{booktabs}
\usepackage{amsmath} % 数学公式支持
\usepackage{multirow} % 表格多行合并
\usepackage{booktabs} % 表格更好的线条
\usepackage{subcaption} % 支持 subtable 和 subfigure
\usepackage{todonotes}
\RequirePackage{xspace}
\makeatletter
\DeclareRobustCommand\onedot{\futurelet\@let@token\@onedot}
\def\@onedot{\ifx\@let@token.\else.\null\fi\xspace}

\makeatother

\definecolor{adptorange}{RGB}{248, 205, 172}
\definecolor{cmpblue}{RGB}{189, 215, 238}
\definecolor{cmpblue}{RGB}{189, 215, 238}

\definecolor{our_red}{RGB}{232,157,160}
\definecolor{our_blue}{RGB}{136,206,230}
\definecolor{our_orange}{RGB}{246,200,168}
\definecolor{our_green}{RGB}{178,211,164}

\definecolor{attn_code0}{RGB}{247,215,200}
\definecolor{attn_code1}{RGB}{238,169,139}
\definecolor{mlp_code0}{RGB}{204,201,221}
\definecolor{mlp_code1}{RGB}{102,95,153}

\definecolor{token_blue}{RGB}{84, 120, 140}
\usepackage{pifont}       % \ding{xx}
\usepackage{bbding}       % \Checkmark,\XSolid,... (需要和pifont宏包共同使用)
\usepackage{fontawesome}
\usepackage{xspace}

\usepackage{float}

\newlength\savewidth

\newcolumntype{x}[1]{>{\centering\arraybackslash}p{#1pt}}
\newcolumntype{y}[1]{>{\raggedright\arraybackslash}p{#1pt}}
\newcolumntype{z}[1]{>{\raggedleft\arraybackslash}p{#1pt}}

\renewcommand{\paragraph}[1]{\vspace{1mm}\noindent\textbf{#1}}
\usepackage{colortbl}
\usepackage{xcolor}
\usepackage{wrapfig}

\renewcommand{\paragraph}[1]{\vspace{1.25mm}\noindent\textbf{#1}}

\usepackage{algorithm}
\usepackage{listings}

\definecolor{codeblue}{rgb}{0.25, 0.5, 0.5}
\definecolor{codekw}{rgb}{0.35, 0.35, 0.75}
\lstdefinestyle{Pytorch}{
    language = Python,
    backgroundcolor = \color{white},
    basicstyle = \fontsize{9pt}{8pt}\selectfont\ttfamily\bfseries,
    columns = fullflexible,
    aboveskip=1pt,
    belowskip=1pt,
    breaklines = true,
    captionpos = b,
    commentstyle = \color{codeblue},
    keywordstyle = \color{codekw},
}

\definecolor{green}{HTML}{009000}
\definecolor{red}{HTML}{ea4335}

\usepackage[utf8]{inputenc} % allow utf-8 input
\usepackage[T1]{fontenc}    % use 8-bit T1 fonts
\usepackage{hyperref}       % hyperlinks
\usepackage{url}            % simple URL typesetting
\usepackage{booktabs}       % professional-quality tables
\usepackage{amsfonts}       % blackboard math symbols
\usepackage{nicefrac}       % compact symbols for 1/2, etc.
\usepackage{microtype}      % microtypography
\usepackage{xcolor}         % colors

\usepackage{amsfonts}
\usepackage{amsmath}
\usepackage{amssymb}
\usepackage{lineno}
\usepackage{multirow}
\usepackage{adjustbox}
\usepackage{bm}

\usepackage{graphicx}
\usepackage{float}
\usepackage{caption}
\usepackage{algpseudocode}
\usepackage{algorithm}
\usepackage{enumitem}
\usepackage{xspace}
\usepackage{wrapfig}
\usepackage{float}

\definecolor{Watermelon_Red}{RGB}{218, 91, 110}
\definecolor{Watermelon_Green}{RGB}{137, 152, 83}
\definecolor{Teal_Blue}{RGB}{0, 128, 128}
\definecolor{mygray}{gray}{0.92}

\newcommand{\method}{\texttt{Lumos-1}\xspace}

\crefname{section}{Sec.}{Secs.}
\Crefname{section}{Section}{Sections}
\Crefname{table}{Table}{Tables}
\crefname{table}{Tab.}{Tabs.}

\title{\method: On Autoregressive Video Generation \\ with Discrete Diffusion from a Unified Model Perspective}

\author[1,2,3]{Hangjie Yuan}
\author[1,2,\dagger]{Weihua Chen}
\author[1,2,3]{Jun Cen}
\author[1]{Hu Yu}
\author[1,2]{Jingyun Liang}
\author[1,2,3]{Shuning Chang}
\author[1,2,3]{Zhihui Lin}
\author[4]{Tao Feng}
\author[3]{Pengwei Liu}
\author[3]{Jiazheng Xing}
\author[1,2]{Hao Luo}
\author[1,2]{Jiasheng Tang}
\author[1]{Fan Wang}
\author[3]{Yi Yang}

\affiliation[1]{DAMO Academy, Alibaba Group}
\affiliation[2]{Hupan Lab}
\affiliation[3]{Zhejiang University}
\affiliation[4]{Tsinghua University\\}

\contribution[\dagger]{Corresponding author\\}

\abstract{

Autoregressive large language models (LLMs) have unified a vast range of language tasks, inspiring preliminary efforts in autoregressive (AR) video generation.
Existing AR video generators either diverge from standard LLM architectures, depend on bulky external text encoders, or incur prohibitive latency due to next-token decoding.
In this paper, we introduce \method, an LLM-based unified model for AR video generation with efficient discrete diffusion.
Firstly, to fit videos with LLMs, we identify that 1D RoPE is ill-suited for visual spatiotemporal correlation modeling, and while demonstrated to be useful, naive 3D RoPE exhibits imbalanced frequency spectra.
Therefore, we propose MM‑RoPE, which preserves the original textual RoPE while seamlessly accommodating video data with comprehensive frequency spectra and scaled 3D positions.
Secondly, to fit the video data's nature and overcome the inefficiency of next-token decoding, we adopt a parallel and mask-based discrete diffusion with the intra-frame bidirectional and inter-frame causal attention masks.
Based on this attention mask, we uncover the frame‑wise loss imbalance issue caused by spatial information redundancy and propose Autoregressive Discrete Diffusion Forcing, which introduces temporal tube masking during training with a compatible inference‑time masking policy to avoid quality degradation.
Despite using only 48 GPUs for pre-training and fine-tuning, limited data and a discrete tokenizer, \method achieves results 
surpassing those of Show-o2 on GenEval, COSMOS-Video2World on VBench-I2V, and OpenSoraPlan on VBench-T2V.
\vspace{-.2cm}
{Code and models: \url{https://github.com/alibaba-damo-academy/Lumos}}
\vspace{-.45cm}

{Email: \texttt{hj.yuan@zju.edu.cn}}
\vspace{-.2cm}

}

\date{\today}

\begin{document}
\maketitle

% \vspace{-.4cm}
\section{Introduction}
% \vspace{-.3cm}

\begin{figure*}[!t]
\centering
\includegraphics[width=1.0\linewidth]{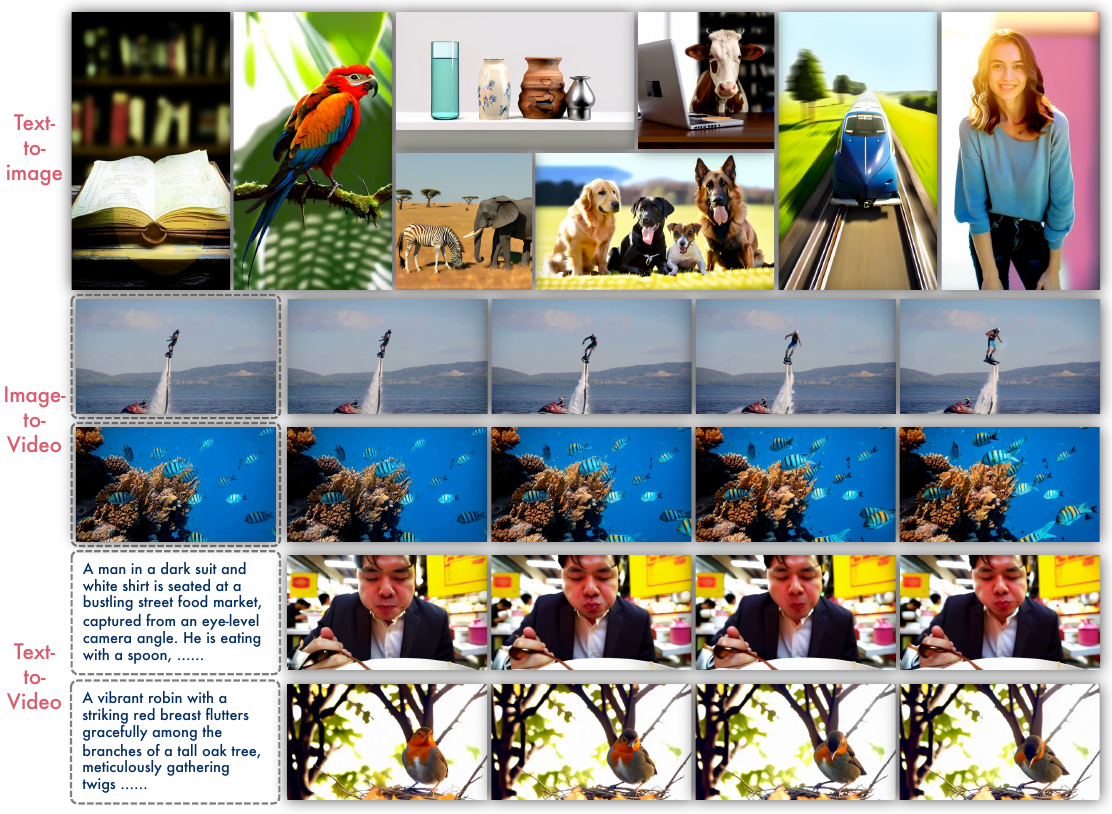}
% \vspace{-.6cm}
\caption{
Visualization of examples generated by \method.
\method supports text-to-image, image-to-video and text-to-video tasks.
}
\vspace{-.33cm}
\label{fig:teaser}
\end{figure*}

Autoregressive (AR) models have demonstrated significant advancement in the field of language processing~\cite{openai2023gpt4,touvron2023llama,ouyang2022InstructGPT,bai2023qwen,liu2024deepseek-v3} by unifying diverse language tasks into one single framework and expanding the scale of large language models to unprecedented sizes. 
Following this trail of research, significant efforts have emerged in autoregressive visual generation~\cite{li2024MAR,Muse,chang2022MaskGiT,luo2024open-magvit2}.
By utilizing similar architectural and generation designs as those used in LLMs, there is substantial potential for advancing LLMs towards a unified model capable of both visual generation and understanding~\cite{wang2024emu3,xie2024show-o,zhou2024transfusion,tong2024metamorph}.
% }

%%%%%%%%%%%%%%%%%%%%%%%%%%%%%%%%%%%%% Para 2 %%%%%%%%%%%%%%%%%%%%%%%%%%%%%%%%%%%%%%%%%
% Pioneering research efforts have been dedicated to instantiating autoregressive visual generation, with a focus on key aspects, spanning improving autoregressive paradigm~\cite{pang2024next-patch-pred,tian2024VAR,chang2022MaskGiT,luo2024open-magvit2}, improving tokenizer format~\cite{li2024MAR} (\textit{e.g.}, from discrete to continuous), and enhancing tokenizer capabilities~\cite{tang2024vidtok,agarwal2025cosmos,wang2024omnitokenizer,shi2024IBQ}.
% However, an autoregressive video generation paradigm that is fully compatible with unified models remains underexplored.
% Preliminary attempts either exhibit architectural distinctions with LLM architectures~\cite{deng2024NOVA,villegas2022phenaki}, rely on external text encoders~\cite{deng2024NOVA}, or trail in generation efficiency~\cite{wang2024loong}, leaving much room for advancement.

% \hangjie{
Pioneering research efforts have been dedicated to instantiating AR visual generation, with a focus on key aspects, spanning improving autoregressive paradigm~\cite{pang2024next-patch-pred,tian2024VAR,chang2022MaskGiT,luo2024open-magvit2}, improving tokenizer format~\cite{li2024MAR} (\textit{e.g.}, from discrete to continuous), and enhancing tokenizer capabilities~\cite{tang2024vidtok,wang2024omnitokenizer,shi2024IBQ}.
However, an AR video generation paradigm that is fully compatible with LLM-based unified models (\textit{e.g.}, Chameleon~\cite{team2024Chameleon} and EMU3~\cite{wang2024emu3}) remains underexplored.
% \hangjie{Emphasize EMU3 and why this is important.}
Preliminary attempts either exhibit architectural distinctions with LLM architectures (\textit{e.g.}, NOVA~\cite{deng2024NOVA} and Phenaki~\cite{villegas2022phenaki}), rely on external text encoders (\textit{e.g.}, LlamaGen~\cite{sun2024LlamaGen} and Fluid~\cite{fan2024fluid}), or trail in generation efficiency due to next-token prediction (\textit{e.g.}, Loong~\cite{wang2024loong}).
% They under-explore a core research area that motivates 
\textit{Therefore, the introduction of LLMs to visual generation motivates careful designs for modeling the spatiotemporal visual space.}
Adapting LLMs for video generation presents two fundamental challenges. 
\textit{First}, the standard positional encoding in LLMs, such as 1D Rotary Position Embeddings (RoPE)~\cite{su2024roformer}, is inherently designed for sequential text and is ill-suited for modeling the complex 3D spatiotemporal correlations of video. 
\textit{Second}, the standard autoregressive paradigm of next-token prediction, while effective for text, is notoriously inefficient for visual data and fails to properly model videos' unique properties: spatial bidirectionality within a frame and strong temporal causality across frames. 
Addressing these two challenges—RoPE representation and prediction paradigm—is critical for building an effective and efficient model for video generation.

\begin{wraptable}{r}{0.6\textwidth} 
% \begin{table}[h!]
\centering
\vspace{-.3cm}
\caption{
\small
% \textcolor{Watermelon_Red}{
\textbf{Design comparison with other types of RoPE}.
% }
}
% \vspace{-.2cm}
\resizebox{1.\linewidth}{!}{
\begin{tabular}{@{}lcccc@{}}
\toprule
\textbf{\shortstack{RoPE \\ Type}} & \textbf{\shortstack{Compatiable with \\ Text RoPE}} & \textbf{\shortstack{3D \\ Structure}} & \textbf{\shortstack{Comprehensive \\ Frequency Allocation}} & \textbf{\shortstack{Strategic \\ Scaling}} \\
\midrule
M-RoPE    & \ding{52}  &  \ding{52}  &  \ding{55}  &  \ding{55} \\
U-RoPE    & \ding{52}  &  \ding{52}  &  \ding{55}  &  \ding{55} \\
IL-RoPE   & \ding{55}  &  \ding{52}  &  \ding{55}  &  \ding{55} \\
VideoRoPE & \ding{52}  &  \ding{52}  &  \ding{55}  &  \ding{52} \\
HoPE      & \ding{52}  &  \ding{52}  &  \ding{55}  &  \ding{52} \\
\midrule
MM-RoPE   & \ding{52}  &  \ding{52}  &  \ding{52}  &  \ding{52} \\
\bottomrule
\end{tabular}}
\vspace{-.3cm}
\label{tab:diff_RoPE_design_main_paper}
% \end{table}
\end{wraptable}

% \hangjie{
To account for the spatiotemporal nature of videos using the RoPE technique, we start with systematic experiments and analysis (in~\cref{sec:MM-RoPE}) of extending 1D RoPE to a vanilla 3D RoPE~\cite{hong2022cogvideo,kong2024hunyuanvideo}, revealing that 1D RoPE is far from optimal, and the incorporation of 3D RoPE facilitates autoregressive generative learning.
However, existing 3D RoPE still suffers from imbalanced frequency spectrum ranges for temporal and spatial modeling, as shown in~\cref{sec:MM-RoPE}.
Building on this insight, we propose MM-RoPE, a new family of RoPE that better accommodates the structure of video data.
As summarized in~\cref{tab:diff_RoPE_design_main_paper}, the core enhancement lies in more distributed frequency allocations for more comprehensive context modeling, substantially improving the visual generation capability.
Moreover, we design a principled scaling to the 3D positions, which improves vision-language modality balancing.
Though MM-RoPE modifies RoPE for visual tokens, the RoPE for text tokens remains consistent with native LLMs, therefore preserving the language learning capabilities.
To account for the nature of videos (\textit{i.e.}, spatial bidirectionality and temporal causality), we build our model upon the intra-frame bidirectional and inter-frame causal token dependency strategy, so that we can introduce the use of the efficient and parallel discrete diffusion~\cite{chang2022MaskGiT}.
% To account for the temporal causality of videos, we build our model upon the intra-frame bidirectional and inter-frame causal token dependency strategy, so that we can introduce the use of the efficient and parallel discrete diffusion~\cite{chang2022MaskGiT}.
Discrete diffusion requires training with masks~\cite{li2024MAR}.
However, the naive global random mask prediction leads to the loss imbalance issue, which means that masked visual tokens in the later frames tend to have much lower loss, impeding effective video modeling.
% the naive global random masks  or more advanced diffusion forcing~\cite{chen2024diffusion} 
We attribute this to videos' spatial information redundancy, as spatial information leakage allows the model to easily predict the masked token by attending to unmasked ones in previous frames.
To address this, we propose Autoregressive Discrete Diffusion Forcing (AR-DF).
The core of AR-DF training involves temporal tube masking~\cite{tong2022videomae}, which repeats a frame-level random mask pattern across the temporal axis.
% Such a design blocks the information flow from masked tokens in later frames from referring to previous ones.
Such a design prevents masked tokens in later frames from learning a shortcut--copying previous ones.
The core of AR-DF inference is the strategic inference-time masks that align with partial observation of history during training, enabling inference without degraded frame quality and motion.
% }

%%%%%%%%%%%%%%%%%%%%%%%%%%%%%%%%%%%%% Para 6 %%%%%%%%%%%%%%%%%%%%%%%%%%%%%%%%%%%%%%%%%

% We leverage Llama~\cite{touvron2023llama2} as our architecture and leverage a unified discrete codebook for language and visual data.
% Through a stage-wise training with GPU memory-friendly techniques, we pre-train \method \textit{from scratch} on 60 million images and 10 million videos using only 48 GPUs, achieving performance comparable to EMU3 on GenEval, COSMOS-Video2World on VBench-I2V, and OpenSoraPlan on VBench-T2V.

% \hangjie{
Collectively, we propose \method, a model building on Llama~\cite{touvron2023llama2} and discrete diffusion to achieve AR video generation.
Through a stage-wise training with GPU memory-friendly techniques, we pre-train \method \textit{from scratch} on 60 million images and 10 million videos using only 48 GPUs, achieving performance comparable to EMU3 on GenEval, COSMOS-Video2World on VBench-I2V, and OpenSoraPlan on VBench-T2V (\cref{fig:teaser} visualizes examples generated by \method).
% }

%%%%%%%%%%%%%%%%%%%%%%%%%%%%%%%%%%%%% Para 7: summarize the contributions %%%%%%%%%%%%%%%%%%%%%%%%%%%%%%%%%%%%%%%%%
% % \hangjie{
% Our contributions can be summarized as:
% \textbf{1)} We identify that while being useful, naive 3D RoPE suffers imbalanced frequency spectra in AR video generation.
% % We identify the issue of imbalanced frequency spectrum ranges when introducing 3D RoPE to AR video generation.
% % We introduce the use of 3D RoPE to native LLMs for visual generation and identify the issue of imbalanced frequency spectrum ranges.
% Therefore, we propose MM-RoPE with more distributed frequency allocations and principled scaling, to achieve better dependency modeling and enforce modality balancing.
% \textbf{2)} We identify the loss imbalance issue during mask prediction training, and propose AR-DF to enforce training with temporal modeling and introduce inference-time masks to avoid quality degradation. 
% \textbf{3)} Through extensive experiments, we comprehensively validate the efficacy of the above designs.
% % using generation metrics and validation loss.
% \method achieves competitive results on T2I, I2V and T2V tasks, while using limited training computing, data and a discrete tokenizer.
% % }

% \hangjie{
Our contributions can be summarized as:
\textbf{1)} We propose \method, a pure LLM-based unified architecture with discrete diffusion for AR video generation.
\textbf{2)} We identify that although being useful, naive 3D RoPE exhibits imbalanced frequency spectra.
Therefore, we propose MM-RoPE with more distributed frequency allocations and principled scaling, to achieve better dependency modeling and modality balancing.
\textbf{3)} We identify the loss imbalance issue during mask prediction training, and propose AR-DF to enforce training with temporal modeling and introduce inference-time masks to avoid quality degradation. 
\textbf{4)} We validate the efficacy of the above designs through extensive experiments.
\method achieves competitive results on T2I, I2V and T2V tasks, while using limited training computing, data and a discrete tokenizer.
\section{Related Work}
\noindent\textbf{Autoregressive video generation.}
Autoregressive video synthesis follows a trajectory similar to images (from a dedicated architecture to a unified one like LLMs), but the substantial computational overhead encouraged researchers to explore different granularities of autoregression.
At the \textit{macro} level, methods such as CausVid~\cite{yin2024CausVid}, Pyramidal Flow~\cite{jin2024pyramidal_flow}, and MAGI-1~\cite{MAGI-1} generate video clips by recursively invoking diffusion models.
A second line of work adopts a \textit{hybrid AR-diffusion} strategy that couples a pre-trained diffusion backbone with an external autoregressive planner, as exemplified by ARLON~\cite{li2024arlon} and Mardini~\cite{liu2024mardini}.
Our focus, however, is on \textit{micro} AR approaches~\cite{gu2025FAR-video,zhou2025MAGI,yu2023MAGVIT,yu2023MAGVIT-v2,villegas2022phenaki,kondratyuk2023VideoPoet,wang2024emu3,deng2024NOVA,wang2024loong}, which treat the entire spatiotemporal token sequence as a single context and generate videos end-to-end with one autoregressive transformer (\textit{e.g.}, Phenaki~\cite{villegas2022phenaki}, VideoPoet~\cite{kondratyuk2023VideoPoet}, Loong~\cite{wang2024loong}, EMU3~\cite{wang2024emu3}, NOVA~\cite{deng2024NOVA}). 
Building on this paradigm, we systematically study the use of RoPE, and further introduce AR-DF, which ensures efficiency and is compatible with intra-frame bidirectionality and inter-frame temporal causality, enabling effective training and inference with LLM architectures.

% \noindent\textbf{RoPE for vision-language data.}
% Since RoPE's successful introduction to LLMs~\cite{su2024roformer,touvron2023llama}, DiT~\cite{peebles2023DiT}, a widely adapted category of models, incorporated RoPE as a technique to inject priors into the generation process.
% However, this technique is heavily underexplored in autoregressive video generation.
% Related works, including M-RoPE~\cite{wang2024qwen2-vl}, RoPE-2D~\cite{agrawal2024pixtral-12B}, TAD-RoPE~\cite{gao2024Tad-RoPE} and VideoRoPE~\cite{wei2025VideoRoPE}, incorporated structure priors (\textit{i.e.}, spatiotemporal or its subsets) into RoPE for better vision-language understanding.
% VoPE~\cite{chen2024JEPA-T2I_VoPE} improves high-resolution image generation by continuous resolution learning.
% We propose a distributed and scaled RoPE variant, MM-RoPE, to enhance the efficacy of autoregressive video generation, while ensuring compatibility with unified models.

%%%%%%%%%%%%%%%%%%%%%%%% RoPE for vision-language data %%%%%%%%%%%%%%%%%%%%%%%%
%%% The following paragraph has been revised.
\noindent\textbf{RoPE for vision-language data.} 
RoPE was first proven effective for LLMs~\cite{su2024roformer,touvron2023llama,huang2025revisiting_RoPE} and later adopted in DiTs~\cite{peebles2023DiT} to inject spatial priors. 
However, its potential for AR video generation is heavily underexplored. 
% \textcolor{Watermelon_Red}{
Related works, including M-RoPE~\cite{wang2024qwen2-vl}, U-RoPE~\cite{tang2025U-RoPE}, RoPE-2D~\cite{agrawal2024pixtral-12B}, VoPE~\cite{chen2024JEPA-T2I_VoPE}, TAD-RoPE~\cite{gao2024Tad-RoPE}, IL-RoPE~\cite{liao2025Mogao}, VideoRoPE~\cite{wei2025VideoRoPE} and HoPE~\cite{li2025HoPE}, incorporated structure priors into RoPE (\textit{i.e.}, spatiotemporal information or its subsets) or improve the frequency allocation of RoPE for better vision-language understanding.
% }
% \textcolor{Watermelon_Red}{
However, they bear drawbacks: 
\textbf{1)} some of them are not compatible with the original text RoPE;
\textbf{2)} their frequency allocations remain suboptimal for local-global dependency modeling;
\textbf{3)} their scaling strategy remains suboptimal for spatiotemporal data.
% }
% VoPE~\cite{chen2024JEPA-T2I_VoPE} improves high-resolution image generation by continuous resolution learning.
We propose MM‑RoPE, a distributed and scaled RoPE technique that improves AR video synthesis while remaining plug‑and‑play for unified models.
% unified multimodal models.

\section{\textbf{\method}}\label{sec:method}
% \vspace{-.3cm}

%%%%%%%%%%%%%%%%%%%%%%%%%%% Comment out during submission %%%%%%%%%%%%%%%%%%%%%%%%%%%%

% % In this section, we commence with preliminaries.
% This section aims to introduce the design philosophy behind \method.
% % , which utilizes the autoregressive LLM architecture for video generation.
% In~\cref{sec:MM-RoPE}, we introduce MM-RoPE that enables better spatiotemporal awareness of an LLM for modeling visual data.
% In~\cref{sec:tube_masking}, we introduce AR-DF that enables effective training and inference.
% In~\cref{sec:implementation}, we introduce important techniques for implementing \method, including architecture, memory-friendly techniques, \textit{etc}.
% % architecture, tokenizers, memory-friendly techniques, stage-wise training, \textit{etc}.

\subsection{Spatiotemporal Correlation Injection via MM-RoPE}
\label{sec:MM-RoPE}

\begin{figure}[!t]
\centering
\includegraphics[width=1.0\linewidth]{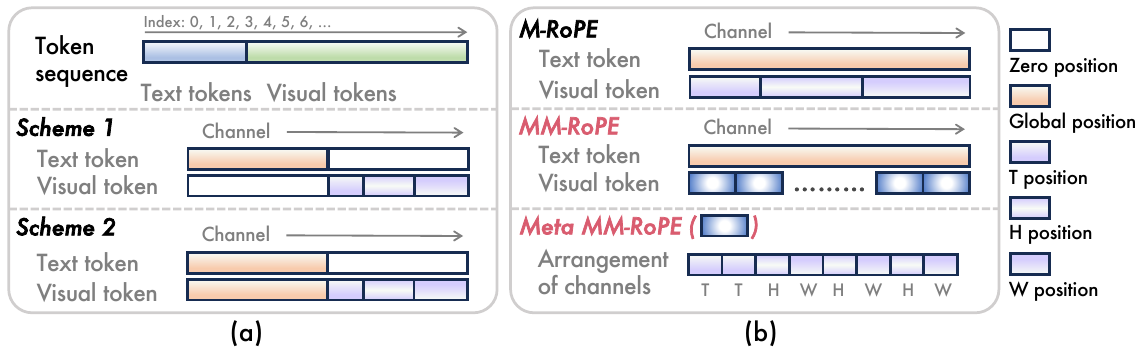}
\vspace{-.3cm}
\caption{
\small
(a) \textbf{Initial exploration of 3D RoPE in autoregressive video generation}.
The sequence is comprised of text tokens and visual tokens (default arrangement in this paper), with global indices starting from 0.
Two schemes are exemplified using one text token and one visual token.
% In these two schemes, temporal, height and width positions start from 0.
In Scheme 1 and 2, temporal, height and width positions start from 0.
(b) \textbf{Details of MM-RoPE} compared to M-RoPE.
The figure illustrates the distributed channel allocation of MM-RoPE.
Temporal, height and width positions are indexed after text tokens.
}
\vspace{-.33cm}
\label{fig:mmrope_details}
\end{figure}

\noindent\textbf{Preliminaries of 3D RoPE.}
RoPE~\cite{su2024roformer} aims to encode the absolute position with a rotation matrix while incorporating the explicit relative position dependency in the attention mechanism.
If we denote $f_q(\bm{x}_m, m)$ and $f_k(\bm{x}_n, n)$ as query and key features encoding positions $m$ and $n$, the attention calculation in RoPE can be rewritten as:
\begin{equation}
    f_q(\bm{x}_m, m)^{\rm T} f_k(\bm{x}_n, n) = \bm{x}_m^{\rm T} \bm{W}_q^{\rm T} \bm{R}^{d}_{\Theta, \tau} \bm{W}_k \bm{x}_n, \quad \tau = n - m
\end{equation}
where $\bm{W}_{q,k}$ is the projection matrix; $\bm{R}^{d}_{\Theta, \tau}$ is the rotary matrix with pre-defined parameters $\Theta = \{\theta_i=\beta^{-2(i-1)/d}, i=[1,2,...,d/2]\}$ ($d$ is the feature dimension and $\beta$ is the base frequency).
We formulate $\bm{R}^{d}_{\Theta, \tau}$ using a base rotary matrix $R_{\theta, \tau}$, 
with $\theta$ as the frequency and $\tau$ as the relative position:
\begin{equation}
    {\scriptsize
    \bm{R}^{d}_{\Theta, \tau} = 
    \begin{bmatrix}
        R_{\theta_{1}, \tau} & 0 & \cdots & 0 \\
        0 & R_{\theta_{2}, \tau} & \cdots & 0 \\
        \vdots &\vdots & \ddots & 0 \\
        0 & 0 & \cdots & R_{\theta_{d/2}, \tau} \\
    \end{bmatrix}, \quad 
    R_{\theta, \tau} = 
    \begin{bmatrix}
        \cos \tau \theta & -\sin \tau \theta \\
        \sin \tau \theta & \cos \tau \theta
    \end{bmatrix}
    }
\end{equation}
However, applying the original RoPE to modeling visual data remains suboptimal considering the spatiotemporal nature of visual tokens.
Diffusion models~\cite{ho2022video_diffusion_models} improved upon this by proposing 3D RoPE that jointly injects spatiotemporal latent coordinates during attention calculation~\cite{hong2022cogvideo,kong2024hunyuanvideo}.
If we slightly abuse the annotation by denoting $\bm{x}_m^{\rm T} \bm{W}_q^{\rm T}$ and $\bm{W}_k \bm{x}_n$ as $\bm{X}_m^{\rm T}$ and $\bm{X}_n$, we can write the attention calculation based on 3D RoPE as:
\begin{equation}
    \resizebox{\textwidth}{!}{$
    \begin{aligned}
    % f_q(\bm{x}_m, m)^{\rm T} f_k(\bm{x}_n, n) =
    \bm{X}_{m, t_s:t_e}^{\rm T}
    \begin{bmatrix}
        R_{\theta_{t_s + 1}, \tau_t} & \cdots & 0 \\
        \vdots                   & \ddots & \vdots \\
        0                        & \cdots & R_{\theta_{t_e}, \tau_t} \\
    \end{bmatrix} 
    \bm{X}_{n, t_s:t_e} 
    +
    \bm{X}_{m, h_s:h_e}^{\rm T}
    \begin{bmatrix}
        R_{\theta_{h_s + 1}, \tau_h} & \cdots & 0 \\
        \vdots                   & \ddots & \vdots \\
        0                        & \cdots & R_{\theta_{h_e}, \tau_h} \\
    \end{bmatrix} 
    \bm{X}_{n, h_s:h_e}
    % &+ \\
    +
    \bm{X}_{m, w_s:w_e}^{\rm T}
    \begin{bmatrix}
        R_{\theta_{w_s + 1}, \tau_w} & \cdots & 0 \\
        \vdots                   & \ddots & \vdots \\
        0                        & \cdots & R_{\theta_{w_e}, \tau_w} \\
    \end{bmatrix}
    \bm{X}_{n, w_s:w_e}
    \end{aligned}$}
\end{equation}
where $\{t_s,t_e\} = \{0,\frac{2}{16}d\}$, $\{h_s,h_e\} = \{\frac{2}{16}d, \frac{5}{16}d\}$ and $\{w_s,w_e\} = \{\frac{5}{16}d, \frac{1}{2} d\}$ denote the start and end dimension index for encoding temporal, height and width relative position; 
$\bm{X}_{m, t_s:t_e}^{\rm T}$ denotes the submatrix extracted from $\bm{X}_{m}^{\rm T}$ using row indices $[2t_s, 2t_e)$;
other matrices are similarly defined.

%%% 把全文所有公式的index顺序换为0开始，从公式2下面就要开始了
%%% $\bm{X}_{m, t_s:t_e}^{\rm T}$用右下角的index取submatrix，要强调前闭后开，然后我们应该x2，数值上才是对的。

% \begin{figure}[!t]
% \centering
% \includegraphics[width=1.0\linewidth]{LlamaVGen/figures/rope_analysis.pdf}
% \vspace{-.66cm}
% \caption{
% \small
% (a) \textbf{Frequency allocation in the previous 3D RoPE};
% (b) \textbf{Comparison of rotary speed} of the first dimension for temporal dimensions (high frequencies) and height dimensions (meddle frequencies) in the previous 3D RoPE.
% }
% \label{fig:rope_analysis}
% \end{figure}

\begin{figure}[!t]
\centering
\includegraphics[width=1.0\linewidth]{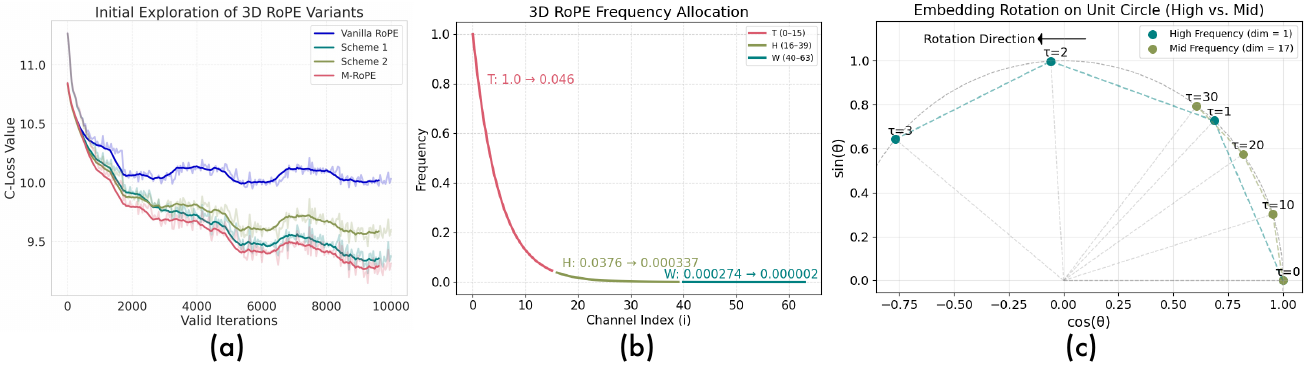}
\vspace{-.3cm}
\caption{
\small
(a) \textbf{Validation loss curve of a toy experiment exploring the necessity of 3D RoPE} using the 0.5B model;
(b) \textbf{Frequency allocation in the vanilla 3D RoPE};
% (c) \textbf{Comparison of rotary speed} of the first dimension for temporal dimensions (high frequencies) and height dimensions (middle frequencies) in the vanilla 3D RoPE.
(c) \textbf{Comparison of rotary speed} of the first dimension for temporal and height dimensions (high and middle frequencies) in the vanilla 3D RoPE.
}
\vspace{-.42cm}
\label{fig:mmrope_curves}
\end{figure}

% \begin{wrapfigure}{r}{0.5\textwidth}
% % \begin{figure}[!t]
% \centering
% \includegraphics[width=1.\linewidth]{LlamaVGen/figures/mmrope_details.pdf}
% \vspace{-.65cm}
% \caption{
% \small
% \textbf{Details of MM-RoPE}, exemplified using one text token and one visual token.
% The figure illustrates the distributed channel allocation of MM-RoPE.
% The temporal, height and width positions are indexed after text tokens.
% }
% \label{fig:mmrope_details}
% % \end{figure}
% \end{wrapfigure}

\noindent\textbf{Initial exploration of incorporating 3D RoPE.}
% \hangjie{mmrope curves a should be referred}
% \hangjie{mmrope details a should be referred}
Our preliminary exploration starts with the introduction of 3D RoPE to AR video generation.
We follow previous works~\cite{fan2024fluid,li2024MAR} to \textit{utilize the validation loss to observe the effectiveness} due to its strong correlation with evaluation metrics.
By default, we use the cross-entropy loss (C-Loss), following standard LLM training.
% It's worthing noting that the applied RoPE should keep the capability of modeling language context unchanged, therefore, 
We compare the vanilla LLM RoPE with three schemes, as shown in~\cref{fig:mmrope_details}:
\textbf{1)} \textbf{\textit{Scheme 1}}, which allocates the first $1/2$ channels to encode the global position (\textit{i.e.}, the index in the global sequence) and the second $1/2$ channels to encode the temporal, height and width positions with a ratio of $2:3:3$.
For text tokens, we only use the first half channels to encode the global positions to ensure the language modeling capability, while for visual tokens, we only use the second half to encode 3D positions;
\textbf{2)} \textbf{\textit{Scheme 2}}, which extends upon scheme 1 by leveraging the first half channels of the visual tokens to encode the global positions.
\textbf{3)} \textbf{\textit{M-RoPE}}~\cite{wang2024qwen2-vl}, which uses all channels of the visual tokens to encode the 3D positions.
From~\cref{fig:mmrope_curves}(a), we can observe that:
\textbf{1)} The incorporation of the spatiotemporal correlation in RoPE can significantly improve the performance of the model fitting data by comparing vanilla RoPE and Scheme 1.
\textbf{2)} The injection of the raster-scan order position information to visual tokens (\textit{i.e.}, the global position in Scheme 2 of \cref{fig:mmrope_details}(a)) can degrade the performance.
\textbf{3)} A comprehensive channel utilization (M-RoPE) is better than partial channel utilization (Scheme 1).
% Therefore, it is promising to inject such priors in this generative model.
Therefore, it is promising to inject such priors in \method.
% \hangjie{Should indicate that we use 0.5B model, whose architectural details are placed in Sec XXX. We can do this in the caption title.}

% \hangjie{Add experiments for w/ and w/o RoPE.}
% We observe a clear improvement with the vanilla version of 3D RoPE, demonstrating its necessity. 
% Based on this, we further propose a question:
% which way might be the most effective 

%%% The following paragraph has been revised.
% \noindent\textbf{Understanding 3D RoPE's mechanism and core limitation.}
\noindent\textbf{Peering into 3D RoPE and identifying its limitations.}
Although 3D RoPE proves effective in practice, its design remains suboptimal.
% In XXX, we visualize the frequencies allocated to modeling temporal, height and width, where \textit{temporal channels dominate large frequency spectrum}, while height and width channels are relegated to near-zero frequencies.
In~\cref{fig:mmrope_curves}(b), we visualize how the frequencies are allocated to model temporal, height, and width dimensions. 
We observe that the \textit{temporal channels dominate large frequency spectrum}, whereas the height and width channels are relegated to near-zero frequencies.
For the sine function, the relative positions $\tau$ (when $\tau >= 0$) should not exceed one period to avoid ambiguity since radians beyond $2\pi$ start repeating patterns in the function.
Beyond this range, the model cannot distinguish fine-grained positional differences since rotary embeddings of this channel lose uniqueness.
% Suppose the head dimension in the attention mechanism is 64 and the base frequency is $\beta = 5\times10^5$, the channel index allocated to the temporal dimension is 1 to 16, whose frequencies range from $0.8146$ to $0.0376$ and the quarter of the period ranges from $1.93$ to $41.78$.
For the low-indexed channels, the embeddings can rotate significantly faster than high-indexed channels (as shown in~\cref{fig:mmrope_curves}(c)), leading to accelerated aliasing and loss of embedding uniqueness.
For the high-indexed channels, the embeddings rotate so slowly that they lack sufficient resolution to model subtle local changes.
% minute local changes.
% Moreover, the height and width channels are symmetrically important, while they occupy significantly different frequency spectrums.
Finally, while height and width are symmetrically important, they occupy disproportionately small and different segments of the overall frequency spectrum, reducing their capacity to capture spatial details effectively.

\begin{wrapfigure}{r}{0.5\textwidth}
% \begin{figure}[!t]
\centering
\vspace{-.5cm}
\includegraphics[width=\linewidth]{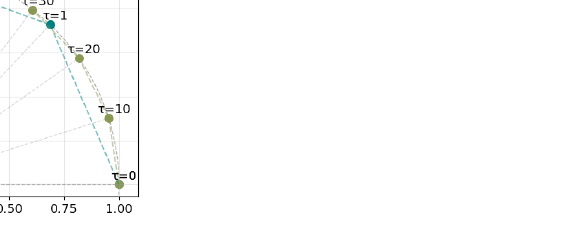}
\vspace{-.3cm}
\caption{
\small
(a) \textbf{Details of scaled 3D positions in MM-RoPE.}
% (b) \textbf{The temporal causal mask used in \method.}
(b) \textbf{Temporal causal mask used in \method.}
}
\vspace{-.5cm}
\label{fig:mmrope_scaling}
% \end{figure}
\end{wrapfigure}

%%% The following paragraph has been revised.
\noindent\textbf{MM-RoPE: a distributed and scaled 3D RoPE mechanism.}
To elegantly solve the aforementioned limitation, we propose MM-RoPE, a distributed 3D RoPE mechanism.
Compared with M-RoPE~\cite{wang2024qwen2-vl} that is widely adopted in vision-language models, one core idea of MM-RoPE is to encode relative positions across comprehensive frequency spectra for all 3D information.
% 还要有实验提到M-RoPE。
As illustrated in~\cref{fig:mmrope_details}(b), the RoPE for text tokens in MM-RoPE follows the standard design of an LLM, whereas the RoPE for visual tokens is comprised of multiple meta MM-RoPE components.
Within each meta MM-RoPE, we keep the ratio of 3D information identical to 3D RoPE (\textit{i.e.}, $2:3:3$), while minimizing the overall dimensionality to maintain a more distributed design.
Concretely, we first allocate channels for temporal modeling, then symmetrically interleave height and width channels to model spatial information.
We can formulate the attention calculation of the first meta MM-RoPE as:
% mathematically
% Let us take the first meta MM-RoPE component as an example. Its attention calculation is given by:
\begin{equation}
    {\small
    \begin{aligned}
    % f_q(\bm{x}_m, m)^{\rm T} f_k(\bm{x}_n, n) =
    \bm{X}_{m, 0:2}^{\rm T}
    \begin{bmatrix}
        R_{\theta_1, \tau_t} & 0 \\
        0 & R_{\theta_2, \tau_t} \\
    \end{bmatrix} 
    \bm{X}_{n, 0:2} 
    +
    \bm{X}_{m, 2:4}^{\rm T}
    \begin{bmatrix}
        R_{\theta_3, \tau_h} & 0 \\
        0 & R_{\theta_4, \tau_w} \\
    \end{bmatrix} 
    \bm{X}_{n, 2:4}
    & + \\
    \bm{X}_{m, 4:6}^{\rm T}
    \begin{bmatrix}
        R_{\theta_5, \tau_h} & 0 \\
        0 & R_{\theta_6, \tau_w} \\
    \end{bmatrix} 
    \bm{X}_{n, 4:6} 
    +
    \bm{X}_{m, 6:8}^{\rm T}
    \begin{bmatrix}
        R_{\theta_7, \tau_h} & 0 \\
        0 & R_{\theta_8, \tau_w} \\
    \end{bmatrix} 
    \bm{X}_{n, 6:8}
    \end{aligned}
    }
\end{equation}
% where a meta MM-RoPE component consists of $16$ channels, repeatedly to compose the RoPE strategy for visual tokens.
where each meta MM-RoPE component comprises $16$ channels; additional components are defined analogously and collectively form the RoPE strategy for visual tokens.

%%%%%%%%%%%%%%%%%%%%%%%%%% V2 %%%%%%%%%%%%%%%%%%%%%%%%%%
\begin{figure}[t]
\centering
\begin{minipage}[t]{0.39\linewidth}
\begin{algorithm}[H]
\scriptsize
\caption{\small AR-DF Training Procedure}
\label{alg:AR-DF_training}
\textbf{Require:} video dataset $\mathcal{D}$, training mask ratio $\rho_{tra}$, generative model $G_\phi$, number of latent frames $T$
\begin{algorithmic}[1]
\For{\textbf{each} video $\bm{X} \in \mathcal{D}$}
  \State \(\bm{X}_p,\bm{X}_v \leftarrow \text{tokenize text, frames}\)
  \State \(\textbf{Sample a mask pattern for all frames}\):
  \State \quad \(\rho \sim \mathrm{Uniform}([\rho_{tra}, 1])\); 
  \State \quad \(\bm{M} \sim \mathrm{Bernoulli}(1-\rho)\) 
  % \Comment{Mask pattern for all frames}
  \For{\(t = 1 \to T\)} \Comment{Apply $\bm{M}$ to all frames}
    \State \(\widetilde{\bm{X}}_v^{(t)} 
      = \bm{M}\odot \bm{X}_v^{(t)}
        + (1-\bm{M}) \odot [\text{MASK}]\)
  \EndFor
  \State Form \(\widetilde{\bm{X}} = \{\bm{X}_p,\, \widetilde{\bm{X}}_v^{(1)},\dots,\widetilde{\bm{X}}_v^{(T)}\}\)
  \State Generate attention mask $AttnMask$ for \(\widetilde{\bm{X}}\)
  \State \(\widehat{\bm{X}} \leftarrow G_\phi(\widetilde{\bm{X}}, AttnMask)\) 
  \State Compute loss \(\mathcal{L}(\widehat{\bm{X}}, \bm{X},\bm{M})\)
  \State Backprop \& update \(\phi\)
\EndFor
\end{algorithmic}
\end{algorithm}
\vspace{-0.6cm}
% \caption{
% \textbf{Training algorithm with AR-DF}. 
% For simplicity, this only includes training on videos.
% }
\end{minipage}\hfill% % \hfill%里面的%非常重要，不能删掉。
% \end{figure}
% \begin{figure}[t]
% \centering
\begin{minipage}[t]{0.59\linewidth}
\begin{algorithm}[H]
% \small
\scriptsize
\caption{\small AR-DF Inference Procedure}
\label{alg:AR-DF_inference}
\textbf{Require:} text prompt $\bm{X}_p$, trained model $G_\phi$, inference mask ratio $\rho_{inf}$, number of latent frames $T$, number of generation steps $N_{steps}$, number of tokens in a latent frame $N_f$, KV cache $\mathcal{C} = \varnothing$, generated latent list $\mathcal{V} = \varnothing$
\begin{algorithmic}[1]
\State \(\textbf{Initialize text cache}\):
  \State \quad Generate text causal mask \(AttnMask^{(p)}\) for \(\bm{X}_p\)
  % \State \quad \(\mathcal{C} \leftarrow G_\phi(\widetilde{\bm{X}}, AttnMask^{(p)})\) \Comment{Store cache for the prompt}
  \State \quad \(\mathcal{C} \leftarrow G_\phi(\bm{X}_p, AttnMask^{(p)})\) \Comment{Store cache for the prompt}
\State \(\textbf{Sample cache mask}\):
  \State \quad \(\bm{M}_{inf} \sim \mathrm{Bernoulli}(1 - \rho_{inf})\)
\For{\(t = 1 \to T\)}  \Comment{Reused for all frames}
  \State Initialize all tokens in \(\bm{X}_v^{(t)}\) as \([\text{MASK}]\)
  \State Generate temporal causal mask \(AttnMask^{(t)}\) for \(\{\bm{X}_p, \bm{X}_v^{(t)}\}\)

  % \For{\(n = 1 \to N_{steps}\)}
  %   \State {%\color{blue} 
  %     \(\alpha \leftarrow \cos(\frac{\pi}{2}\frac{n}{N_{steps}})\)
  %   }
  %   \Comment{Cosine schedule factor in [0,1]}
  %   \State $U \leftarrow \text{floor}(\alpha \times N_f)$ \Comment{Number of tokens to unmask in this iteration}
  %   \State $\bm{P} \leftarrow G_\phi\bigl(\bm{X}_v^{(t)}, AttnMask^{(t)}, \mathcal{C}\bigr)$ 
  %   \Comment{Predicted token distribution for all tokens}
  %   \State Perturb $\bm{P}$ with gumbel noise
  %   \State Select the $(N_f - U)$ masked positions with the lowest confidence in $\bm{P}$
  %   % \State \(\textbf{Sample tokens:}\)
  %   \State \(\textit{sampled\_ids} \;\leftarrow\; \mathrm{multinomial}(\bm{P},\, 1)\)
  %   % \State \(\textbf{Identify lowest-confidence positions:}\)
  %   \State \(\textit{low\_conf\_pos} 
  %      \;\leftarrow\; \mathrm{SelectLowConfidence}\bigl(\bm{P},\, (N_f - U)\bigr)\)
  %   % \State \(\textbf{Replace them with [MASK]:}\)
  %   \State \(\textit{sampled\_ids}[\textit{low\_conf\_pos}] 
  %      \;\leftarrow\; [\text{MASK}]\)
  %   % \State \(\textbf{Update tokens for this frame:}\)
  %   \State \(\bm{X}_v^{(t)} \;\leftarrow\; \textit{sampled\_ids}\)
  % \EndFor
  % \State Append \(\bm{X}_v^{(t)}\) to \(\mathcal{V}\) % used for decoding
  
  \State Generate $t$-th frame \(\bm{X}_v^{(t)}\) and append it to \(\mathcal{V}\)

  % Mask the generated frame
  \State \textbf{Cache partial observation of the generated frame}:
    \State \quad \(\widetilde{\bm{X}}_v^{(t)} 
      = \bm{M}_{inf} \odot \bm{X}_v^{(t)}
      + (1 - \bm{M}_{inf}) \odot [\text{MASK}]\)
    \State \quad \(\mathcal{C} \leftarrow G_\phi(\widetilde{\bm{X}}_v^{(t)}, AttnMask^{(t)}, \mathcal{C})\) \Comment{Store cache for the \(t\)-th frame} 
\EndFor
% \State Decode \(T\) latents \(\mathcal{V}\) to RGB space
% \State \textbf{return} the generated video frames
\end{algorithmic}
\end{algorithm}
\vspace{-0.3cm}
% \caption{
% \textbf{Inference algorithm with AR-DF.} 
% The complete version is included in the Appendix.
% }
% \caption{\textbf{Inference algorithm with AR-DF.}
% We first cache the text prompt in $\mathcal{C}$, then iteratively generate frame latents $\bm{X}_v^{(t)}$ in a MaskGiT-like fashion. 
% After each frame is generated, we create a partially masked version $\widetilde{\bm{X}}_v^{(t)}$ and pass it into the model to update $\mathcal{C}$, maintaining consistent context across frames.}
\end{minipage}
\caption{
\small
Left: \textbf{Training algorithm with AR-DF}. 
For simplicity, this only includes training on videos.
Right: \textbf{Inference algorithm with AR-DF.} 
A more detailed version for inference is included in the Appendix.
% A more detailed version of the inference algorithm
}
\vspace{-.5cm}
\end{figure}

Moreover, for a model that jointly processes text and visual tokens, the interplay between two modalities is significant to ensure the vision-language alignment.
However, the range of positions for representing texts or visual data tends to differ.
Contemporary visual generation systems are typically trained with extremely long and descriptive captions~\cite{2023dalle3,wan2.1}, despite the low latent resolution of visual data (\textit{e.g.}, a video of resolution $448\times256\times25$ becomes \(56 \times 32 \times 7\) after a \(8 \times 8 \times 4\) compression).
% Contemporary visual generation systems are typically trained with extremely long and descriptive captions~\cite{2023dalle3,wan2.1}, despite the low latent resolution of visual data (\textit{e.g.}, a video of resolution $448\times256\times25$ has a latent size of $56\times32\times7$ if the compression ratio of visual tokenizer is $8\times8\times4$).
To balance the two modalities, we propose scaling the 3D positions to ensure a balanced learning.
Specifically, we empirically scale the latent 3D positions to the RGB space by multiplying the compression ratio, as shown in~\cref{fig:mmrope_scaling}(a).
This simple scaling operation, from another perspective, improves the RoPE resolution for visual tokens by slightly accelerating the rotation speed. 
Experiments in the experiment section demonstrate its efficacy, thereby showing the importance of balancing the two modalities from the RoPE perspective.
We acknowledge, however, that given the autoregressive generation nature of videos, this scaling may not be the optimal solution.
More advanced and sophisticated solutions are left for future work.
% However, we acknowledge that this might not be the optimal solution considering the autoregressive generation nature of video data.
% More advanced and sophisticated solutions are left for future work.

% 需要带上scaled表示，scaled原因一个是balance两个modality，一个是增加RoPE resolution。
% 画一个图，是怎么scale的，就是每个维度的rope分布是如何的。

% 关于grouped frequency allocation有没有好的motivation。
% 关于grouped frequency allocation有没有更好的数学表示。
% 关于spatial scaling有没有更好的数学表示

% \vspace{-.2cm}
\subsection{Autoregressive Discrete Diffusion Forcing}
% \subsection{Discrete Diffusion Forcing with Temporal Tube Masking}
% \subsection{Autoregressive Consistent-Mask Diffusion Forcing}
\label{sec:tube_masking}
% \vspace{-.2cm}

The most naive generation paradigm (\textit{i.e.}, next-token prediction) suffers from incompatibility with the nature of videos and low generation efficiency, rendering it impractical for autoregressive visual generation.
% 表达的意思：这种naive的approach，不是特别的好，因为low efficiency。所以in this work，我们xxxx。
% To facilitate an LLM-like architecture to perform video generation, the most naive approach is to follow the paradigm next-token prediction~\cite{wang2024emu3}.
% However, its low efficiency renders it an impractical choice for autoregressive video generation.
In this work, we resort to discrete diffusion~\cite{xie2024show-o,swerdlow2025UniDisc} to generate visual content, together with spatially bidirectional and temporally causal token dependency.
% to enable a temporally autoregressive generation paradigm.
However, due to the autoregressive nature of \method, naive random masking (\textit{i.e.}, a globally random mask) or temporally independent masking (\textit{i.e.}, diffusion forcing~\cite{chen2024diffusion_forcing,song2025history-guided}) both lead to significant loss imbalance, \textit{i.e.}, visual tokens in the later frames tend to have much lower loss.
% Since the task difficulty of predicting frames with ample history frame context is considerably easier than predicting the first image given a text prompt or predicting the second frame given the first frame, the model would lean towards optimizing simpler tasks, leading to a degradation of temporal learning.
%%% shorter
Since the task difficulty of predicting frames with ample history frame context is considerably easier than predicting the first image given a text prompt, the model would lean towards optimizing simpler tasks, leading to a degradation of temporal learning.

\noindent\textbf{Training scheme.}
To resolve this issue, we build upon the basic nature of videos--spatial information redundancy.
The core issue of imbalanced loss during training stems from spatial information leakage.
% This issue is overlooked 
% The original diffusion forcing and its application to video generation
It is worth noting that the original video diffusion transformer~\cite{song2025history-guided} that adopts diffusion forcing does not encounter this issue due to its usage of bidirectional dependency.
Facing this challenge, we introduce Autoregressive Discrete Diffusion Forcing (AR-DF), which adopts temporal tube masking during the training of AR video generation.
Concretely, for every video, we randomly generate a mask pattern for the first frame and then apply this mask pattern repeatedly to later frames in this video.
If we denote the multi-modal token sequence $\bm{X}$ composed of text tokens $\bm{X}_p$ and visual tokens $\bm{X}_v$ and sample a mask ratio $\rho$, the training masking strategy in AR-DF can be formulated as:
\begin{equation}\small
    \bm{M}_i \sim \mathrm{Bernoulli}(1 - \rho) \quad \text{for } i=1, \dots, N_f
\end{equation}
\begin{equation}\small
    \widetilde{\bm{X}}_v^{(t)} \;=\; \bm{M} \odot \bm{X}_v^{(t)} \;+\; (1 - \bm{M})\,\odot\,[\text{MASK}], \quad t = 1,\dots,T
\end{equation}
\begin{equation}\small
    \widetilde{\bm{X}} \;=\;\bigl\{ \bm{X}_p,\; \widetilde{\bm{X}}_v^{(1)},\;\widetilde{\bm{X}}_v^{(2)},\;\dots,\;\widetilde{\bm{X}}_v^{(T)}\bigr\}
\end{equation}
where $N_f$ and $T$ denote the number of tokens in a latent frame and the number of latent frames in $\bm{X}_v$; 
$\bm{X}_v^{(t)}$ denotes the visual tokens of the $t$-th frame;
$\widetilde{\bm{X}}$ denotes the masked multi-modal token sequence prepared for training;
$\odot$ indicates Hadamard multiplication;
$\bm{M}$ denotes the mask pattern; 
$[\text{MASK}]$ denotes the mask token.
After the preparation of the token sequence, it is fed into the model for processing.
To ensure consistency with native LLMs and temporal causality in videos, we adopt a temporal causal mask $AttnMask$ for attention processing, as shown in~\cref{fig:mmrope_scaling}(b).
% To train the model, we use cross-entropy loss and compute the loss only on the unmasked token, 
To train the model, we use cross-entropy loss and compute the loss only on masked tokens, denoted as $\mathcal{L}(\widehat{\bm{X}}, \bm{X},\bm{M})$ and $\widehat{\bm{X}}$ is the token sequence after model processing.
The algorithm is formalized in~\cref{alg:AR-DF_training}.

\noindent\textbf{Inference scheme.}
After training with AR-DF, the most naive inference scheme (\textit{i.e.}, autoregressively generating video frames) would result in significant frame quality and motion degradation.
We observe that this is caused by inconsistent inference with training.
During training, later frames consistently have partial observation of history frames, while the inference stage does not align with this observation pattern.
% Therefore, given a caption, we first generate the first frame by running multiple steps, then save the generated frame for VAE decoding to RGB space.
% After that, the generated frame would randomly replace a pre-defined ratio $\rho_{inf}$ of tokens with the $[\text{MASK}]$ token, run the model again, and then cache the Keys and Values for later swift inference.
Therefore, given a caption, we first generate the first frame by running multiple steps and then randomly replace a pre-defined ratio $\rho_{inf}$ of tokens with the $[\text{MASK}]$ token for the generated image.
We infer the model with this partially observed image and cache the Keys and Values of this image for swift inference.
This process is repeated until the entire video is generated.
% 生成单帧的公式
% drop + cache；一个字典保存输出的完整frames；
% 
The algorithm is formalized in~\cref{alg:AR-DF_inference}.

% \subsection{Implementation}
% About image-video joint training derived from the token dependency strategy.

\begin{table*}[t]
    \centering
    \definecolor{lightblue}{RGB}{240,248,255}
    \newcommand{\colorrow}[1]{\rowcolor{lightgray} #1}
    \caption{
    \small
    \textbf{Performance comparison on GenEval.}
    % In the ``\#Params" column, we present the parameter counts of the entire model, including the main model and external models (\textit{e.g.}, text encoders or diffusion models).
    % In the ``\#Params" column, we present the parameter counts of the main model and \textcolor{gray}{external models} (\textit{e.g.}, pre-trained text encoders or diffusion models), for a fair comparison with unified models following~\cite{zhou2024transfusion}.
    % In the ``\#Params" column, we present the parameter counts of the visual generative model and the significant \textcolor{gray}{language encoders} for a fair comparison with unified models following~\cite{zhou2024transfusion}.
    % Tables below follow this paradigm.
    The ``\#Params" column shows the parameter counts of the generative model and the \textcolor{gray}{language encoder} for a fair comparison to unified models following~\cite{zhou2024transfusion}.
    % \textcolor{Watermelon_Red}{
    \textbf{Gen Rep} denotes the generation representation type: continuous or discrete.
    * denotes results without inference-time scaling.
    \dag ~denotes supervised fine-tuning on small-scale data.
    % }
    }
    % \vspace{-.3cm}
    \label{tab:t2i_geneval}
    \resizebox{1.\linewidth}{!}{
        \begin{tabular}{lcc|c|c|ccccccc}
            \toprule
            \textbf{Model} & \textbf{\#Params} & \textbf{\#Images} & \textbf{Gen Rep} & \textbf{Overall$\uparrow$} & \textbf{Single Obj.} & \textbf{Two Obj.} & \textbf{Counting} & \textbf{Colors} & \textbf{Position} & \textbf{Attr. Bind} \\
            % & \textbf{Ext. Encoder}
            \midrule
            % \midrule
            \multicolumn{10}{l}{\textbf{Diffusion models}} \\ % Span 10 columns
            \midrule
            SD v1.5~\cite{rombach2022stable_diffusion} & 0.9B \textcolor{gray}{+ 0.1B}  & 2B  &  Continuous & 0.43  & 0.97 & 0.38 & 0.35 & 0.76 & 0.04 & 0.06 \\
            % SD v2.1~\cite{rombach2022stable_diffusion} & 0.9B \textcolor{gray}{+ 0.3B}  & 2B & 0.50  &  0.98 & 0.51 & 0.44 & 0.85 & 0.07 & 0.17 \\
            % 1.3B
            SD-XL~\cite{2023SDXL} & 2.6B \textcolor{gray}{+ 0.8B} & -- &  Continuous & 0.55 & 0.98 & 0.74 & 0.39 & 0.85 & 0.15 & 0.23 \\
            % 3.4B
            % 2.6B \textcolor{gray}{+ 0.4B}
            SD 3~\cite{2024SD3} & 8.2B \textcolor{gray}{+ 2.8B}  & -- &  Continuous & 0.68 & 0.98 &  0.84 & \textbf{0.66} & 0.74 & 0.40 & 0.43 \\
            % 12.7B (from Mint) 
            DALL-E 2~\cite{2022DALLE2} & 4.2B \textcolor{gray}{+ 1.0B} & 650M &  Continuous &  0.52 & 0.94 & 0.66 & 0.49 & 0.77 & 0.10 & 0.19 \\
            % \textcolor{Watermelon_Red}{SANA 1.5}
            SANA 1.5~\cite{SANA-1.5} & 1.6B \textcolor{gray}{+ 2.6B} & 50M &  Continuous & 0.66 & -- & -- & -- & -- & -- & -- \\
            FLUX~\cite{flux2024} & 12B \textcolor{gray}{+ 2.5B} & -- &  Continuous & 0.67 & 0.99 & 0.85 & 0.75 & 0.77 & 0.22 & 0.42 \\
            % \textcolor{Watermelon_Red}{Meissonic}
            Meissonic~\cite{bai2025Meissonic} & 1B \textcolor{gray}{+ 0.4B} & 210M &  Discrete & 0.54 & 0.99 & 0.66 & 0.42 & 0.86 & 0.10 & 0.22 \\
            % \textcolor{Watermelon_Red}{Muddit}
            Muddit~\cite{shi2025Muddit} & 1B \textcolor{gray}{+ 0.4B} & 50M &  Discrete & 0.61 & 0.98 & 0.72 & 0.54 & 0.82 & 0.19 & 0.41 \\
            % \textcolor{Watermelon_Red}{UniDisc}
            UniDisc~\cite{swerdlow2025UniDisc} & 1.4B  & 280M &  Discrete & 0.42 & 0.92 & 0.47 & 0.15 & 0.67 & 0.13 & 0.19 \\
            % \textcolor{Watermelon_Red}{D-DiT}
            D-DiT~\cite{li2025D-Dit} & 2B  & -- &  Discrete & 0.65 & 0.97 & 0.80 & 0.54 & 0.76 & 0.32 & 0.50 \\
            % \textcolor{Watermelon_Red}{MMaDA}
            MMaDA~\cite{yang2025MMaDA} & 8B  & -- &  Discrete & 0.63 & 0.99 & 0.76 & 0.61 & 0.84 & 0.20 & 0.37 \\
            % 6.5B janusflow, infinity
            % 4.2B \textcolor{gray}{+ 1.0B}
            % DALL-E 3~\cite{2023dalle3} & -- & -- & 0.67 & 0.96 & 0.87 & 0.47 & 0.83 & 0.43 & 0.45 \\
            % FLUX~\cite{flux2024} & 12B \textcolor{gray}{+ 2.5B} & -- & 0.665 & 0.988 & 0.849 & 0.747 & 0.766 & 0.218 & 0.423 \\
            % IF-XL~\cite{2023IF} & 10.1B &  & 0.61 & 0.97 & 0.74 & 0.66 & 0.81 & 0.13 & 0.35 \\
            \midrule
            \multicolumn{10}{l}{\textbf{Autoregressive models}} \\ % Span 10 columns
            \midrule
            % LWM~\cite{liu2024LWM} & 7B & 1B & 0.47 & 0.93 & 0.41 & 0.46 & 0.79 & 0.09 & 0.15 \\
            SEED-X~\cite{ge2024seed-x} & 17B & -- &  Continuous & 0.49 & 0.97 & 0.58 & 0.26 & 0.80 & 0.19 & 0.14 \\
            Transfusion~\cite{zhou2024transfusion} & 7.3B & 3.5B &  Continuous & 0.63 & -- & -- & -- & -- & -- & -- \\
            Fluid~\cite{fan2024fluid} & 10.5B \textcolor{gray}{+ 4.7B} & 680M &  Continuous & 0.69 & 0.96 & 0.83 & 0.63 & 0.80 & 0.39 & 0.51 \\
            % \textcolor{Watermelon_Red}{Show-o2}
            Show-o2~\cite{xie2025show-o2} & 7B & 66M &  Continuous & 0.76 & \textbf{1.00} & 0.87 & 0.58 & \textbf{0.92} & 0.52 & 0.62 \\
            % \colorrow{Janus~\cite{2024Janus} & 1.3B &  & 0.61 & 0.97 & 0.68 & 0.30 & 0.84 & 0.46 & 0.42 \\}
            % \colorrow{JanusFlow~\cite{ma2024janusflow} & 1.3B &  & 0.63 & 0.97 & 0.59 & 0.45 & 0.83 & 0.53 & 0.42 \\}
            % \colorrow{\textbf{MINT (Ours)} & 1.3B &  & \textbf{0.73} & \textbf{0.98} & \textbf{0.82} & \textbf{0.66} & 0.79 & \textbf{0.55} & \textbf{0.56} \\}
            LlamaGen~\cite{sun2024LlamaGen} & 0.8B \textcolor{gray}{+ 2.9B} & 60M  & Discrete & 0.32 & 0.71 & 0.34 & 0.21 & 0.58 & 0.07 & 0.04 \\
            % Show-o~\cite{xie2024show-o} & 1.3B & 2B & 0.53 & 0.95 & 0.52 & 0.49 & 0.82 & 0.11 & 0.28 \\
            % Higher show-o
            Show-o~\cite{xie2024show-o} & 1.3B & 2B &  Discrete & 0.68 & 0.98 & 0.80 & 0.66 & 0.84 & 0.31 & 0.50 \\
            Chameleon~\cite{team2024Chameleon} & 34B  & 1.4B &  Discrete & 0.39 & -- & -- & -- & -- & -- & -- \\
            % \textcolor{Watermelon_Red}{Lumina-mGPT}
            Lumina-mGPT~\cite{liu2024Lumina-mGPT} & 7B  & 1.4B &  Discrete & 0.56 & -- & 0.77 & 0.27 & -- & -- & 0.32 \\
            % \textcolor{Watermelon_Red}{Lumina-mGPT 2.0\textsuperscript{*}} 
            Lumina-mGPT 2.0\textsuperscript{*}~\cite{xin2025lumina-mGPT-2} & 7B  & -- &  Discrete & 0.73 & \textbf{1.00} & \textbf{0.87} & 0.49 & 0.85 & 0.52 & 0.62 \\
            EMU3~\cite{zhou2024transfusion} & 8B & -- &  Discrete & 0.66 & 0.99 & 0.81 & 0.42 & 0.80 & 0.49 & 0.45 \\
            \midrule
            \method ($352\times 352$) & 1.5B & 60M &  Discrete & 0.601 & 0.959 & 0.732 & 0.375 & 0.774 & 0.365 & 0.400 \\
            % \method (Qwen32B caption, cfg = 16)
            \method ($352\times 352$) & 3.6B & 60M &  Discrete & 0.664 & 0.953 & 0.806 & 0.463 & 0.806 & 0.483 & 0.475 \\
            % \method (Qwen32B caption, cfg = 16)
            \method ($512\times 512$)\textsuperscript{\dag} & 1.5B & 60M &  Discrete & 0.725 & 0.984 & \textbf{0.869} & 0.519 & 0.862 & 0.558 & 0.558  \\
            %%% Original 3B
            % \method ($512\times 512$)\textsuperscript{\dag} & 3.6B & 60M &  Discrete & \textbf{0.777} & 0.991 & \textbf{0.897} & 0.566 & 0.878 & \textbf{0.640} & \textbf{0.690} \\
            %%% 3B 重新0.9预训练+SFT
            \method ($512\times 512$)\textsuperscript{\dag} & 3.6B & 60M &  Discrete & \textbf{0.791} & 0.991 & 0.866 & \textbf{0.678} & 0.840 & \textbf{0.683} & \textbf{0.688} \\
            \bottomrule
        \end{tabular}}
\vspace{-.4cm}
\end{table*}

% \vspace{-.2cm}
\subsection{Implementation}\label{sec:implementation}
% \vspace{-.2cm}

%%%%%%%%%%%%%%%%%%%%%%%% Original Architecture %%%%%%%%%%%%%%%%%%%%%%%%
% \noindent\textbf{Architecture.} 
% The architecture of \method follows Llama~\cite{touvron2023llama2,grattafiori2024llama-3}, meaning that it incorporates RMSNorm~\cite{zhang2019RMSNorm} and SwiGLU~\cite{shazeer2020SwiGLU} by default.
% To stabilize training, we integrate query-key normalization (QK-Norm) following Chameleon~\cite{team2024Chameleon}.
% % In Table~\cref{tab:architecture_training_summary}, we present the architectural details for different model scales.
% We have models of three scales (\method 0.5B, 1B and 3B), whose architectural details are placed in the Appendix.

%%%%%%%%%%%%%%%%%%%%%%%% Short Architecture %%%%%%%%%%%%%%%%%%%%%%%%
\noindent\textbf{Architecture.} 
\method's architecture follows Llama~\cite{touvron2023llama2,grattafiori2024llama-3}.
To stabilize training, we integrate QK-Norm following Chameleon~\cite{team2024Chameleon}.
We have models of three scales (0.5B, 1B and 3B), whose architectural details are placed in the Appendix.

%%%%%%%%%%%%%%%%%%%%%%%% Original Tokenizers %%%%%%%%%%%%%%%%%%%%%%%%
% \noindent\textbf{Tokenizers.}
% % To jointly process visual tokens and text tokens, 
% To unify visual and text token processing, we adopt the discrete version of Cosmos Tokenizer~\cite{agarwal2025cosmos} that achieves spatiotemporal compression rates of $8 \times 8 \times 4$.
% For text tokens, we retain Chameleon's text tokenizer~\cite{team2024Chameleon}.
% Therefore, \method's total codebook size is 129,536, partitioned into 65,536 text tokens and 64,000 visual tokens.

%%%%%%%%%%%%%%%%%%%%%%%% Short Tokenizers %%%%%%%%%%%%%%%%%%%%%%%%
\noindent\textbf{Tokenizers.}
We adopt the discrete Cosmos Tokenizer~\cite{agarwal2025cosmos} that achieves spatiotemporal compression rates of $8 \times 8 \times 4$.
For text tokens, we retain Chameleon's text tokenizer.
\method's total codebook size is 129,536, partitioned into 65,536 text tokens and 64,000 visual tokens.

\begin{table*}[t]
    \setlength{\tabcolsep}{3pt}
    \centering
    \definecolor{lightblue}{RGB}{240,248,255}
    \newcommand{\colorrow}[1]{\rowcolor{lightgray} #1}
    \caption{
    \small
    \textbf{Performance comparison on VBench-I2V benchmark.}
    We list partial metrics due to space limits.
    % \textcolor{Watermelon_Red}{\dag ~denotes supervised fine-tuning on small-scale data.}
    }
    % \vspace{-.3cm}
    \label{tab:i2v_vbench}
    \resizebox{1.\linewidth}{!}{
        \begin{tabular}{lcc|c|ccc|ccccc}
            \toprule
            \textbf{Model} & \textbf{\#Params} & \textbf{\#Videos} & \textbf{Gen Rep} & \textbf{Total$\uparrow$} & \textbf{I2V Score} & \textbf{Quality Score} & \textbf{I2V Sub.} & \textbf{I2V Back.} & \textbf{Sub. Cons.} & \textbf{Back. Cons.} & \textbf{Img. Quality} \\
            % \shortstack{A\\B}
            \midrule
            \multicolumn{11}{l}{\textbf{Diffusion models}} \\ % Span 10 columns
            \midrule
            VideoCrafter-I2V~\cite{chen2023videocrafter1} & 2.6B  & 20M & Continuous & 82.57  & 86.31 & 78.84 & 91.17 & 91.31 & 97.86 & 98.79 & 71.68 \\
            ConsistI2V~\cite{ren2024ConsistI2V} & 1.3B \textcolor{gray}{+ 0.3B} & 10M & Continuous & 84.07  & 91.91 & 76.22 & 95.82 & 95.95 & 95.27 & 98.28 & 66.92 \\
            SEINE~\cite{chen2023SEINE} & 0.9B \textcolor{gray}{+ 0.1B} & 10M & Continuous & 84.88  & 92.39 & 77.37 & 96.57 & 96.80 & 94.2 & 97.26 & 70.97 \\
            I2VGen-XL~\cite{zhang2023i2vgen-xl} & 1.4B \textcolor{gray}{+ 1.0B} & 35M & Continuous & 85.28  & 92.11 & 78.44 & 96.48 & 96.83 & 94.18 & 97.09 & 69.14 \\
            CogVideoX~\cite{yang2024cogvideox} & 5.6B \textcolor{gray}{+ 4.8B} & -- & Continuous & 86.70  & 94.79 & 78.61 & 97.19 & 96.74 & 94.34 & 96.42 & 70.01 \\
            \midrule
            \multicolumn{11}{l}{\textbf{Autoregressive models}} \\ % Span 10 columns
            \midrule
            % COSMOS-Video2World~\cite{sun2024LlamaGen}
            COSMOS~\cite{sun2024LlamaGen} & 5B \textcolor{gray}{+ 11B} & 100M  & Continuous & 84.16 & 92.51 & 75.81 & 95.99 & 97.36 & 97.12 & 96.59 & 59.90\\
            VideoMAR~\cite{yu2025videomar} & 1.4B \textcolor{gray}{+ 1.5B} & 0.5M  & Continuous & 84.82 & 94.02 & 75.6 & 97.85 & 98.38 & 97.13 & 97.20 & 62.34 \\
            \midrule
            \method ($672\times 384 \times25$) & 1.5B & 10M & Discrete & 84.16 & 91.80 & 76.53 & 96.06  & 96.58 & 95.90 & 96.25 & 67.06 \\
            % \method (Qwen32B caption, cfg = 16)
            \method ($672\times 384 \times25$) & 3.6B & 10M & Discrete & 84.72 & 93.34 & 76.10 & 97.42 & 97.40 & 97.42 & 96.91 & 69.23 \\
            % \method ($672\times 384 \times25$)\textsuperscript{\dag} & 3.6B & 10M & Discrete &  &  &  &  &   &   &   &  \\
            % \method (Qwen32B caption, cfg = 16)
            \bottomrule
        \end{tabular}}
\vspace{-.4cm}
\end{table*}

\begin{table*}[t]
    \setlength{\tabcolsep}{3pt}
    \centering
    \definecolor{lightblue}{RGB}{240,248,255}
    \newcommand{\colorrow}[1]{\rowcolor{lightgray} #1}
    \caption{
    \small
    \textbf{Performance comparison on VBench-T2V benchmark.}
    We list partial metrics due to space limits.
    % \textcolor{Watermelon_Red}{
    \dag ~denotes supervised fine-tuning on small-scale data.
    % }
    }
    % \vspace{-.3cm}
    \label{tab:t2v_vbench}
    \resizebox{1.\linewidth}{!}{
        \begin{tabular}{lcc|c|ccc|cccccc}
            \toprule
            \textbf{Model} & \textbf{\#Params} & \textbf{\#Videos} & \textbf{Gen Rep} & \textbf{Total$\uparrow$} & \textbf{Quality} & \textbf{Semantic} & \textbf{Sub. Cons.} & \textbf{Back. Cons.} & \textbf{Img. Quality} & \textbf{Obj. Class} & \textbf{Color} & \textbf{Overall Cons.} \\
            % \shortstack{A\\B}
            \midrule
            \multicolumn{11}{l}{\textbf{Diffusion models}} \\ % Span 10 columns
            \midrule
            ModelScopeT2V~\cite{wang2023modelscopeT2V} & 1.4B \textcolor{gray}{+ 0.3B} & 10M & Continuous & 75.75 & 78.05 & 66.54 & 89.87 & 95.29 & 58.57 & 82.25 & 81.72 & 25.67 \\
            InstructVideo~\cite{yuan2024instructvideo} & 1.4B \textcolor{gray}{+ 0.3B} & 10M & Continuous & 76.61 & 81.56 & 56.81 & 95.30 & 96.97 & 68.01 & 73.26 & 77.14 & 19.91 \\
            LaVie~\cite{wang2023lavie} & 2.5B \textcolor{gray}{+ 0.5B} & 25M & Continuous & 77.08 & 78.78 & 70.31 & 91.41 & 97.47 & 61.90 & 91.82 & 86.39 & 26.41 \\
            % Show-1~\cite{zhang2023show-1} & xxxB \textcolor{gray}{+ 4.8B} & 25M & 77.08 & 78.78 & 70.31 & 91.41 & 97.47 & 61.90 & 91.82 & 86.39 & 26.41 \\
            OpenSoraPlan V1.3~\cite{lin2024OpenSoraPlan} & 2.7B \textcolor{gray}{+ 13B} & 70M & Continuous & 77.23 & 80.14 & 65.62 & 97.79 & 97.24 & 56.21 & 85.56 & 79.30 & 24.47 \\
            % OpenSora V1.2~\cite{zheng2024OpenSora}  \\
            LTX-Video~\cite{hacohen2024LTX-Video} & 1.9B \textcolor{gray}{+ 4.8B} & -- & Continuous & 80.00 & 82.30 & 70.79 & 96.56 & 97.20 & 60.28 & 83.45 & 81.45 & 25.19 \\
            CogVideoX~\cite{yang2024cogvideox} & 5.6B \textcolor{gray}{+ 4.8B} & -- & Continuous & 81.91 & 83.05 & 77.33 & 96.45 & 96.71 & 63.33 & 85.07 & 83.03 & 27.65 \\
            % VideoCrafter-I2V~\cite{chen2023videocrafter1} & 1.3B  & 20M & 82.57  & 86.31 & 78.84 & 91.17 & 91.31 & 97.86 & 98.79 & 71.68 \\
            % ConsistI2V~\cite{ren2024ConsistI2V} & 1.3B \textcolor{gray}{+ 0.3B} & 10M & 84.07  & 91.91 & 76.22 & 95.82 & 95.95 & 95.27 & 98.28 & 66.92 \\
            % SEINE~\cite{chen2023SEINE} & 0.9B \textcolor{gray}{+ 0.1B} & 10M & 84.88  & 92.39 & 77.37 & 96.57 & 96.80 & 94.2 & 97.26 & 70.97 \\
            % I2VGen-XL~\cite{zhang2023i2vgen-xl} & 1.4B \textcolor{gray}{+ 1.0B} & 35M & 85.28  & 92.11 & 78.44 & 96.48 & 96.83 & 94.18 & 97.09 & 69.14 \\
            % CogVideoX~\cite{yang2024cogvideox} & 5.6B \textcolor{gray}{+ 4.8B} & -- & 86.70  & 94.79 & 78.61 & 97.19 & 96.74 & 94.34 & 96.42 & 70.01 \\
            \midrule
            \multicolumn{11}{l}{\textbf{Autoregressive models}} \\ % Span 10 columns
            \midrule
            % COSMOS-Video2World~\cite{sun2024LlamaGen} & 5B \textcolor{gray}{+ 11B} & --  & 84.16 & 92.51 & 75.81 & 95.99 & 97.36 & 97.12 & 96.59 & 59.90\\
            % CogVideo~\cite{sun2024LlamaGen} & 5B \textcolor{gray}{+ 11B} & --  & 84.16 & 92.51 & 75.81 & 95.99 & 97.36 & 97.12 & 96.59 & 59.90 \\
            CogVideo~\cite{hong2022cogvideo} & 9B & 5.4M & Continuous & 67.01 & 72.06 & 46.83 & 92.19 & 96.20 & 41.03 & 73.40  & 79.57 & 7.70\\
            NOVA~\cite{deng2024NOVA} & 0.6B \textcolor{gray}{+ 2.8B} & 20M & Continuous & 80.12 & 80.39 & 79.05 & -- & -- & -- & 92.00  & -- & -- \\
            % NOVA~\cite{deng2024NOVA} & 0.6B \textcolor{gray}{+ 2.8B} & 20M & 78.48 & 78.96 & 76.57 & -- & -- & -- & 91.36  & -- & -- \\
            EMU3~\cite{wang2024emu3} & 8B & -- & Discrete & 80.96 & 84.09 & 68.43 & 95.32 & 97.69 & -- & 86.17 & -- & -- \\
            \midrule
            \method ($672\times 384 \times25$) & 1.5B & 10M & Discrete & 76.34 & 78.27 & 68.65 & 96.04 & 96.29 & 56.16 & 90.47 & 77.49 & 25.07  \\
            % \method (Qwen32B caption, cfg = 16)
            \method ($672\times 384 \times25$) & 3.6B & 10M & Discrete & 78.32 & 79.52 & 73.51 & 95.51 & 96.50 & 58.04 & 90.05 & 82.00 & 25.29\\
            \method ($672\times 384 \times25$)\textsuperscript{\dag} & 1.5B & 10M & Discrete & 78.17 & 79.60 & 72.45 & 95.36 & 96.08 & 57.21 & 93.39 & 77.13 & 25.55 \\
            %%% Original 3B
            % \method ($672\times 384 \times25$)\textsuperscript{\dag} & 3.6B & 10M & Discrete & 78.90  & 79.83 & 75.21 & 96.78 & 96.60 & 61.92 & 94.38 & 81.14 & 25.57 \\
            % \method (Qwen32B caption, cfg = 16)
            %%% 3B 重新0.9预训练+SFT
            \method ($672\times 384 \times25$)\textsuperscript{\dag} & 3.6B & 10M & Discrete & 78.90  & 79.83 & 75.21 & 96.78 & 96.60 & 61.92 & 94.38 & 81.14 & 25.57 \\ 
            \bottomrule
        \end{tabular}}
\vspace{-6mm}
\end{table*}

\noindent\textbf{Stage-wise training.}
Due to the autoregressive nature of \method, the training of video generation can be categorized into training two capabilities:
\textbf{1)} text-to-image and \textbf{2)} image/images to video.
Although the incorporation of AR-DF training substantially ameliorates the imbalanced learning issue, we still observe that the later task is relatively easier.
% By observing frame-wise losses, we observe that following task 
Therefore, a stage-wise training scheme is mandatory to ensure successful video generation training.
% Concretely, the training commences with dedicated text-to-image training with 256p resolution, and then we perform image-video joint training to train the video generation capability on 256p resolution.
% Finally, we perform joint training on visual data of 384p resolution.
% \textcolor{Watermelon_Red}{
Our training consists of three pre-training stages and a supervised fine-tuning (SFT) stage.
The pre-training stages focus on learning the generation capabilities from a vast amount of data:
% } 
\textbf{1)} text-to-image training at 256p, 
\textbf{2)} joint image-video training at 256p, 
and \textbf{3)} joint fine-tuning at 384p.
% \textcolor{Watermelon_Red}{
The final SFT stage focuses on small-scale fine-tuning on data of high quality.
We use around 100k images and 20k videos.
% aligning the model's generative capabilities with human preferences.
% This stage could boost the performance of 
This process teaches the model to more precisely follow instructions and adhere to specific aesthetic styles.
We find that both models trained on 256p or 384p data can be boosted by simple SFT.
% }

Details on \textit{sequence formatting} and \textit{GPU memory friendly implementation} are placed in the Appendix.

\section{Experiments}\label{sec:exp}
% \vspace{-.3cm}

\subsection{Experimental Details}\label{sec:experimental_details}
% \vspace{-.2cm}

\noindent\textbf{Datasets.}
To train \method, we curate a image dataset containing 60 million images and a video dataset containing 10 million videos.
% We preserve the original aspect ratios of images and videos, with videos being clipped to 25 frames for training.
We preserve their original aspect ratios, with videos being clipped to 25 frames for training.
To ensure fine-grained vision-language alignment~\cite{2022DALLE2,wan2.1}, the visual data is re-captioned using visual-language models~\cite{lin2024vila} to obtain long and descriptive captions.

\noindent\textbf{Training, inference and model evaluation.}
% We train our model with a batch size of 768 for images or a batch size of 128 for videos in 256p.
% For the 1.5B model, we train our model with a learning rate of  in 256p resolution and xx in 480p resolution.
% Experiments are conducted on 4 NVIDIA A100s, with the batch size set to $8$ and the learning rate set to $1 \times 10^{-5}$.
% To strike a cost-performance balance, we fine-tune \method with default parameters for 20$\rm k$ steps if not otherwise stated.
% We train the model from scratch and the basic training hyper-parameters are detailed in~\cref{tab:architecture_training_summary}.
We train the model from scratch and the basic training hyper-parameters are placed in the Appendix.
During AR-DF training, the training mask ratio $\rho_{tra}$ is set to $0.7$ following~\cite{li2024MAR}.
During AR-DF inference, classier-free guidance (CFG) is utilized by default to enhance the generation quality. 
The inference mask ratio $\rho_{inf}$ is set to $0.7$ by default.
% \hangjie{Should check? Might use 0.7.}
The number of steps to generate one latent frame is set to $N_{steps}=50$ by default.
When evaluating on GenEval~\cite{ghosh2023geneval} and VBench~\cite{huang2024vbench}, the guidance scale is set to 16 if not otherwise specified.
Since we train our model only on detailed and descriptive captions, it is mandatory for us to rewrite captions using Qwen 32B~\cite{yang2024qwen2.5} when evaluating on GenEval~\cite{ghosh2023geneval}.
When evaluating on VBench~\cite{huang2024vbench}, we use its official long captions by default.

% \vspace{-.2cm}
% \subsection{Comparing with State-of-the-Arts}
\subsection{Comparison with Other Methods on Visual Generation}
% \vspace{-.2cm}

% \subsubsection{Text-to-image generation.}
\noindent\textbf{Text-to-image generation.} We compare \method with competitive image generation methods in~\cref{tab:t2i_geneval}.
% Compared with diffusion models, we observe that \method outperforms models of similar sizes by a margin (\textit{e.g.}, SD-XL~\cite{2023SDXL}) and is even on par with FLUX~\cite{flux2024}.
% % even if it uses continuous tokenizers.
% Compared with autoregressive models, we observe that \method is on par with EMU3~\cite{wang2024emu3}, while being significantly more efficient due to the discrete diffusion inference.
% \textcolor{Watermelon_Red}{
Compared with diffusion models, we observe that \method outperforms competitive models (\textit{e.g.}, SD-XL~\cite{2023SDXL} and FLUX~\cite{flux2024}) using continuous or discrete tokenizers by a clear margin, even if \method only uses a discrete tokenizer.
% even if it uses continuous tokenizers.
Compared with autoregressive models, we observe that \method outperforms EMU3~\cite{wang2024emu3}, Fluid~\cite{fan2024fluid} and Lumina-mGPT series~\cite{xin2025lumina-mGPT-2,liu2024Lumina-mGPT}, while using substantially fewer training images and retaining a small model.
% }
\method achieves superior results in terms of position and attribute binding, demonstrating excellent language understanding and vision-language alignment even without textual pre-training.
% \hangjie{Position and attr, excellent text understanding and VL alignment }
% \textcolor{Watermelon_Red}{
We also observe that with a small-scale supervised fine-tuning using high-quality data, the performance can be largely boosted on GenEval, especially on challenging metrics like position and attribute binding.
% }

\noindent\textbf{Image-to-video generation.}
Thanks to the autoregressive nature of \method, we can perform image-to-video generation by specifying the first frame, although we did not specifically train on this task.
Results are listed in~\cref{tab:i2v_vbench}.
\method outperforms the popular VideoCrafter-I2V model and is on par with the leading COSMOS-Video2World model, which uses substantially more data (100M$>$10M) and training resources (10000 H100s$>$48 H20s), demonstrating the promising performance of \method.

\begin{wraptable}{r}{0.6\textwidth} 
% \begin{table}[h!]
\centering
\vspace{-.5cm}
\caption{
\small
\textbf{Comparison with different training methods} for AR video generation.
Results are measured on GenEval, Vbench-Overall Consistency (OC), and Vbench-Imaging Quality (IQ).
}
% \vspace{-.3cm}
\resizebox{1.\linewidth}{!}{
\begin{tabular}{@{}lccc@{}}
\toprule
\textbf{Training Methods} & \textbf{GenEval} & \textbf{Vbench-OC} & \textbf{Vbench-IQ} \\
\midrule
Random masks        & 0.593 & 0.232 & 0.424 \\
Linear decay loss   & 0.588 & 0.228 & 0.410 \\
Step decay loss     & 0.584 & 0.234 & 0.464 \\
Diffusion forcing masks    & 0.590 & 0.241 & 0.540 \\
\midrule
AR-DF               & 0.591 & 0.249 & 0.559 \\
\bottomrule
\end{tabular}}
\vspace{-.3cm}
\label{tab:diff_training_methods}
% \end{table}
\end{wraptable}

\noindent\textbf{Text-to-video generation.}
We list the comparison in~\cref{tab:t2v_vbench}.
% We can see from the figure that 
Although \method utilizes discrete tokenizers and does not rely on a heavy pretrained text encoder, it can still be on par with diffusion models like OpenSoraPlan.
% even if we do not rely on a heavy pretrained text encoder.
% considering the model size and data we utilize for training.
Due to the autoregressive nature, the video quality can be ensured by first frame quality, enabling \method to excel in object-centric metrics (object class and color).

% 参考：
% PYRAMIDAL FLOW MATCHING FOR EFFICIENT VIDEO GENERATIVE MODELING 一些数据

% \subsection{Visualization}
% \textbf{Visual example comparison.}
% examples.
% \textbf{Human evaluation.}

% \vspace{-.2cm}
% \subsection{Analysis}
\subsection{Analysis and Ablation Studies}
% \vspace{-.2cm}

\begin{figure}[!t]
\centering
\includegraphics[width=1.0\linewidth]{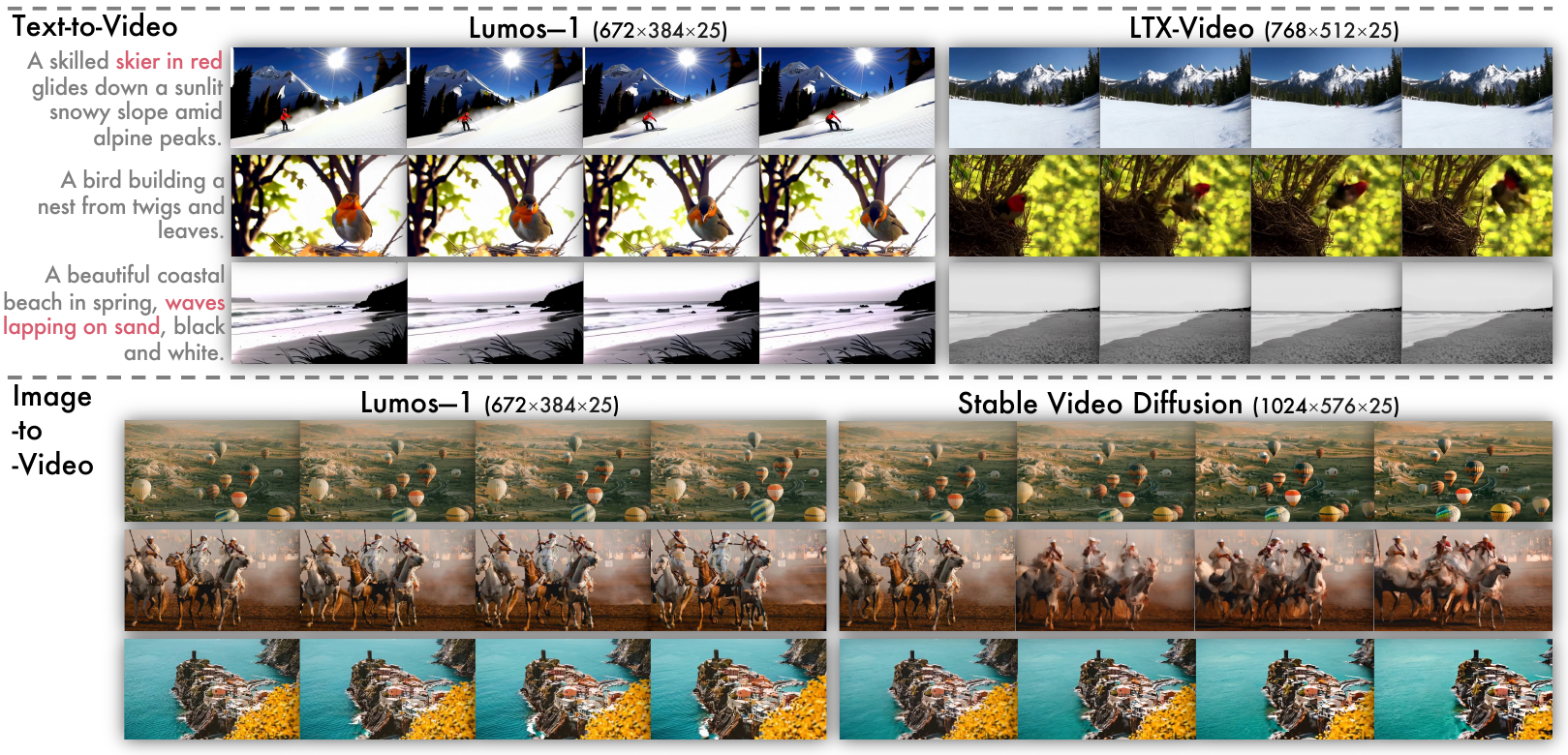}
\vspace{-.3cm}
\caption{
\textbf{Visual comparison} of \method with other methods on text/image-to-video tasks.
}
\vspace{-.3cm}
\label{fig:t2v_i2v_comparison}
\end{figure}

\noindent\textbf{Qualitative visual comparison.}
% \noindent\textbf{Visual example comparison.}
We compare \method with popular video generation methods in~\cref{fig:t2v_i2v_comparison}.
For T2V, the visual quality of our 384p videos does not trail 512p videos from LTX-Video.
In the provided case, \method generates better natural motion (water waves) and aligns with prompts better (skier in red and waves).
For I2V, \method handles multiple-object (multiple floating hot air balloons in example 1) and fine-grained motion (subtle ripples around the shoreline in example 3) substantially better than Stable Video Diffusion~\cite{blattmann2023StableVideoDiffusion} (SVD), which generates global camera movement only.
In example 2, SVD produces significant blurs, whereas \method animates subjects smoothly.
\textbf{\textit{More visualizations are placed in the Appendix}}.
% SVD do not use captions, .

% \noindent\textbf{The effectiveness of temporal tube masks in AR-DF training.}
% To demonstrate the effectiveness of using temporal tube masks, we compare it with using global random masks on video generation training.
% It is worth noting that the mask ratios $\rho$ between the two strategies are identical, set to $0.7$ following MAR~\cite{li2024MAR}.
% The frame-wise losses (latent frame 1, 3 and 6) are visualized in \hangjie{XXX}.
% We can observe that:
% 1) Due to the information leakage when using random masks, the loss of frame 6 converges significantly faster than using temporal tube masks, showing that random masks lead to an overly easy task that impairs performance.
% 2) When using temporal tube masks, at the initial stage of learning, the loss of frame 6 is even higher than previous frames, and their gap narrows gradually, showing that the model gradually learns to model temporal dynamics.

\begin{figure}[!t]
\centering
\includegraphics[width=1.0\linewidth]{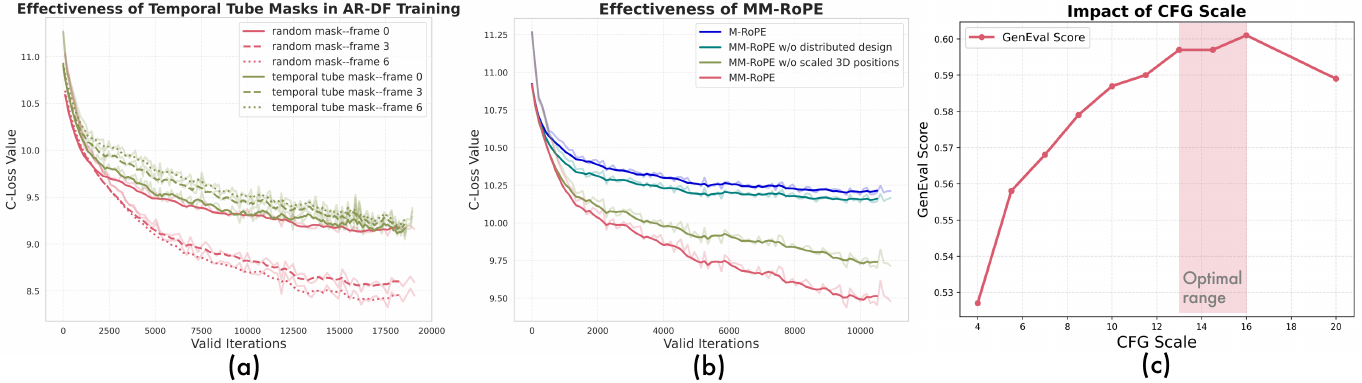}
\vspace{-.3cm}
\caption{
\small
(a) \textbf{Effectiveness of temporal-tube masks} during AR-DF training (on videos) using the 0.5B model;
(b) \textbf{Effectiveness of MM-RoPE};
(c) \textbf{Sensitivity analysis of CFG scale} on GenEval using 1B model.
}
\vspace{-.4cm}
\label{fig:AR-DF_curves}
\end{figure}

%%% This has been revised.
\noindent\textbf{Effectiveness of temporal‑tube masks during AR‑DF training.}
In~\cref{fig:AR-DF_curves}(a), we compare the frame-wise validation loss (frame 0, 3, 6) when using global random masks and temporal tube masks.
% The training mask ratios $\rho_{tra}$ are both set to $0.7$ following MAR~\cite{li2024MAR}.
For random masks, the loss of frame 6 decreases immediately and becomes lower than that of earlier frames.
This rapid fall indicates pronounced information leakage:
the model can reconstruct a masked token by attending to unmasked tokens in neighbouring frames instead of modelling genuine temporal dynamics, rendering the task overly easy.
For temporal tube masks, frame 6 is the hardest because it has the longest context to learn from, and all pixels in the same spatial locations are masked across the temporal axis, eliminating the learning shortcut. 
% eliminating the shortcut available with random masks. 
As iterations proceed, the gap between frames narrows and eventually levels out, demonstrating that the model is learning to propagate information through time rather than copying it.

% \hangjie{caption: Latent frame-wise loss comparison.}

\begin{figure}[!t]
\centering
\includegraphics[width=1.0\linewidth]{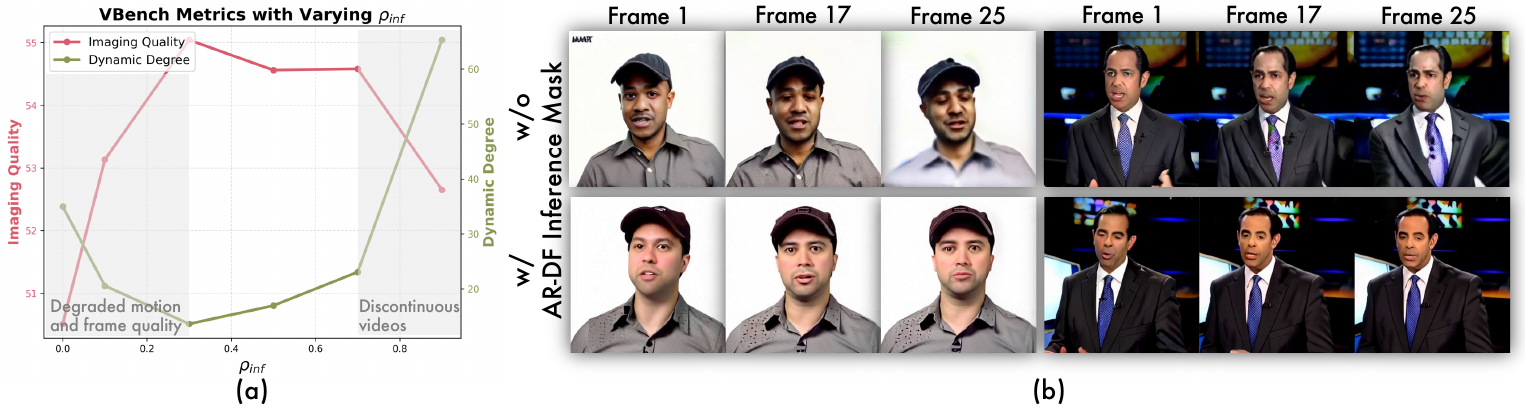}
\vspace{-.3cm}
\caption{
\small
(a) \textbf{Selected VBench metrics with varying inference-time mask ratio $\rho_{inf}$};
(b) \textbf{Text-to-video visualization comparing the effect of inference-time masks}.
% in AR-DF inference
(w/o: $\rho_{inf}=0.0$, w/: $\rho_{inf}=0.7$)
}
\vspace{-.6cm}
\label{fig:AR-DF_inference_masks}
\end{figure}

\begin{wraptable}{r}{0.6\textwidth} 
% \begin{table}[t]
\centering
% \scriptsize
\small
\vspace{-.3cm}
\caption{\small 
\textbf{Inference time analysis with different RoPEs and decoding strategies}. 
Results are reported on one H20 with batch size set to 1, $N_{steps}=50$ and CFG used by default.
}
% \vspace{-.2cm}
\label{tab:inference_efficiency}
\resizebox{1.\linewidth}{!}{
\begin{tabular}{lccc}
\toprule
 & \textbf{1D RoPE} (1B / 3B) & \textbf{M‑RoPE} (1B / 3B) & \textbf{MM‑RoPE} (1B / 3B) \\
\midrule
Image ($448\times256$)         & 7.4s / 16.3s   & 7.7s / 16.9s   & 7.7s / 16.9s   \\
Video ($448\times256\times25$) & 75.1s / 173.6s & 77.8s / 178.5s   & 77.8s / 178.5s \\
\bottomrule
\toprule
 & \textbf{Next-Token Pred} & \textbf{Mask Pred w/o KV Cache} & \textbf{Mask Pred w/ KV Cache} \\
\midrule
 Video ($448\times256\times25$) & 960.0s & 383.0s & 77.8s \\
\bottomrule
\end{tabular}}
\vspace{-.3cm}
% \end{table}
\end{wraptable}

%%% This has been revised.
\noindent\textbf{Effect of AR‑DF inference masks and sensitivity to $\rho_{inf}$.}
AR‑DF requires the same partial‑context masking at inference as during training; omitting these masks severely harms quality.
In~\cref{fig:AR-DF_inference_masks}(b), we observe that the ``w/o inference mask'' setting produces visible artifacts and flickering, whereas using masks preserves coherence.
In~\cref{fig:AR-DF_inference_masks}(a), we select two VBench metrics (imaging quality and dynamic degree) to assess the impact of $\rho_{inf}$.
% to quantitatively assess the impact of $\rho_{inf}$.
A broad plateau between $0.3$ and $0.7$ yields smooth, visually pleasing videos.
% When $\rho_{inf}$ is below $0.3$, insufficient context forces the model to degrade both motion and per‑frame quality, thus inflating dynamic degree values. 
When $\rho_{inf} < 0.3$, insufficient context causes motion and per‑frame quality degradation, thus inflating dynamic degree values. 
% When $\rho_{inf}$ is above $0.7$, overly aggressive masking disrupts temporal continuity, driving up dynamic degree values.
When $\rho_{inf} > 0.7$, aggressive masking disrupts temporal continuity, driving up dynamic degree values.
We set $\rho_{inf}$ to $0.7$ to ensure obvious motion. 
% We empirically set $\rho_{inf}$ to $0.7$ to ensure obvious motion. 

\noindent\textbf{Training strategy comparison.}
Temporal tube masking is a key in AR-DF training.
In~\cref{tab:diff_training_methods}, we compare it with other methods, including global random masks, loss reweighting (step decay loss~\cite{wang2024loong} and its modified version, linear decay loss), and diffusion forcing masks, by performing joint training on \method 1B using 8 H20 for 10k steps.
We observe that loss reweighting cannot solve the information leakage issue.
Diffusion forcing can ameliorate this, but temporal tube masking in AR-DF is particularly suited for videos, thus surpassing all other methods.

\begin{wraptable}{r}{0.5\textwidth} 
% \begin{table}[h!]
\centering
\vspace{-.5cm}
\caption{
\small
\textbf{Ablation study on MM-RoPE.}
% Results are measured on GenEval, Vbench-Overall Consistency (OC), and Vbench-Imaging Quality (IQ).
}
% \vspace{-.2cm}
\resizebox{1.\linewidth}{!}{
\begin{tabular}{@{}ccccc@{}}
\toprule
\textbf{Distributed} & \textbf{Scaled 3D} & \textbf{GenEval} & \textbf{Vbench-OC} & \textbf{Vbench-IQ} \\
\midrule
&            & 0.310 & 0.122 & 0.350 \\
& \checkmark & 0.414 & 0.211 & 0.451 \\
\checkmark & & 0.566 & 0.245 & 0.535 \\
\midrule
\checkmark & \checkmark & 0.591 & 0.249 & 0.559 \\
\bottomrule
\toprule
\multicolumn{2}{c}{\textbf{Scaling Factors}} & \textbf{GenEval} & \textbf{Vbench-OC} & \textbf{Vbench-IQ} \\
\midrule
\multicolumn{2}{c}{(1, 1, 1)}         & 0.566 & 0.245 & 0.535 \\
\multicolumn{2}{c}{(2, 4, 4)}         & 0.583 & 0.248 & 0.545 \\
\multicolumn{2}{c}{(4, 8, 8)}         & 0.591 & 0.249 & 0.559 \\
\multicolumn{2}{c}{(8, 16, 16)}       & 0.593 & 0.250 & 0.553 \\
\bottomrule
\end{tabular}}
\vspace{-.3cm}
\label{tab:MM-RoPE_ablation_study}
% \end{table}
\end{wraptable}

\noindent\textbf{Efficacy of MM-RoPE.}
% \hangjie{XXX} 
\cref{fig:AR-DF_curves}(b) plots the validation loss of the 0.5B model under four RoPE settings.
Note that M‑RoPE means that both designs are removed.
We can observe that MM‑RoPE consistently converges faster and settles at the lowest loss, confirming the benefit of modelling fine‑grained spatiotemporal information.
Although ablating either component raises the loss, dropping the distributed design hurts more than dropping the design of scaled position, indicating that comprehensive frequency allocation is the dominant factor.
Removing both enhancements gives the slowest convergence and the highest plateau, showing that the two mechanisms are complementary for effective video generation.
We quantitatively ablate on the MM-RoPE design in~\cref{tab:MM-RoPE_ablation_study} by performing joint training on \method 1B using 8 H20 for 10k steps.
We can find that the distributed design contributes significantly to the boost and speeds up the model's convergence.
By varying the scaling factors, we confirm that (4,8,8) is a suitable and well-justified choice for visual generation.

%%%%%%%%%%%%%%%%%%%%%%%%%%% Inference overhead analysis: Two separate %%%%%%%%%%%%%%%%%%%%%%%%%%%%%
% \noindent\textbf{Inference overhead analysis for MM-RoPE.}
% % \textbf{Additional inference overhead of MM-RoPE.}
% Similar to M-RoPE, MM-RoPE needs to locate the start of visual tokens and then apply the RoPE mechanism, which requires a small amount of computation.
% In~\cref{tab:inference_efficiency}, we compare the inference speed of generating images and videos using the vanilla 1D RoPE, M-RoPE and MM-RoPE.
% We can observe that:
% 1) Compared with 1D RoPE, the incorporation of 3D priors introduces only 3.5\%-4.1\% inference latency.
% 2) Compared to M-RoPE, MM-RoPE introduces no additional latency.

% \noindent\textbf{Inference speed analysis.}
% We also observe that MM-RoPE introduces no additional inference latency compared with M-RoPE, as detailed in the Appendix.

%%%%%%%%%%%%%%%%%%%%%%%%%%% Inference overhead analysis: One para %%%%%%%%%%%%%%%%%%%%%%%%%%%%%
\noindent\textbf{Inference time analysis.}
% \noindent\textbf{Inference overhead analysis.}
% Similar to M-RoPE, MM-RoPE needs to locate the start of visual tokens and then apply the RoPE mechanism, which requires a small amount of computation.
In~\cref{tab:inference_efficiency}, we compare the inference speed using the vanilla 1D RoPE, M-RoPE and MM-RoPE.
We can observe that:
\textbf{1)} Compared with 1D RoPE, the incorporation of 3D priors introduces only 3.5\%-4.1\% inference latency.
\textbf{2)} Compared to M-RoPE, MM-RoPE introduces no additional latency.
We also compare the speed using different decoding strategies.
% in~\cref{tab:inference_efficiency}.
We observe a clear efficiency boost compared with vanilla next-token prediction by using discrete diffusion (mask prediction) and KV cache.

\begin{wraptable}{r}{0.45\textwidth} 
% \begin{table}[h!]
\centering
\vspace{-.5cm}
\caption{
\small
% \textcolor{Watermelon_Red}{
\textbf{Performance comparison with other types of RoPE} on GenEval and VBench.
% Results are measured on GenEval, Vbench-Overall Consistency (OC), and Vbench-Imaging Quality (IQ).
% }
}
% \vspace{-.2cm}
\resizebox{1.\linewidth}{!}{
\begin{tabular}{@{}lccc@{}}
\toprule
\textbf{RoPE Type} & \textbf{GenEval} & \textbf{Vbench-OC} & \textbf{Vbench-IQ} \\
\midrule
M-RoPE    & 0.310 & 0.122 & 0.350 \\
U-RoPE    & 0.402 & 0.165 & 0.423 \\
IL-RoPE   & 0.541 & 0.225 & 0.513 \\
VideoRoPE & 0.569 & 0.243 & 0.540 \\
HoPE      & 0.570 & 0.246 &0.545  \\
\midrule
MM-RoPE   & 0.591 & 0.249 & 0.559 \\
\bottomrule
\end{tabular}}
\vspace{-.3cm}
\label{tab:diff_RoPE_comparison_main_paper}
% \end{table}
\end{wraptable}

% \textcolor{Watermelon_Red}{
\noindent\textbf{Comparison with different RoPE designs.}
To validate the efficacy of MM-RoPE, we benchmark it against M-RoPE~\cite{wang2024qwen2-vl}, VideoRoPE~\cite{wei2025VideoRoPE}, U-RoPE~\cite{tang2025U-RoPE}, IL-RoPE~\cite{liao2025Mogao} and HoPE~\cite{li2025HoPE}. 
As shown in~\cref{tab:diff_RoPE_comparison_main_paper}, MM-RoPE consistently outperforms them. 
This superiority stems from its unique design, which holistically incorporates a comprehensive frequency allocation strategy and a strategic scaling for modality alignment and resolution enhancement. 
A detailed breakdown of each method and a full analysis are provided in~\cref{app:more_analysis}.
% }

\noindent\textbf{Sensitivity analysis of CFG scale.}
We study the impact of the guidance scale on GenEval using 1B model in~\cref{fig:AR-DF_curves}(c).
We find that scale values from 13 to 16 (default value) lead to decent results.

% square and tall

% \hangjie{Should be appended if there is room.}
% We test on 
% Although there is only very limited data of aspect ratio ()

% \textbf{Failure cases and future work discussions.}
\vspace{-.2cm}
\section{Conclusion}
\vspace{-.2cm}
%%%%%%% v2 %%%%%%
In this paper, we introduce \method, which utilizes the LLM architecture for AR video generation.
We propose MM-RoPE for better spatiotemporal dynamics modeling and propose AR-DF for effective training and inference considering intra-frame bidirectionality and inter-frame temporal causality.
We anticipate that \method represents a significant step toward building a foundational unified model.
%%%%%%% v1 %%%%%%
% In this paper, we introduce \method, a model that utilizes the LLM architecture for autoregressive video generation.
% We study the basic techniques of RoPE and propose MM-RoPE for better spatiotemporal dynamics modeling.
% Moreover, we propose AR-DF for effective training and inference under the token dependency strategy of intra-frame bidirectionality and inter-frame temporal causality.
% We anticipate this to be an important step on the way to a foundational unifed model.

\section*{Ethics Statement}
\vspace{-.2cm}

Our work introduces \method, an autoregressive model for text-to-video generation. 
Similar to other generative models, this technology carries societal risks if misused. 
The primary concerns include the potential for creating deceptive or misleading content for misinformation campaigns, the generation of harmful or disturbing visuals, and the potential amplification of societal biases present in the training data.
To address these, we have taken several steps and advocate for further safeguards, as outlined in~\cref{app:societal_impact_safeguards}. 
We firmly state that \method is currently a research-oriented project. 
We recommend that any future public release of this technology must be preceded by rigorous safety measures, also detailed in~\cref{app:societal_impact_safeguards}.

\section*{Reproducibility Statement}
\vspace{-.2cm}

We are committed to ensuring the reproducibility of our work. 
All essential details for reproducing our results are provided within the paper and the appendix.
The details of architectures and their training details are presented in and \cref{sec:implementation}, \cref{sec:experimental_details} and \cref{app:more_architectural_implementation_details}.
Our technical contribution, MM-RoPE and AR-DF, are described with comprehensive details in~\cref{sec:method}.
The evaluation protocols are detailed in~\cref{sec:experimental_details}.
We will ensure the release of our source code, pre-trained model weights, and evaluation scripts upon publication of this work.
We can also provide any source code if any reviewer asks for a detailed implementation.

% \begin{itemize}
%     \item Architectures and their training details: These details are presented in and \cref{sec:implementation}, \cref{sec:experimental_details} and \cref{app:more_architectural_implementation_details}.
    
%     \item Methods: Our technical contribution, MM-RoPE and AR-DF, are described with comprehensive details in~\cref{sec:method}.

%     \item 
% \end{itemize}

% https://iclr.cc/Conferences/2026/AuthorGuide

\bibliography{refs}

\begin{thebibliography}{100}

\bibitem{agarwal2025cosmos}
Niket Agarwal, Arslan Ali, Maciej Bala, Yogesh Balaji, Erik Barker, Tiffany Cai, Prithvijit Chattopadhyay, Yongxin Chen, Yin Cui, Yifan Ding, et~al.
\newblock Cosmos world foundation model platform for physical ai.
\newblock {\em arXiv preprint arXiv:2501.03575}, 2025.

\bibitem{agrawal2024pixtral-12B}
Pravesh Agrawal, Szymon Antoniak, Emma~Bou Hanna, Baptiste Bout, Devendra Chaplot, Jessica Chudnovsky, Diogo Costa, Baudouin De~Monicault, Saurabh Garg, Theophile Gervet, et~al.
\newblock Pixtral 12b.
\newblock {\em arXiv preprint arXiv:2410.07073}, 2024.

\bibitem{bai2025Meissonic}
Jinbin Bai, Tian Ye, Wei Chow, Enxin Song, Qing-Guo Chen, Xiangtai Li, Zhen Dong, Lei Zhu, and Shuicheng Yan.
\newblock Meissonic: Revitalizing masked generative transformers for efficient high-resolution text-to-image synthesis.
\newblock In {\em The Thirteenth International Conference on Learning Representations}.

\bibitem{bai2023qwen}
Jinze Bai, Shuai Bai, Yunfei Chu, Zeyu Cui, Kai Dang, Xiaodong Deng, Yang Fan, Wenbin Ge, Yu~Han, Fei Huang, et~al.
\newblock Qwen technical report.
\newblock {\em arXiv preprint arXiv:2309.16609}, 2023.

\bibitem{2023dalle3}
James Betker, Gabriel Goh, Li~Jing, Tim Brooks, Jianfeng Wang, Linjie Li, Long Ouyang, Juntang Zhuang, Joyce Lee, Yufei Guo, et~al.
\newblock Improving image generation with better captions.
\newblock {\em Computer Science}, 2023.

\bibitem{blattmann2023StableVideoDiffusion}
Andreas Blattmann, Tim Dockhorn, Sumith Kulal, Daniel Mendelevitch, Maciej Kilian, Dominik Lorenz, Yam Levi, Zion English, Vikram Voleti, Adam Letts, et~al.
\newblock Stable video diffusion: Scaling latent video diffusion models to large datasets.
\newblock {\em arXiv preprint arXiv:2311.15127}, 2023.

\bibitem{Muse}
Huiwen Chang, Han Zhang, Jarred Barber, AJ~Maschinot, Jose Lezama, Lu~Jiang, Ming-Hsuan Yang, Kevin Murphy, William~T Freeman, Michael Rubinstein, et~al.
\newblock Muse: Text-to-image generation via masked generative transformers.
\newblock {\em arXiv preprint arXiv:2301.00704}, 2023.

\bibitem{chang2022MaskGiT}
Huiwen Chang, Han Zhang, Lu~Jiang, Ce~Liu, and William~T Freeman.
\newblock Maskgit: Masked generative image transformer.
\newblock In {\em Proceedings of the IEEE/CVF Conference on Computer Vision and Pattern Recognition}, pages 11315--11325, 2022.

\bibitem{chen2024diffusion_forcing}
Boyuan Chen, Diego~Marti Monso, Yilun Du, Max Simchowitz, Russ Tedrake, and Vincent Sitzmann.
\newblock Diffusion forcing: Next-token prediction meets full-sequence diffusion.
\newblock {\em arXiv preprint arXiv:2407.01392}, 2024.

\bibitem{chen2024JEPA-T2I_VoPE}
Dengsheng Chen, Jie Hu, Tiezhu Yue, and Xiaoming Wei.
\newblock High-resolution image synthesis via next-token prediction.
\newblock {\em arXiv preprint arXiv:2411.14808}, 2024.

\bibitem{chen2023videocrafter1}
Haoxin Chen, Menghan Xia, Yingqing He, Yong Zhang, Xiaodong Cun, Shaoshu Yang, Jinbo Xing, Yaofang Liu, Qifeng Chen, Xintao Wang, et~al.
\newblock Videocrafter1: Open diffusion models for high-quality video generation.
\newblock {\em arXiv preprint arXiv:2310.19512}, 2023.

\bibitem{chen2025blip3o}
Jiuhai Chen, Zhiyang Xu, Xichen Pan, Yushi Hu, Can Qin, Tom Goldstein, Lifu Huang, Tianyi Zhou, Saining Xie, Silvio Savarese, et~al.
\newblock Blip3-o: A family of fully open unified multimodal models-architecture, training and dataset.
\newblock {\em arXiv preprint arXiv:2505.09568}, 2025.

\bibitem{chen2023SEINE}
Xinyuan Chen, Yaohui Wang, Lingjun Zhang, Shaobin Zhuang, Xin Ma, Jiashuo Yu, Yali Wang, Dahua Lin, Yu~Qiao, and Ziwei Liu.
\newblock Seine: Short-to-long video diffusion model for generative transition and prediction.
\newblock In {\em The Twelfth International Conference on Learning Representations}, 2023.

\bibitem{dao2023Flashattention-2}
Tri Dao.
\newblock Flashattention-2: Faster attention with better parallelism and work partitioning.
\newblock {\em arXiv preprint arXiv:2307.08691}, 2023.

\bibitem{deng2024NOVA}
Haoge Deng, Ting Pan, Haiwen Diao, Zhengxiong Luo, Yufeng Cui, Huchuan Lu, Shiguang Shan, Yonggang Qi, and Xinlong Wang.
\newblock Autoregressive video generation without vector quantization.
\newblock {\em arXiv preprint arXiv:2412.14169}, 2024.

\bibitem{2024SD3}
Patrick Esser, Sumith Kulal, Andreas Blattmann, Rahim Entezari, Jonas M{\"u}ller, Harry Saini, Yam Levi, Dominik Lorenz, Axel Sauer, Frederic Boesel, et~al.
\newblock Scaling rectified flow transformers for high-resolution image synthesis.
\newblock 2024.

\bibitem{fan2024fluid}
Lijie Fan, Tianhong Li, Siyang Qin, Yuanzhen Li, Chen Sun, Michael Rubinstein, Deqing Sun, Kaiming He, and Yonglong Tian.
\newblock Fluid: Scaling autoregressive text-to-image generative models with continuous tokens.
\newblock {\em arXiv preprint arXiv:2410.13863}, 2024.

\bibitem{gao2024Tad-RoPE}
Mingze Gao, Jingyu Liu, Mingda Li, Jiangtao Xie, Qingbin Liu, Bo~Zhao, Xi~Chen, and Hui Xiong.
\newblock Tc-llava: Rethinking the transfer from image to video understanding with temporal considerations.
\newblock {\em arXiv preprint arXiv:2409.03206}, 2024.

\bibitem{ge2024seed-x}
Yuying Ge, Sijie Zhao, Jinguo Zhu, Yixiao Ge, Kun Yi, Lin Song, Chen Li, Xiaohan Ding, and Ying Shan.
\newblock Seed-x: Multimodal models with unified multi-granularity comprehension and generation.
\newblock {\em arXiv preprint arXiv:2404.14396}, 2024.

\bibitem{ghosh2023geneval}
Dhruba Ghosh, Hannaneh Hajishirzi, and Ludwig Schmidt.
\newblock Geneval: An object-focused framework for evaluating text-to-image alignment.
\newblock {\em Advances in Neural Information Processing Systems}, 36:52132--52152, 2023.

\bibitem{grattafiori2024llama-3}
Aaron Grattafiori, Abhimanyu Dubey, Abhinav Jauhri, Abhinav Pandey, Abhishek Kadian, Ahmad Al-Dahle, Aiesha Letman, Akhil Mathur, Alan Schelten, Alex Vaughan, et~al.
\newblock The llama 3 herd of models.
\newblock {\em arXiv preprint arXiv:2407.21783}, 2024.

\bibitem{gu2025FAR-video}
Yuchao Gu, Weijia Mao, and Mike~Zheng Shou.
\newblock Long-context autoregressive video modeling with next-frame prediction.
\newblock {\em arXiv preprint arXiv:2503.19325}, 2025.

\bibitem{hacohen2024LTX-Video}
Yoav HaCohen, Nisan Chiprut, Benny Brazowski, Daniel Shalem, Dudu Moshe, Eitan Richardson, Eran Levin, Guy Shiran, Nir Zabari, Ori Gordon, et~al.
\newblock Ltx-video: Realtime video latent diffusion.
\newblock {\em arXiv preprint arXiv:2501.00103}, 2024.

\bibitem{he2025NAR}
Yefei He, Yuanyu He, Shaoxuan He, Feng Chen, Hong Zhou, Kaipeng Zhang, and Bohan Zhuang.
\newblock Neighboring autoregressive modeling for efficient visual generation.
\newblock {\em arXiv preprint arXiv:2503.10696}, 2025.

\bibitem{ho2020denoising_DDPM}
Jonathan Ho, Ajay Jain, and Pieter Abbeel.
\newblock Denoising diffusion probabilistic models.
\newblock {\em Advances in Neural Information Processing Systems}, 33:6840--6851, 2020.

\bibitem{ho2022video_diffusion_models}
Jonathan Ho, Tim Salimans, Alexey Gritsenko, William Chan, Mohammad Norouzi, and David~J Fleet.
\newblock Video diffusion models.
\newblock {\em arXiv preprint arXiv:2204.03458}, 2022.

\bibitem{hong2022cogvideo}
Wenyi Hong, Ming Ding, Wendi Zheng, Xinghan Liu, and Jie Tang.
\newblock Cogvideo: Large-scale pretraining for text-to-video generation via transformers.
\newblock {\em arXiv preprint arXiv:2205.15868}, 2022.

\bibitem{huang2025revisiting_RoPE}
Jie Huang, Xuejing Liu, Sibo Song, Ruibing Hou, Hong Chang, Junyang Lin, and Shuai Bai.
\newblock Revisiting multimodal positional encoding in vision-language models.
\newblock {\em arXiv preprint arXiv:2510.23095}, 2025.

\bibitem{huang2025NFIG}
Zhihao Huang, Xi~Qiu, Yukuo Ma, Yifu Zhou, Chi Zhang, and Xuelong Li.
\newblock Nfig: Autoregressive image generation with next-frequency prediction.
\newblock {\em arXiv preprint arXiv:2503.07076}, 2025.

\bibitem{huang2024vbench}
Ziqi Huang, Yinan He, Jiashuo Yu, Fan Zhang, Chenyang Si, Yuming Jiang, Yuanhan Zhang, Tianxing Wu, Qingyang Jin, Nattapol Chanpaisit, et~al.
\newblock Vbench: Comprehensive benchmark suite for video generative models.
\newblock In {\em Proceedings of the IEEE/CVF Conference on Computer Vision and Pattern Recognition}, pages 21807--21818, 2024.

\bibitem{jin2024pyramidal_flow}
Yang Jin, Zhicheng Sun, Ningyuan Li, Kun Xu, Hao Jiang, Nan Zhuang, Quzhe Huang, Yang Song, Yadong Mu, and Zhouchen Lin.
\newblock Pyramidal flow matching for efficient video generative modeling.
\newblock {\em arXiv preprint arXiv:2410.05954}, 2024.

\bibitem{kondratyuk2023VideoPoet}
Dan Kondratyuk, Lijun Yu, Xiuye Gu, Jos{\'e} Lezama, Jonathan Huang, Grant Schindler, Rachel Hornung, Vighnesh Birodkar, Jimmy Yan, Ming-Chang Chiu, et~al.
\newblock Videopoet: A large language model for zero-shot video generation.
\newblock {\em arXiv preprint arXiv:2312.14125}, 2023.

\bibitem{kong2024hunyuanvideo}
Weijie Kong, Qi~Tian, Zijian Zhang, Rox Min, Zuozhuo Dai, Jin Zhou, Jiangfeng Xiong, Xin Li, Bo~Wu, Jianwei Zhang, et~al.
\newblock Hunyuanvideo: A systematic framework for large video generative models.
\newblock {\em arXiv preprint arXiv:2412.03603}, 2024.

\bibitem{flux2024}
Black~Forest Labs.
\newblock Flux.
\newblock \url{https://github.com/black-forest-labs/flux}, 2024.

\bibitem{li2025HoPE}
Haoran Li, Yingjie Qin, Baoyuan Ou, Lai Xu, and Ruiwen Xu.
\newblock Hope: Hybrid of position embedding for length generalization in vision-language models.
\newblock {\em arXiv preprint arXiv:2505.20444}, 2025.

\bibitem{li2024MAR}
Tianhong Li, Yonglong Tian, He~Li, Mingyang Deng, and Kaiming He.
\newblock Autoregressive image generation without vector quantization.
\newblock {\em arXiv preprint arXiv:2406.11838}, 2024.

\bibitem{li2025D-Dit}
Zijie Li, Henry Li, Yichun Shi, Amir~Barati Farimani, Yuval Kluger, Linjie Yang, and Peng Wang.
\newblock Dual diffusion for unified image generation and understanding.
\newblock In {\em Proceedings of the Computer Vision and Pattern Recognition Conference}, pages 2779--2790, 2025.

\bibitem{li2024arlon}
Zongyi Li, Shujie Hu, Shujie Liu, Long Zhou, Jeongsoo Choi, Lingwei Meng, Xun Guo, Jinyu Li, Hefei Ling, and Furu Wei.
\newblock Arlon: Boosting diffusion transformers with autoregressive models for long video generation.
\newblock {\em arXiv preprint arXiv:2410.20502}, 2024.

\bibitem{liao2025Mogao}
Chao Liao, Liyang Liu, Xun Wang, Zhengxiong Luo, Xinyu Zhang, Wenliang Zhao, Jie Wu, Liang Li, Zhi Tian, and Weilin Huang.
\newblock Mogao: An omni foundation model for interleaved multi-modal generation.
\newblock {\em arXiv preprint arXiv:2505.05472}, 2025.

\bibitem{lin2024OpenSoraPlan}
Bin Lin, Yunyang Ge, Xinhua Cheng, Zongjian Li, Bin Zhu, Shaodong Wang, Xianyi He, Yang Ye, Shenghai Yuan, Liuhan Chen, et~al.
\newblock Open-sora plan: Open-source large video generation model.
\newblock {\em arXiv preprint arXiv:2412.00131}, 2024.

\bibitem{lin2024vila}
Ji~Lin, Hongxu Yin, Wei Ping, Pavlo Molchanov, Mohammad Shoeybi, and Song Han.
\newblock Vila: On pre-training for visual language models.
\newblock In {\em Proceedings of the IEEE/CVF conference on computer vision and pattern recognition}, pages 26689--26699, 2024.

\bibitem{liu2024deepseek-v3}
Aixin Liu, Bei Feng, Bing Xue, Bingxuan Wang, Bochao Wu, Chengda Lu, Chenggang Zhao, Chengqi Deng, Chenyu Zhang, Chong Ruan, et~al.
\newblock Deepseek-v3 technical report.
\newblock {\em arXiv preprint arXiv:2412.19437}, 2024.

\bibitem{liu2024Lumina-mGPT}
Dongyang Liu, Shitian Zhao, Le~Zhuo, Weifeng Lin, Yu~Qiao, Hongsheng Li, and Peng Gao.
\newblock Lumina-mgpt: Illuminate flexible photorealistic text-to-image generation with multimodal generative pretraining.
\newblock {\em arXiv preprint arXiv:2408.02657}, 2024.

\bibitem{liu2024mardini}
Haozhe Liu, Shikun Liu, Zijian Zhou, Mengmeng Xu, Yanping Xie, Xiao Han, Juan~C P{\'e}rez, Ding Liu, Kumara Kahatapitiya, Menglin Jia, et~al.
\newblock Mardini: Masked autoregressive diffusion for video generation at scale.
\newblock {\em arXiv preprint arXiv:2410.20280}, 2024.

\bibitem{loshchilov2018adamW}
Ilya Loshchilov and Frank Hutter.
\newblock Decoupled weight decay regularization.
\newblock In {\em ICLR}, 2018.

\bibitem{luo2024open-magvit2}
Zhuoyan Luo, Fengyuan Shi, Yixiao Ge, Yujiu Yang, Limin Wang, and Ying Shan.
\newblock Open-magvit2: An open-source project toward democratizing auto-regressive visual generation.
\newblock {\em arXiv preprint arXiv:2409.04410}, 2024.

\bibitem{ma2025token-shuffle}
Xu~Ma, Peize Sun, Haoyu Ma, Hao Tang, Chih-Yao Ma, Jialiang Wang, Kunpeng Li, Xiaoliang Dai, Yujun Shi, Xuan Ju, et~al.
\newblock Token-shuffle: Towards high-resolution image generation with autoregressive models.
\newblock {\em arXiv preprint arXiv:2504.17789}, 2025.

\bibitem{openai2023gpt4}
OpenAI.
\newblock {GPT-4} technical report, 2023.

\bibitem{ouyang2022InstructGPT}
Long Ouyang, Jeffrey Wu, Xu~Jiang, Diogo Almeida, Carroll Wainwright, Pamela Mishkin, Chong Zhang, Sandhini Agarwal, Katarina Slama, Alex Ray, et~al.
\newblock Training language models to follow instructions with human feedback.
\newblock {\em Advances in Neural Information Processing Systems}, 35:27730--27744, 2022.

\bibitem{pang2024next-patch-pred}
Yatian Pang, Peng Jin, Shuo Yang, Bin Lin, Bin Zhu, Zhenyu Tang, Liuhan Chen, Francis~EH Tay, Ser-Nam Lim, Harry Yang, et~al.
\newblock Next patch prediction for autoregressive visual generation.
\newblock {\em arXiv preprint arXiv:2412.15321}, 2024.

\bibitem{peebles2023DiT}
William Peebles and Saining Xie.
\newblock Scalable diffusion models with transformers.
\newblock In {\em Proceedings of the IEEE/CVF International Conference on Computer Vision}, pages 4195--4205, 2023.

\bibitem{2023SDXL}
Dustin Podell, Zion English, Kyle Lacey, Andreas Blattmann, Tim Dockhorn, Jonas M{\"u}ller, Joe Penna, and Robin Rombach.
\newblock {SDXL}: Improving latent diffusion models for high-resolution image synthesis.
\newblock In {\em ICLR}, 2024.

\bibitem{polyak2024MovieGen}
Adam Polyak, Amit Zohar, Andrew Brown, Andros Tjandra, Animesh Sinha, Ann Lee, Apoorv Vyas, Bowen Shi, Chih-Yao Ma, Ching-Yao Chuang, et~al.
\newblock Movie gen: A cast of media foundation models.
\newblock {\em arXiv preprint arXiv:2410.13720}, 2024.

\bibitem{2022DALLE2}
Aditya Ramesh, Prafulla Dhariwal, Alex Nichol, Casey Chu, and Mark Chen.
\newblock Hierarchical text-conditional image generation with {CLIP} latents.
\newblock {\em arXiv preprint arXiv:2204.06125}, 2022.

\bibitem{Dalle}
Aditya Ramesh, Mikhail Pavlov, Gabriel Goh, Scott Gray, Chelsea Voss, Alec Radford, Mark Chen, and Ilya Sutskever.
\newblock Zero-shot text-to-image generation.
\newblock In {\em International conference on machine learning}, pages 8821--8831. Pmlr, 2021.

\bibitem{ren2025xAR}
Sucheng Ren, Qihang Yu, Ju~He, Xiaohui Shen, Alan Yuille, and Liang-Chieh Chen.
\newblock Beyond next-token: Next-x prediction for autoregressive visual generation.
\newblock {\em arXiv preprint arXiv:2502.20388}, 2025.

\bibitem{ren2024M-VAR}
Sucheng Ren, Yaodong Yu, Nataniel Ruiz, Feng Wang, Alan Yuille, and Cihang Xie.
\newblock M-var: Decoupled scale-wise autoregressive modeling for high-quality image generation.
\newblock {\em arXiv preprint arXiv:2411.10433}, 2024.

\bibitem{ren2024ConsistI2V}
Weiming Ren, Huan Yang, Ge~Zhang, Cong Wei, Xinrun Du, Wenhao Huang, and Wenhu Chen.
\newblock Consisti2v: Enhancing visual consistency for image-to-video generation.
\newblock {\em arXiv preprint arXiv:2402.04324}, 2024.

\bibitem{rombach2022stable_diffusion}
Robin Rombach, Andreas Blattmann, Dominik Lorenz, Patrick Esser, and Bj{\"o}rn Ommer.
\newblock High-resolution image synthesis with latent diffusion models.
\newblock In {\em Proceedings of the IEEE/CVF Conference on Computer Vision and Pattern Recognition}, pages 10684--10695, 2022.

\bibitem{MAGI-1}
Sand-AI.
\newblock Magi-1: Autoregressive video generation at scale, 2025.

\bibitem{shi2024IBQ}
Fengyuan Shi, Zhuoyan Luo, Yixiao Ge, Yujiu Yang, Ying Shan, and Limin Wang.
\newblock Taming scalable visual tokenizer for autoregressive image generation.
\newblock {\em arXiv preprint arXiv:2412.02692}, 2024.

\bibitem{shi2025Muddit}
Qingyu Shi, Jinbin Bai, Zhuoran Zhao, Wenhao Chai, Kaidong Yu, Jianzong Wu, Shuangyong Song, Yunhai Tong, Xiangtai Li, Xuelong Li, et~al.
\newblock Muddit: Liberating generation beyond text-to-image with a unified discrete diffusion model.
\newblock {\em arXiv preprint arXiv:2505.23606}, 2025.

\bibitem{song2025history-guided}
Kiwhan Song, Boyuan Chen, Max Simchowitz, Yilun Du, Russ Tedrake, and Vincent Sitzmann.
\newblock History-guided video diffusion.
\newblock {\em arXiv preprint arXiv:2502.06764}, 2025.

\bibitem{su2024roformer}
Jianlin Su, Murtadha Ahmed, Yu~Lu, Shengfeng Pan, Wen Bo, and Yunfeng Liu.
\newblock Roformer: Enhanced transformer with rotary position embedding.
\newblock {\em Neurocomputing}, 568:127063, 2024.

\bibitem{sun2024LlamaGen}
Peize Sun, Yi~Jiang, Shoufa Chen, Shilong Zhang, Bingyue Peng, Ping Luo, and Zehuan Yuan.
\newblock Autoregressive model beats diffusion: Llama for scalable image generation.
\newblock {\em arXiv preprint arXiv:2406.06525}, 2024.

\bibitem{swerdlow2025UniDisc}
Alexander Swerdlow, Mihir Prabhudesai, Siddharth Gandhi, Deepak Pathak, and Katerina Fragkiadaki.
\newblock Unified multimodal discrete diffusion.
\newblock {\em arXiv preprint arXiv:2503.20853}, 2025.

\bibitem{tang2024vidtok}
Anni Tang, Tianyu He, Junliang Guo, Xinle Cheng, Li~Song, and Jiang Bian.
\newblock Vidtok: A versatile and open-source video tokenizer.
\newblock {\em arXiv preprint arXiv:2412.13061}, 2024.

\bibitem{tang2025U-RoPE}
Feilong Tang, Xiang An, Haolin Yang, Yin Xie, Kaicheng Yang, Ming Hu, Zheng Cheng, Xingyu Zhou, Zimin Ran, Imran Razzak, et~al.
\newblock Univit: Unifying image and video understanding in one vision encoder.
\newblock In {\em The Thirty-ninth Annual Conference on Neural Information Processing Systems}.

\bibitem{team2024Chameleon}
Chameleon Team.
\newblock Chameleon: Mixed-modal early-fusion foundation models.
\newblock {\em arXiv preprint arXiv:2405.09818}, 2024.

\bibitem{wan2.1}
Wan Team.
\newblock Wan: Open and advanced large-scale video generative models.
\newblock 2025.

\bibitem{tian2024VAR}
Keyu Tian, Yi~Jiang, Zehuan Yuan, Bingyue Peng, and Liwei Wang.
\newblock Visual autoregressive modeling: Scalable image generation via next-scale prediction.
\newblock {\em arXiv preprint arXiv:2404.02905}, 2024.

\bibitem{tong2024metamorph}
Shengbang Tong, David Fan, Jiachen Zhu, Yunyang Xiong, Xinlei Chen, Koustuv Sinha, Michael Rabbat, Yann LeCun, Saining Xie, and Zhuang Liu.
\newblock Metamorph: Multimodal understanding and generation via instruction tuning.
\newblock {\em arXiv preprint arXiv:2412.14164}, 2024.

\bibitem{tong2022videomae}
Zhan Tong, Yibing Song, Jue Wang, and Limin Wang.
\newblock Videomae: Masked autoencoders are data-efficient learners for self-supervised video pre-training.
\newblock {\em Advances in neural information processing systems}, 35:10078--10093, 2022.

\bibitem{touvron2023llama}
Hugo Touvron, Thibaut Lavril, Gautier Izacard, Xavier Martinet, Marie-Anne Lachaux, Timothée Lacroix, Baptiste Rozière, Naman Goyal, Eric Hambro, Faisal Azhar, Aurelien Rodriguez, Armand Joulin, Edouard Grave, and Guillaume Lample.
\newblock Llama: Open and efficient foundation language models, 2023.

\bibitem{touvron2023llama2}
Hugo Touvron, Louis Martin, Kevin Stone, Peter Albert, Amjad Almahairi, Yasmine Babaei, Nikolay Bashlykov, Soumya Batra, Prajjwal Bhargava, Shruti Bhosale, et~al.
\newblock Llama 2: Open foundation and fine-tuned chat models.
\newblock {\em arXiv preprint arXiv:2307.09288}, 2023.

\bibitem{villegas2022phenaki}
Ruben Villegas, Mohammad Babaeizadeh, Pieter-Jan Kindermans, Hernan Moraldo, Han Zhang, Mohammad~Taghi Saffar, Santiago Castro, Julius Kunze, and Dumitru Erhan.
\newblock Phenaki: Variable length video generation from open domain textual descriptions.
\newblock In {\em International Conference on Learning Representations}, 2022.

\bibitem{wang2023modelscopeT2V}
Jiuniu Wang, Hangjie Yuan, Dayou Chen, Yingya Zhang, Xiang Wang, and Shiwei Zhang.
\newblock Modelscope text-to-video technical report.
\newblock {\em arXiv preprint arXiv:2308.06571}, 2023.

\bibitem{wang2024omnitokenizer}
Junke Wang, Yi~Jiang, Zehuan Yuan, Binyue Peng, Zuxuan Wu, and Yu-Gang Jiang.
\newblock Omnitokenizer: A joint image-video tokenizer for visual generation.
\newblock {\em arXiv preprint arXiv:2406.09399}, 2024.

\bibitem{wang2025SimpleAR}
Junke Wang, Zhi Tian, Xun Wang, Xinyu Zhang, Weilin Huang, Zuxuan Wu, and Yu-Gang Jiang.
\newblock Simplear: Pushing the frontier of autoregressive visual generation through pretraining, sft, and rl.
\newblock {\em arXiv preprint arXiv:2504.11455}, 2025.

\bibitem{wang2024qwen2-vl}
Peng Wang, Shuai Bai, Sinan Tan, Shijie Wang, Zhihao Fan, Jinze Bai, Keqin Chen, Xuejing Liu, Jialin Wang, Wenbin Ge, et~al.
\newblock Qwen2-vl: Enhancing vision-language model's perception of the world at any resolution.
\newblock {\em arXiv preprint arXiv:2409.12191}, 2024.

\bibitem{wang2024emu3}
Xinlong Wang, Xiaosong Zhang, Zhengxiong Luo, Quan Sun, Yufeng Cui, Jinsheng Wang, Fan Zhang, Yueze Wang, Zhen Li, Qiying Yu, et~al.
\newblock Emu3: Next-token prediction is all you need.
\newblock {\em arXiv preprint arXiv:2409.18869}, 2024.

\bibitem{wang2023lavie}
Yaohui Wang, Xinyuan Chen, Xin Ma, Shangchen Zhou, Ziqi Huang, Yi~Wang, Ceyuan Yang, Yinan He, Jiashuo Yu, Peiqing Yang, et~al.
\newblock Lavie: High-quality video generation with cascaded latent diffusion models.
\newblock {\em arXiv preprint arXiv:2309.15103}, 2023.

\bibitem{wang2024loong}
Yuqing Wang, Tianwei Xiong, Daquan Zhou, Zhijie Lin, Yang Zhao, Bingyi Kang, Jiashi Feng, and Xihui Liu.
\newblock Loong: Generating minute-level long videos with autoregressive language models.
\newblock {\em arXiv preprint arXiv:2410.02757}, 2024.

\bibitem{wei2025VideoRoPE}
Xilin Wei, Xiaoran Liu, Yuhang Zang, Xiaoyi Dong, Pan Zhang, Yuhang Cao, Jian Tong, Haodong Duan, Qipeng Guo, Jiaqi Wang, et~al.
\newblock Videorope: What makes for good video rotary position embedding?
\newblock {\em arXiv preprint arXiv:2502.05173}, 2025.

\bibitem{SANA-1.5}
Enze Xie, Junsong Chen, Yuyang Zhao, Jincheng YU, Ligeng Zhu, Yujun Lin, Zhekai Zhang, Muyang Li, Junyu Chen, Han Cai, et~al.
\newblock Sana 1.5: Efficient scaling of training-time and inference-time compute in linear diffusion transformer.
\newblock In {\em Forty-second International Conference on Machine Learning}.

\bibitem{xie2024show-o}
Jinheng Xie, Weijia Mao, Zechen Bai, David~Junhao Zhang, Weihao Wang, Kevin~Qinghong Lin, Yuchao Gu, Zhijie Chen, Zhenheng Yang, and Mike~Zheng Shou.
\newblock Show-o: One single transformer to unify multimodal understanding and generation.
\newblock {\em arXiv preprint arXiv:2408.12528}, 2024.

\bibitem{xie2025show-o2}
Jinheng Xie, Zhenheng Yang, and Mike~Zheng Shou.
\newblock Show-o2: Improved native unified multimodal models.
\newblock {\em arXiv preprint arXiv:2506.15564}, 2025.

\bibitem{xin2025lumina-mGPT-2}
Yi~Xin, Juncheng Yan, Qi~Qin, Zhen Li, Dongyang Liu, Shicheng Li, Victor Shea-Jay Huang, Yupeng Zhou, Renrui Zhang, Le~Zhuo, et~al.
\newblock Lumina-mgpt 2.0: Stand-alone autoregressive image modeling.
\newblock {\em arXiv preprint arXiv:2507.17801}, 2025.

\bibitem{yang2024qwen2.5}
An~Yang, Baosong Yang, Beichen Zhang, Binyuan Hui, Bo~Zheng, Bowen Yu, Chengyuan Li, Dayiheng Liu, Fei Huang, Haoran Wei, et~al.
\newblock Qwen2. 5 technical report.
\newblock {\em arXiv preprint arXiv:2412.15115}, 2024.

\bibitem{yang2025MMaDA}
Ling Yang, Ye~Tian, Bowen Li, Xinchen Zhang, Ke~Shen, Yunhai Tong, and Mengdi Wang.
\newblock Mmada: Multimodal large diffusion language models.
\newblock {\em arXiv preprint arXiv:2505.15809}, 2025.

\bibitem{yang2024cogvideox}
Zhuoyi Yang, Jiayan Teng, Wendi Zheng, Ming Ding, Shiyu Huang, Jiazheng Xu, Yuanming Yang, Wenyi Hong, Xiaohan Zhang, Guanyu Feng, et~al.
\newblock Cogvideox: Text-to-video diffusion models with an expert transformer.
\newblock {\em arXiv preprint arXiv:2408.06072}, 2024.

\bibitem{yin2024CausVid}
Tianwei Yin, Qiang Zhang, Richard Zhang, William~T Freeman, Fredo Durand, Eli Shechtman, and Xun Huang.
\newblock From slow bidirectional to fast autoregressive video diffusion models.
\newblock {\em arXiv preprint arXiv:2412.07772}, 2, 2024.

\bibitem{yu2025videomar}
Hu~Yu, Biao Gong, Hangjie Yuan, DanDan Zheng, Weilong Chai, Jingdong Chen, Kecheng Zheng, and Feng Zhao.
\newblock Videomar: Autoregressive video generation with continuous tokens.
\newblock {\em Advances in neural information processing systems}, 2025.

\bibitem{yu2025FAR}
Hu~Yu, Hao Luo, Hangjie Yuan, Yu~Rong, and Feng Zhao.
\newblock Frequency autoregressive image generation with continuous tokens.
\newblock {\em arXiv preprint arXiv:2503.05305}, 2025.

\bibitem{Parti}
Jiahui Yu, Yuanzhong Xu, Jing~Yu Koh, Thang Luong, Gunjan Baid, Zirui Wang, Vijay Vasudevan, Alexander Ku, Yinfei Yang, Burcu~Karagol Ayan, et~al.
\newblock Scaling autoregressive models for content-rich text-to-image generation.
\newblock {\em arXiv preprint arXiv:2206.10789}, 2(3):5, 2022.

\bibitem{yu2023MAGVIT}
Lijun Yu, Yong Cheng, Kihyuk Sohn, Jos{\'e} Lezama, Han Zhang, Huiwen Chang, Alexander~G Hauptmann, Ming-Hsuan Yang, Yuan Hao, Irfan Essa, et~al.
\newblock Magvit: Masked generative video transformer.
\newblock In {\em Proceedings of the IEEE/CVF Conference on Computer Vision and Pattern Recognition}, pages 10459--10469, 2023.

\bibitem{yu2023MAGVIT-v2}
Lijun Yu, Jos{\'e} Lezama, Nitesh~B Gundavarapu, Luca Versari, Kihyuk Sohn, David Minnen, Yong Cheng, Vighnesh Birodkar, Agrim Gupta, Xiuye Gu, et~al.
\newblock Language model beats diffusion--tokenizer is key to visual generation.
\newblock {\em arXiv preprint arXiv:2310.05737}, 2023.

\bibitem{yu2024RAR}
Qihang Yu, Ju~He, Xueqing Deng, Xiaohui Shen, and Liang-Chieh Chen.
\newblock Randomized autoregressive visual generation.
\newblock {\em arXiv preprint arXiv:2411.00776}, 2024.

\bibitem{yuan2024instructvideo}
Hangjie Yuan, Shiwei Zhang, Xiang Wang, Yujie Wei, Tao Feng, Yining Pan, Yingya Zhang, Ziwei Liu, Samuel Albanie, and Dong Ni.
\newblock Instructvideo: Instructing video diffusion models with human feedback.
\newblock In {\em Proceedings of the IEEE/CVF Conference on Computer Vision and Pattern Recognition}, pages 6463--6474, 2024.

\bibitem{zhang2025FramePack}
Lvmin Zhang and Maneesh Agrawala.
\newblock Packing input frame context in next-frame prediction models for video generation.
\newblock {\em arXiv preprint arXiv:2504.12626}, 2025.

\bibitem{zhang2022gddim}
Qinsheng Zhang, Molei Tao, and Yongxin Chen.
\newblock gddim: Generalized denoising diffusion implicit models.
\newblock {\em arXiv preprint arXiv:2206.05564}, 2022.

\bibitem{zhang2023i2vgen-xl}
Shiwei Zhang, Jiayu Wang, Yingya Zhang, Kang Zhao, Hangjie Yuan, Zhiwu Qin, Xiang Wang, Deli Zhao, and Jingren Zhou.
\newblock I2vgen-xl: High-quality image-to-video synthesis via cascaded diffusion models.
\newblock {\em arXiv preprint arXiv:2311.04145}, 2023.

\bibitem{zhou2024transfusion}
Chunting Zhou, Lili Yu, Arun Babu, Kushal Tirumala, Michihiro Yasunaga, Leonid Shamis, Jacob Kahn, Xuezhe Ma, Luke Zettlemoyer, and Omer Levy.
\newblock Transfusion: Predict the next token and diffuse images with one multi-modal model.
\newblock {\em arXiv preprint arXiv:2408.11039}, 2024.

\bibitem{zhou2025MAGI}
Deyu Zhou, Quan Sun, Yuang Peng, Kun Yan, Runpei Dong, Duomin Wang, Zheng Ge, Nan Duan, Xiangyu Zhang, Lionel~M Ni, et~al.
\newblock Taming teacher forcing for masked autoregressive video generation.
\newblock {\em arXiv preprint arXiv:2501.12389}, 2025.

\bibitem{zhuang2025vargpt}
Xianwei Zhuang, Yuxin Xie, Yufan Deng, Liming Liang, Jinghan Ru, Yuguo Yin, and Yuexian Zou.
\newblock Vargpt: Unified understanding and generation in a visual autoregressive multimodal large language model.
\newblock {\em arXiv preprint arXiv:2501.12327}, 2025.

\end{thebibliography}
\bibliographystyle{plain}

\clearpage
\appendix

\counterwithin*{figure}{section}
\counterwithin*{table}{section}
\renewcommand{\thefigure}{\thesection\arabic{figure}}
\renewcommand{\thetable}{\thesection\arabic{table}}

\section*{Appendix}

In this Appendix, we provide additional content organized as follows:
\begin{itemize}
    \item \cref{app:more_related_work} presents more related work and discussions:
    \begin{itemize}
        \item More related work on autoregressive image generation.
        \item The reasons for choosing and retaining the LLM architecture.
        \item Discussion on other LLM-based visual generation architectures.
        \item Discussion on AR-DF’s relatedness to Diffusion Forcing and FramePack. 
    \end{itemize}

    \item \cref{app:detailed_RoPE_Preliminary} presents a detailed version introducing vanilla 3D RoPE.

    \item \cref{app:detailed_inference_algorithm} presents a detailed version of the AR-DF inference algorithm.
    
    \item \cref{app:more_architectural_implementation_details} presents more architectural and implementation details. 
    \begin{itemize}
        \item Model architectural details and more training details.
        \item Detailed introduction of the token sequence formatting.
        \item GPU memory friendly implementation.
    \end{itemize}

    % \item \cref{app:more_analysis_and_ablations} provides 1) ablation studies on the effect of the number of meta MM-RoPEs and the effect of the scaling factors in MM-RoPE (\cref{app:more_ablations}), 2) \hangjie{Add new experiments} and 3) more visualizations to demonstrate image and video generation capabilities (\cref{app:more_visualizations}). \hangjie{Add more.}

    \item \cref{app:more_analysis_and_ablations} presents more analysis and ablation studies.
    \begin{itemize}
        \item More ablation studies, including
        \begin{itemize}
            \item The effect of the number of meta MM-RoPEs in MM-RoPE.
            \item The effect of the scaling factors in MM-RoPE.
        \end{itemize}
        
        \item More analysis, including 
        \begin{itemize}
            \item \method's robustness to aspect ratios.
            \item \method's comparison with other RoPE designs.
            \item \method's training resource comparison.
        \end{itemize}

        \item More visualizations, including
        \begin{itemize}
            \item Qualitative visual comparison on text-to-image generation.
            \item More text-to-image visualizations.
            \item More image-to-video and text-to-video visualizations.
        \end{itemize}
    \end{itemize}
    
    \item \cref{app:discussions} presents more discussions.
    \begin{itemize}
        \item A road map to unified models for understanding and generation.
        \item Limitations and future work.
        \item Potential societal impact and safeguards.
    \end{itemize}
    
    % \item \cref{app:discussions} presents discussions covering 1) limitations and future work (\cref{app:limitations_future_work}) and 2) potential societal impact and safeguards (\cref{app:societal_impact_safeguards}). 

    \item \cref{app:LLM_usage} presents a disclosure of LLM usage in this paper. 
\end{itemize}

% \hangjie{Append videos in the supp for reference.}

% \hangjie{Append detailed prompts in the supp for reference.}

% In this appendix we include supplementary material organized as follows:

% \begin{itemize}
% \item \cref{app\:detailed\_inference\_algorithm};–,A complete, step-by-step version of the AR-DF inference procedure.

% ```
% \item \cref{app:more_architectural_implementation_details}\;–\,Additional architectural and implementation details, including the end-to-end training pipeline.

% \item \cref{app:more_analysis_and_ablations}\;–\,Extended experiments:  
%       \begin{enumerate*}[label=(\roman*)]
%           \item ablations on the number of meta MM-RoPE groups and on the 3-D scaling factors (\cref{app:more_ablations}), and  
%           \item further qualitative results illustrating image- and video-generation capabilities (\cref{app:more_visualizations}).
%       \end{enumerate*}

% \item \cref{app:discussions}\;–\,Broader discussion covering  
%       \begin{enumerate*}[label=(\roman*)]
%           \item current limitations and avenues for future work (\cref{app:limitations_future_work}), and  
%           \item potential societal impacts together with the safeguards we employ (\cref{app:societal_impact_safeguards}).
%       \end{enumerate*}
% ```

% \end{itemize}

\section{More Related Work and Discussions} \label{app:more_related_work}
As a supplement to the main paper, we provide 
\textbf{1)} more related work on autoregressive image generation, 
\textbf{2)} the reasons for choosing LLMs, 
\textbf{3)} the comparison with LLM-based visual generation models and 
\textbf{4)} AR-DF’s relatedness to Diffusion Forcing and FramePack.

\noindent\textbf{Autoregressive image generation.}
% Due to the surge and success of large language models, the research focus of visual generation has shifted from diffusion paradigms~\cite{rombach2022stable_diffusion,ho2020denoising_DDPM,zhang2022gddim,2024SD3} to an autoregressive one~\cite{sun2024LlamaGen,tian2024VAR,yu2025FAR,huang2025NFIG,ren2025xAR,zhou2025MAGI,yu2024RAR,he2025NAR,ma2025token-shuffle,wang2025SimpleAR,ren2024M-VAR} due to its potential of integrating into unified models~\cite{team2024Chameleon,zhuang2025vargpt}.
Spurred by the progress of LLMs, visual generation research is rapidly shifting from diffusion paradigms~\cite{rombach2022stable_diffusion,ho2020denoising_DDPM,zhang2022gddim,2024SD3} to an autoregressive one~\cite{sun2024LlamaGen,tian2024VAR,yu2025FAR,huang2025NFIG,ren2025xAR,zhou2025MAGI,yu2024RAR,he2025NAR,ma2025token-shuffle,wang2025SimpleAR,ren2024M-VAR}, due to its potential of being integrated into unified models~\cite{team2024Chameleon,zhuang2025vargpt} with minimal modification.
% Preliminary research such as Parti~\cite{Parti}, DALL-E~\cite{Dalle}, MaskGiT~\cite{chang2022maskgit} and LLamaGen~\cite{sun2024LlamaGen}, have demonstrated the generation effectiveness of aligning with language models using discrete tokenizers.
% MAR~\cite{li2024MAR} introduces the diffusion loss to autoregressive image generation, enabling the use of continuous tokens.
% VAR~\cite{tian2024VAR} and FAR~\cite{yu2025FAR} introduce new autoregressive paradigms, \textit{i.e.}, next-scale prediction and next-frequenct, to  the generation effectiveness.
Preliminary research, such as Parti~\cite{Parti}, DALL-E~\cite{Dalle}, MaskGiT~\cite{chang2022MaskGiT} and LLamaGen~\cite{sun2024LlamaGen}, have demonstrated the efficacy of autoregressive generation using discrete tokenizers.
MAR~\cite{li2024MAR} incorporates a diffusion-style objective into autoregressive training, thereby accommodating continuous tokens.
VAR~\cite{tian2024VAR} and FAR~\cite{yu2025FAR} reformulate the learning target as next-scale and next-frequency prediction, respectively, improving generation fidelity.
Unlike them, we dive into the design of the RoPE technique in visual generation.
% In contrast to these directions, we revisit a foundational component—rotary positional embedding (RoPE)—and systematically analyze how its design choices influence autoregressive video generation in our Lumos-1 framework.

\noindent\textbf{The reasons for choosing and retaining the LLM architecture.}
One significant group of unified models, like Chameleon~\cite{team2024Chameleon} and EMU3~\cite{wang2024emu3}, adopts the original LLM architecture to unify understanding and generation tasks. 
Some hybrid architectures combining an autoregressive transformer and a diffusion head, like BLIP3o-Next~\cite{chen2025blip3o}, also rely on the autoregressive visual generation capabilities.
Although the autoregressive LLM transformers are very capable language models since they are designed for language tasks, the visual generation results are not satisfactory due to the lack of research exploration. 
We align our model architecture with the LLM so that the insights obtained and techniques designed (e.g., MM-RoPE and AR-DF) in this paper could transfer smoothly to the unified models to come (but we also want to mention that techniques like MM-RoPE can be transferred to other types of visual generation models). This architectural alignment also means that \method is not just a specialized generator; the addition of understanding-oriented data to its pre-training can unlock both generation and comprehension capabilities, which is a promising future work.

% \noindent\textbf{Discussion on other LLM-based visual generation architectures.}
% % Unified models LLMs
% We briefly compare our model with LlamaGen and DALLE in~\cref{tab:comparison_with_other_LLM_AR_models} to clarify the key distinctions. 
% First of all, both of them 
% \textbf{1)} relied on the slow next-token prediction for generating visual content and
% \textbf{2)} failed to support video generation, which we aim to address. Apart from these two, regarding DALLE, although it retains an LLM, some key designs like the core position embedding is different from the one used in contemporary LLMs (RoPE is published after DALLE), leaving the topic worth exploring. Regarding LlamaGen, it retains an LLM for the visual generation part while still utilizing the 2D RoPE and relying on externally trained text encoders, departing from a fully unified design. 

% \textcolor{Watermelon_Red}{
\noindent\textbf{Discussion on other LLM-based unified visual generation architectures.}
We briefly compare our model with representative LLM-based unified generative models, including DALL·E~\cite{Dalle}, LlamaGen~\cite{sun2024LlamaGen}, Loong~\cite{wang2024loong} and Lumina-mGPT~\cite{liu2024Lumina-mGPT} in~\cref{tab:comparison_with_other_LLM_AR_models} to clarify the key distinctions. 
This comparison highlights three key advantages of Lumos-1's design:
\textbf{1)} Truly unified architecture: Unlike hybrid models like LlamaGen which stitch together a T5 encoder and an LLM (for visual generation), Lumos-1 uses a single, end-to-end LLM backbone. This is a simpler and more elegant design.
\textbf{2)} Advanced positional encoding (MM-RoPE): Lumos-1 is the only model that incorporates a native 3D-aware RoPE (MM-RoPE) that is also compatible with standard text RoPE used in LLMs. This allows for principled spatiotemporal modeling for visual data, a capability absent in models using 1D, 2D, or non-RoPE embeddings.
\textbf{3)} Efficient prediction paradigm: Crucially, Lumos-1 is the only model in this comparison to leverage fast, parallelizable mask-based prediction. Other models are constrained by the slow, sequential nature of next-token prediction, which becomes prohibitively inefficient for videos.
These targeted architectural choices allow Lumos-1 to function as a fast and versatile image-video generator while being consistent with the core LLM structure.
% }

\begin{table}[!h]
\centering
\small
\caption{
% \textcolor{Watermelon_Red}{
\small
\textbf{Comparison of different LLM-based visual generation models.} The table highlights their architecture, position embedding techniques, prediction paradigms, generated media types, and performance on the GenEval benchmark.
% }
}
\label{tab:model_comparison}
\resizebox{\textwidth}{!}{
\begin{tabular}{@{}llllll@{}}
\toprule
\textbf{Model} & \textbf{Architecture} & \textbf{Position Embedding} & \textbf{Prediction Paradigm} & \textbf{Generated Media} & \textbf{GenEval} \\
\midrule
DALL·E & LLM & $\sin/\cos$ PE & Next-token prediction (Slow) & Image & -- \\
LLamaGen & T5 + LLM (2.9B + 0.8B) & Na\"ive 2D RoPE & Next-token prediction (Slow) & Image & 0.32 \\
Loong    & LLM      & 1D RoPE & Next-token prediction (Slow)       & Image + Video & -- \\
Lumina-mGPT & LLM  (7B)  & 1D RoPE & Next-token prediction (Slow)       & Image & 0.56 \\
\method & LLM (1.5B) & MM-RoPE (distributed and scaled 3D RoPE) & Mask prediction (Fast) & Image + Video & 0.725 \\
\bottomrule
\end{tabular}
}
\label{tab:comparison_with_other_LLM_AR_models}
\end{table}

\noindent\textbf{Discussion on AR-DF's relatedness to Diffusion Forcing and FramePack.}
% The drifting-forgetting dilemma is a common challenge for autoregressive models. 
For Diffusion Forcing~\cite{chen2024diffusion_forcing} and AR-DF, they are related since both of them are methods to overcome the fundamental \textit{drifting problem} in video generation: A small error made early on gets passed down and magnified over time, causing the final video to "drift" far away from the intended or realistic result.
The temporal dependency of the video data, to some extent, leads to the drifting problem since later frames' predictions rely on previous frames. 
We identify that this is exacerbated by frame-wise loss imbalance since the model tends to optimize those easier tasks, encouraging later frames to rely on "copying" previous frames. 
The temporal tube masks in AR-DF or the per-frame independent noise in Diffusion Forcing Transformer~\cite{song2025history-guided} both aim to ameliorate this issue by reducing the reliance of later frames on previous frames. 
The inference-time masks in \method also mitigate error accumulation by dropping partially generated context.
Our temporal tube masking is particularly suited for video data, as it directly targets the spatial information redundancy between frames. 
This can be revealed by the experiments in~\cref{tab:diff_training_methods}, where we compare our model with a variant trained by following Diffusion Forcing Transformer.
We observe that they have similar performance on image generation, while \method performs better on video generation.
For FramePack~\cite{zhang2025FramePack}, it compresses the history in a context with a fixed length. 
Due to its strategic method for history compression, it focuses more on overcoming the \textit{forgetting problem} (\textit{i.e.}, later frames forget history context), which is not the focus of \method. 
It also proposes anti-drifting sampling to get rid of drifting. However, it relies on access to future frames.

% In addition to the above arguments, we also want to emphasize the architectural differences between these models. Therefore, choosing one of them depends on the specific goal. If we want to build one unified model capable of generating videos, Lumos-1 might be a suitable choice; if we target less forgetting, then we should choose FramePack; if we cannot get access to any future information, then we should choose Lumos-1 or Diffusion Forcing Transformer.

\section{Detailed Preliminary of 3D RoPE} \label{app:detailed_RoPE_Preliminary}

One de facto design of contemporary LLM is RoPE~\cite{su2024roformer}, whose overall aim is to encode the absolute position with a rotation matrix while incorporating the explicit relative position dependency in the attention mechanism.
This can be formulated as:
\begin{equation}\label{eqn:attn_calculation}
    \langle f_q(\bm{x}_m, m), f_k(\bm{x}_n, n) \rangle = 
    g(\bm{x}_m, \bm{x}_n, m-n)
\end{equation}
where $f_q(\bm{x}_m, m)$ encodes the position $m$ for the embedding $\bm{x}_m$ to obtain the query feature ($f_k(\bm{x}_n, n)$ is analogously defined);
$g(\bm{x}_m, \bm{x}_n, m-n)$ is a function that defines the inner product between the query and key vectors that explicitly encodes the relative position $m-n$.
The resultant form of function $f_{\{q,k\}}$ can be formulated as:
\begin{equation}
    f_{\{q,k\}}(\bm{x}_m, m) = \bm{R}^{d}_{\Theta, m} \bm{W}_{q,k} \bm{x}_m
\end{equation}
where $\bm{W}_{q,k}$ is the projection matrix; $\bm{R}^{d}_{\Theta, m}$ is the rotary matrix with pre-defined parameters $\Theta = \{\theta_i=\beta^{-2(i-1)/d}, i=[1,2,...,d/2]\}$ with $d$ acting as the dimension of the embeddings and $\beta$ acting as the base frequency.
In such a formulation, the attention calculation can be rewritten as:
\begin{equation}
    f_q(\bm{x}_m, m)^{\rm T} f_k(\bm{x}_n, n) = \bm{x}_m^{\rm T} \bm{W}_q^{\rm T} \bm{R}^{d}_{\Theta, \tau} \bm{W}_k \bm{x}_n, \quad \tau = n - m
\end{equation}
where the detailed formulation of $\bm{R}^{d}_{\Theta, \tau}$ can be formulated using a base rotary matrix $R_{\theta, \tau}$, with $\theta$ as the frequency and $\tau$ as the relative position:
\begin{equation}
    {%\scriptsize
    \bm{R}^{d}_{\Theta, \tau} = 
    \begin{bmatrix}
        R_{\theta_{1}, \tau} & 0 & \cdots & 0 \\
        0 & R_{\theta_{2}, \tau} & \cdots & 0 \\
        \vdots &\vdots & \ddots & 0 \\
        0 & 0 & \cdots & R_{\theta_{d/2}, \tau} \\
    \end{bmatrix}, \quad 
    R_{\theta, \tau} = 
    \begin{bmatrix}
        \cos \tau \theta & -\sin \tau \theta \\
        \sin \tau \theta & \cos \tau \theta
    \end{bmatrix}
    }
\end{equation}
However, the application of the original RoPE to modeling visual data remains suboptimal considering the spatiotemporal correlation of visual tokens.
One popular type of generative models, diffusion models~\cite{ho2022video_diffusion_models,ho2020denoising_DDPM}, improved upon this technique by proposing 3D RoPE that jointly injects spatiotemporal latent coordinates during attention calculation and demonstrates its effectiveness~\cite{hong2022cogvideo,kong2024hunyuanvideo,wan2.1}.
If we slightly abuse the annotation by denoting $\bm{x}_m^{\rm T} \bm{W}_q^{\rm T}$ and $\bm{W}_k \bm{x}_n$ as $\bm{X}_m^{\rm T}$ and $\bm{X}_n$, we can write the attention calculation in~\cref{eqn:attn_calculation} based on 3D RoPE as:
\begin{equation}
    \resizebox{\textwidth}{!}{$
    \begin{aligned}
    % f_q(\bm{x}_m, m)^{\rm T} f_k(\bm{x}_n, n) =
    \bm{X}_{m, t_s:t_e}^{\rm T}
    \begin{bmatrix}
        R_{\theta_{t_s + 1}, \tau_t} & \cdots & 0 \\
        \vdots                   & \ddots & \vdots \\
        0                        & \cdots & R_{\theta_{t_e}, \tau_t} \\
    \end{bmatrix} 
    \bm{X}_{n, t_s:t_e} 
    +
    \bm{X}_{m, h_s:h_e}^{\rm T}
    \begin{bmatrix}
        R_{\theta_{h_s + 1}, \tau_h} & \cdots & 0 \\
        \vdots                   & \ddots & \vdots \\
        0                        & \cdots & R_{\theta_{h_e}, \tau_h} \\
    \end{bmatrix} 
    \bm{X}_{n, h_s:h_e}
    % &+ \\
    +
    \bm{X}_{m, w_s:w_e}^{\rm T}
    \begin{bmatrix}
        R_{\theta_{w_s + 1}, \tau_w} & \cdots & 0 \\
        \vdots                   & \ddots & \vdots \\
        0                        & \cdots & R_{\theta_{w_e}, \tau_w} \\
    \end{bmatrix}
    \bm{X}_{n, w_s:w_e}
    \end{aligned}$}
\end{equation}
where $\{t_s,t_e\} = \{0,\frac{2}{16}d\}$, $\{h_s,h_e\} = \{\frac{2}{16}d, \frac{5}{16}d\}$ and $\{w_s,w_e\} = \{\frac{5}{16}d, \frac{1}{2} d\}$ denote the start and end dimension index for encoding temporal, height and width relative position; 
$\bm{X}_{m, t_s:t_e}^{\rm T}$ denotes the submatrix extracted from $\bm{X}_{m}^{\rm T}$ using row indices $[2t_s, 2t_e)$;
other matrices are similarly defined.

\section{Detailed AR-DF Inference Algorithm} \label{app:detailed_inference_algorithm}
In this subsection, we present the detailed version of the AR-DF inference algorithm.
After the text prompt $\bm{X}_p$ is encoded once and its KV pairs are cached (Lines\,1–3), we iterate over the $T$ latent frames (Line\,6).
Each frame is generated by running $N_{steps}$ iterations (Lines\,9–22).
During the frame generation process, the model predicts a token distribution, samples tokens only at positions that are still masked, gathers their confidences, and sets the scores of already-revealed tokens to $+\infty$.
After adding Gumbel noise, we re-mask the lowest-confidence tokens.
As the generation proceeds, fewer tokens are re-masked as the running steps grow and the frame becomes sharper and clearer.
The partially revealed frame is then added to the cache before the next frame is generated. 
Once all frames are finished, they are decoded to RGB space to form the final video.

%%%%%%%%%%%%%%%%%%%%%%%%%% V2 %%%%%%%%%%%%%%%%%%%%%%
\begin{figure}[t]
\centering
\begin{algorithm}[H]
\small
\caption{AR-DF Inference Procedure}
\label{alg:AR-DF_inference_detailed}
\textbf{Require:} text prompt $\bm{X}_p$, trained model $G_\phi$, inference mask ratio $\rho_{inf}$, number of latent frames $T$, number of generation steps $N_{steps}$, number of tokens in a latent frame $N_f$, KV cache $\mathcal{C} = \varnothing$, generated latent list $\mathcal{V} = \varnothing$
\begin{algorithmic}[1]
\State \(\textbf{Initialize text cache}\):
  \State \quad Generate text causal mask \(AttnMask^{(p)}\) for \(\bm{X}_p\)
  % \State \quad \(\mathcal{C} \leftarrow G_\phi(\widetilde{\bm{X}}, AttnMask^{(p)})\) \Comment{Store cache for the prompt}
  \State \quad \(\mathcal{C} \leftarrow G_\phi(\bm{X}_p, AttnMask^{(p)})\) \Comment{Store cache for the prompt}
\State \(\textbf{Sample cache mask}\):
  \State \quad \(\bm{M}_{inf} \sim \mathrm{Bernoulli}(1 - \rho_{inf})\)
\For{\(t = 1 \to T\)}  \Comment{Reused for all frames}
  \State Initialize all tokens in \(\bm{X}_v^{(t)}\) as \([\text{MASK}]\)
  \State Generate temporal causal mask \(AttnMask^{(t)}\) for \(\{\bm{X}_p, \bm{X}_v^{(t)}\}\)

  \For{\(n = 1 \to N_{steps}\)}
    \State $\bm{P} \leftarrow G_\phi\bigl(\bm{X}_v^{(t)}, AttnMask^{(t)}, \mathcal{C}\bigr)$ 
    \Comment{Predicted token distribution for all tokens}
    % \State Select the $(N_f - U)$ masked positions with the lowest confidence in $\bm{P}$
    % \State \(\textbf{Sample tokens:}\)
    \State \(\textit{sampled\_ids} \;\leftarrow\; \mathrm{multinomial}(\bm{P},\, 1)\)
    % \State \(\textbf{Identify lowest-confidence positions:}\)

    \State $\textit{sampled\_ids} \gets
       \mathrm{where}\!\bigl(\bm{X}^{(t)}_v=[\text{MASK}],
       \;\textit{sampled\_ids},
       \;\bm{X}^{(t)}_v\bigr)$

    \State $\bm Q \gets \mathrm{Gather}\bigl(\bm P,\; \textit{sampled\_ids}\bigr)$
       \Comment{Select confidence score of every chosen token}

    % \State $\bm Q \gets \text{SetKownInfinity} \!\bigl(\bm Q,\; \bm X^{(t)}_v != \text{[MASK]} \bigr)$  \Comment{set $+\infty$ where the token was \emph{not} masked}
    \State $\bm{Q} \gets \mathrm{SetKnownInfinity}
       \!\bigl(\bm{Q},\; \bm X^{(t)}_v \neq [\text{MASK}] \bigr)$
       \Comment{Set $+\infty$ where the token was \emph{not} masked}

    % \mathsf{MaskKnown}

    \State \textbf{Mask the lowest confident ($\bm{Q}$) tokens in $\textit{sampled\_ids}$ with randomness:}
    \State \quad Perturb $\bm{Q}$ with gumbel noise
    \State \quad {%\color{blue} 
      \(\alpha \leftarrow \cos(\frac{\pi}{2}\frac{n}{N_{steps}})\)
    }
    \Comment{Cosine schedule factor in [0,1]}
    \State \quad $U \leftarrow \text{floor}(\alpha \times N_f)$ \Comment{Number of tokens to mask in this iteration}
    % \State \quad \(\textit{low\_conf\_pos} \;\leftarrow\; \mathrm{SelectLowConfidence}\bigl(\bm{P},\, (N_f - U)\bigr)\)
    \State \quad \(\textit{low\_conf\_pos} \;\leftarrow\; \mathrm{SelectLowConfidence}\bigl(\bm{Q},\,  U\bigr)\)
    % \State \(\textbf{Replace them with [MASK]:}\)
    \State \quad \(\textit{sampled\_ids}[\textit{low\_conf\_pos}] 
       \;\leftarrow\; [\text{MASK}]\)
    % \State \(\textbf{Update tokens for this frame:}\)
    \State \quad \(\bm{X}_v^{(t)} \;\leftarrow\; \textit{sampled\_ids}\)
  \EndFor

  \State Append \(\bm{X}_v^{(t)}\) to \(\mathcal{V}\) % used for decoding
  % Mask the generated frame
  \State \textbf{Cache partial observation of the generated frame}:
    \State \quad \(\widetilde{\bm{X}}_v^{(t)} 
      = \bm{M}_{inf} \odot \bm{X}_v^{(t)}
      + (1 - \bm{M}_{inf}) \odot [\text{MASK}]\)
    \State \quad \(\mathcal{C} \leftarrow G_\phi(\widetilde{\bm{X}}_v^{(t)}, AttnMask^{(t)}, \mathcal{C})\) \Comment{Store cache for the \(t\)-th frame} 
\EndFor
\State Decode \(T\) latents \(\mathcal{V}\) to RGB space
\State \textbf{return} the generated video frames
\end{algorithmic}
\end{algorithm}
\vspace{-0.3cm}
\caption{
\textbf{Detailed inference algorithm with AR-DF.}
}
\vspace{-0.5cm}
% \caption{\textbf{Inference algorithm with AR-DF.}
% We first cache the text prompt in $\mathcal{C}$, then iteratively generate frame latents $\bm{X}_v^{(t)}$ in a MaskGiT-like fashion. 
% After each frame is generated, we create a partially masked version $\widetilde{\bm{X}}_v^{(t)}$ and pass it into the model to update $\mathcal{C}$, maintaining consistent context across frames.}
\end{figure}

\section{More Architectural and Implementation Details} \label{app:more_architectural_implementation_details}

\noindent\textbf{Model architectural details and more training details.}
For all of our models, we adopt the Adam~\cite{loshchilov2018adamW} optimizer to optimize the model, set the weight decay to 0.1 and set $\beta_1$ and $\beta_2$ to $0.9$ and $0.95$.
\cref{tab:architecture_training_summary} presents the architectural details of three versions of \method and their corresponding training details, including their training batch sizes and their learning rates.  
For 256p joint training, we use all 60M images and 10M videos.
For 384p joint training, we curate a higher-quality subset from 256p training data, consisting of 8M images and 0.6M videos.
% When performing joint training, we follow Movie Gen~\cite{polyak2024MovieGen} to iterate over the image batches and video batches.
Following the alternating strategy of MovieGen~\cite{polyak2024MovieGen}, we interleave image and video batches during joint training.
%%%%%%%%%%%%%%% Delete this finally. We can provide this if necessary. %%%%%%%%%%%%%%%
% We perform 256p image pre-training for 100k steps, 256p joint training for 240k steps, and 384p joint training for 90k steps.

\noindent\textbf{Detailed token sequence formatting.}
The visual tokens and text tokens are interleaved within a sequence~\cite{liu2024Lumina-mGPT}, with the text tokens specifying metadata, including the text prompt, video resolution, video fps and the number of frames in this video.
With this design, we train the model using images and videos of varying aspect ratios without resizing them.
% The details of the sequence formatting (\textit{e.g.}, how special tokens are organized) are placed in the Appendix.
The text tokens are formatted as ``Generate a video with a resolution of $<video\_resolution>$, consisting of $<video\_\#frames>$ frames at $<video\_fps>$ frames per second, according to the following prompt:\textbackslash{n} $<text\_prompt>$".
The visual tokens are extracted and organized according to the following structured format: 
``$<video\_start\_token>$, $<video\_duration\_token>$, $<video\_fps\_token>$, $<frame\_tokens>$,...,$<video\_end\_token>$".
$<frame\_tokens>$ contains tokens for each latent frame, which are formatted as:
``$<image\_start\_token>$, $<h\_grid\_token>$, $<w\_grid\_token>$, $<image\_content\_tokens>$, ..., $<image\_end\_token>$", where $<image\_content\_tokens>$ are raster-scan visual tokens with a $<new\_line\_token>$ inserted after each row of visual tokens.
To harmonize text and visual tokens, we utilize MM-RoPE, which encodes spatiotemporal coordinates for visual tokens and global positions for text.

\noindent\textbf{GPU memory friendly implementation.}
By default, we leverage flash attention~\cite{dao2023Flashattention-2} to accelerate attention calculation and reduce memory overhead during training and inference of \method.
Moreover, we observe significant amount of GPU memory consumption during the training of a model with a large codebook size.
To address it, we eliminate the use of language-related loss (\textit{i.e.}, next-token prediction on texts), reducing the final logit matrix size to match only visual tokens.
While text token embeddings, which map the text token index to token embeddings, remain trainable, this focuses the model on video generation.
This loss can be added if we target a unified model that is capable of learning within the language modality.
Finally, we observe substantially high GPU memory consumption during loss computation over 129$k$ token types, which easily leads to the out-of-memory issue.
To solve this, we utilize chunked cross-entropy loss, which maintains full softmax accuracy by upcasting and calculating softmax logits one chunk of token sequences at a time. 
This approach significantly reduces peak memory usage.
By default, we set the chunk size to 2,000.

\begin{table}[t]
\setlength{\tabcolsep}{4.5pt}
\centering
\small
\caption{
\small
\textbf{Model architectural and training details.} 
Batch sizes (*/*) are for images and videos during joint training.
Learning rates (*/*) are for 256p and 384p.
% Batch sizes (*/*) denote the batch sizes for images and videos during joint training.
% The 0.5B version is specifically for fast ablation studies.
The 0.5B version is only for fast ablation studies.
}
\resizebox{\textwidth}{!}{
\begin{tabular}{l|ccccc|ccc}
\toprule
\textbf{Model} & \textbf{\#Params} & \textbf{\#Layers} & \textbf{Hidden size} & \textbf{\#Heads} & \textbf{Head dim} & \textbf{Batch (256p)} & \textbf{Batch (384p)} & \textbf{Learning rate} \\
\midrule
\method 0.5B & 0.5B & 16 & 1024 & 16 & 64 & 896 / 128 & 192 / 32 & 5e-5 / 2.5e-5 \\
\method 1B & 1.5B & 16 & 2048 & 32 & 64 & 640 / 96 & 192 / 32 & 5e-5 / 2.5e-5 \\
\method 3B & 3.6B & 28 & 3072 & 24 & 128 & 768 / 96 & 288 / 48 & 1e-4 / 5.0e-5\\
\bottomrule
\end{tabular}}
\label{tab:architecture_training_summary}
\end{table}

\section{More Analysis and Ablation Studies} \label{app:more_analysis_and_ablations}

\begin{figure}[t]
\centering
\includegraphics[width=1.0\linewidth]{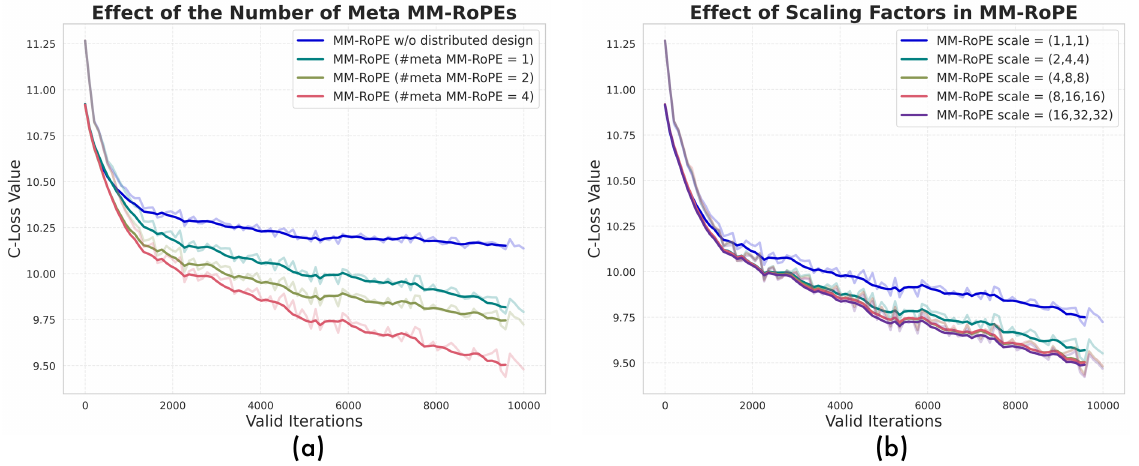}
\vspace{-.3cm}
\caption{
\small
(a) \textbf{Effect of the number of meta MM-RoPEs} in MM-RoPE;
(b) \textbf{Effect of the scaling factors} in MM-RoPE.
Both curves are plotted using the 0.5B model trained on videos.
}
% \vspace{-.33cm}
\label{fig:mmrope_ablation_meta_mmrope_scaling}
\end{figure}

% We might not need it since do not have it as well.
% \textbf{Human evaluation.}
% \textbf{Better vision-language alignment with a unified codebook.}

\subsection{More Ablation Studies} \label{app:more_ablations}
\textbf{Effect of the number of meta MM-RoPEs in MM-RoPE.}
MM-RoPE slices the embedding channels into several meta groups.
More groups mean that one kind of information (temporal, height or width) receives a wider and more comprehensive frequency spectrum for modeling.
% for modeling a certain dimension of data.
\cref{fig:mmrope_ablation_meta_mmrope_scaling}(a) plots the validation loss for a 0.5B model under four settings:
\textbf{1)} \textit{w/o distributed design}: 
This setting uses the previous design, which allocates the first $\frac{2}{8}$ channels for temporal modeling and $\frac{3}{8}$ channels for height and width modeling, respectively.
\textbf{2)} \textit{\#meta MM-RoPE=1}: 
This setting equips a meta MM-RoPE of 64 channels, while maintaining the ratios ($2:3:3$) for modeling temporal, height and width information.
This variant improves upon the previous one by enabling interleaved height and width channels, which increases the frequency spectrum range for the two spatial dimensions.
\textbf{3)} \textit{\#meta MM-RoPE=2}: 
This setting equips a meta MM-RoPE of 32 channels.
This variant improves the frequency spectrum range for temporal, height and width information compared with the previous one.
\textbf{4)} \textit{\#meta MM-RoPE=4}:
This setting is the default design that keeps the number of channels for every meta MM-RoPE minimal (16 channels).
This means that the frequency spectrum range is most comprehensive for the temporal, height, or width dimension.
These results confirm that widening the frequency spectrum for every dimension through increasing the number of meta MM-RoPEs substantially improves spatiotemporal modeling and overall training effectiveness.

% \textbf{Effect of the number of meta MM-RoPEs in MM-RoPE.}
% MM-RoPE benefits significantly from the distributed design, \textit{i.e.}, the frequency spectra for temporal, height and width are designed to be as comprehensive as possible.
% XXX presents the validation loss curves by varying the number of meta MM-RoPEs using the 0.5B model:
% 1) \textit{w/o distributed design}: We use the previous design, which allocates $\frac{2}{8}$ channels for temporal modeling and $\frac{3}{8}$ channels for height and width modeling, respectively.
% 2) \textit{\#meta MM-RoPE=1}: Every meta MM-RoPE has 64 channels. 
% This variant improves upon the previous one by enabling interleaved height and width channels. This increases the frequency spectrum range for the two spatial dimensions.
% 3) \textit{\#meta MM-RoPE=2}: Every meta MM-RoPE has 32 channels. 
% This variant improves the frequency spectrum range compared with the previous one.
% 4) \textit{\#meta MM-RoPE=4}: This is the default design by keeping the number of channels for every meta MM-RoPE minimal (16 channels), which means that the frequency spectrum range is most comprehensive for the temporal, height, or width dimension.
% We can observe from the figure that utilizing the design and making the spectrum range as comprehensive as possible significantly decrease the evaluation loss, demonstrating the effectiveness of our design.

\begin{wraptable}{r}{0.5\textwidth} 
\centering
\small
\setlength{\tabcolsep}{3pt}
\vspace{-.1cm}
\caption{\small 
\textbf{GenEval score across different aspect ratios} using 1B model.}
\vspace{-.3cm}
\resizebox{1.\linewidth}{!}{
\begin{tabular}{lccc}
\toprule
 & $\bm{448\times 256}$ ($7:4$) & $\bm{352\times 352}$ ($1:1$) & $\bm{256\times 448}$  ($4:7$) \\
\midrule
GenEval & 0.605 & 0.601 & 0.569 \\
\bottomrule
\end{tabular}}
\label{tab:aspect_ratios_results}
\vspace{-.3cm}
\end{wraptable}

\textbf{Effect of the scaling factors in MM-RoPE.}
\cref{fig:mmrope_ablation_meta_mmrope_scaling}(b) presents the validation loss curves of varying the scaling factors for modeling temporal, height and width positional information in MM-RoPE.
Two clear trends emerge:
Moving from $(1,1,1)$ → $(2,4,4)$ → $(4,8,8)$ steadily lowers the curve, but further enlarging to $(8,16,16)$ or $(16,32,32)$ yields no additional gain since the three curves almost overlap throughout training.
Therefore, a moderate scale of $(4,8,8)$ is sufficient to balance vision-language ranges and fully realize the benefit of higher-resolution RoPE, while avoiding unnecessary frequency inflation. 
We therefore adopt $(4,8,8)$ as the default scaling for MM-RoPE.

% \textbf{Effect of the scaling factors in MM-RoPE.}
% XXX presents the validation loss curves of varying the scaling factors in MM-RoPE.
% We increase the scale factors from $(1,1,1)$ to $(16,32,32)$ and we find that when increasing from $(1,1,1)$ to $(4,8,8)$, performance gradually improves. However, when the scaling factors continue to increase, the validation loss plateaus, demonstrating that $(4,8,8)$ is sufficient for modeling spatiotemporal data.
% Therefore, our default design settles on the scaling factors of $(4,8,8)$.

% \textbf{Benefits of model size scaling.}

\subsection{More Analysis} \label{app:more_analysis}

\noindent\textbf{Robustness to aspect ratios.}
Although the aspect ratio of training data is mostly $7:4$, \cref{tab:aspect_ratios_results} indicates that \method 1B adapts well to visual generation with different aspect ratios due to the unified codebook design.

\begin{wraptable}{r}{0.55\textwidth} 
% \begin{table}[h!]
\centering
\vspace{-.3cm}
\caption{
\small
% \textcolor{Watermelon_Red}{
\textbf{Design comparison with other types of RoPE}.
% }
}
% \vspace{-.2cm}
\resizebox{1.\linewidth}{!}{
\begin{tabular}{@{}lcccc@{}}
\toprule
\textbf{\shortstack{RoPE \\ Type}} & \textbf{\shortstack{Compatiable with \\ Text RoPE}} & \textbf{\shortstack{3D \\ Structure}} & \textbf{\shortstack{Comprehensive \\ Frequency Allocation}} & \textbf{\shortstack{Strategic \\ Scaling}} \\
\midrule
M-RoPE    & \ding{52}  &  \ding{52}  &  \ding{55}  &  \ding{55} \\
U-RoPE    & \ding{52}  &  \ding{52}  &  \ding{55}  &  \ding{55} \\
IL-RoPE   & \ding{55}  &  \ding{52}  &  \ding{55}  &  \ding{55} \\
VideoRoPE & \ding{52}  &  \ding{52}  &  \ding{55}  &  \ding{52} \\
HoPE      & \ding{52}  &  \ding{52}  &  \ding{55}  &  \ding{52} \\
\midrule
MM-RoPE   & \ding{52}  &  \ding{52}  &  \ding{52}  &  \ding{52} \\
\bottomrule
\end{tabular}}
\vspace{-.3cm}
\label{tab:diff_RoPE_design}
% \end{table}
\end{wraptable}

\noindent\textbf{Comparison with other RoPE designs.} 
To demonstrate the superiority of MM-RoPE, 
% \textcolor{Watermelon_Red}{
we compare it with five other RoPEs, \textit{i.e.}, M-RoPE~\cite{wang2024qwen2-vl}, VideoRoPE~\cite{wei2025VideoRoPE}, U-RoPE~\cite{tang2025U-RoPE}, IL-RoPE~\cite{liao2025Mogao} and HoPE~\cite{li2025HoPE}.
% }
% \textcolor{Watermelon_Red}{
We show their design comparison in~\cref{tab:diff_RoPE_design}.
We mainly compare these methods on four aspects: 
\textbf{1)} their compatibility with text RoPE, so that models can align with contemporary LLMs;
\textbf{2)} the inclusion of 3D information into RoPE to enable enhanced spatiotemporal modeling;
\textbf{3)} the comprehensive frequency allocation that enables more balanced local-global context modeling;
\textbf{4)} the strategic scaling of spatiotemporal indices (or their subsets) to account for different granularities of vision and text information.
Based on the above analysis, MM-RoPE holistically incorporates the four properties.
% }

\begin{wraptable}{r}{0.45\textwidth} 
% \begin{table}[h!]
\centering
\vspace{-.3cm}
\caption{
\small
% \textcolor{Watermelon_Red}{
\textbf{Performance comparison with other types of RoPE}.
Results are measured on GenEval, Vbench-Overall Consistency (OC), and Vbench-Imaging Quality (IQ).
% }
}
% \vspace{-.2cm}
\resizebox{1.\linewidth}{!}{
\begin{tabular}{@{}lccc@{}}
\toprule
\textbf{RoPE Type} & \textbf{GenEval} & \textbf{Vbench-OC} & \textbf{Vbench-IQ} \\
\midrule
M-RoPE    & 0.310 & 0.122 & 0.350 \\
U-RoPE    & 0.402 & 0.165 & 0.423 \\
IL-RoPE   & 0.541 & 0.225 & 0.513 \\
VideoRoPE & 0.569 & 0.243 & 0.540 \\
HoPE      & 0.570 & 0.246 &0.545  \\
\midrule
MM-RoPE   & 0.591 & 0.249 & 0.559 \\
\bottomrule
\end{tabular}}
\vspace{-.3cm}
\label{tab:diff_RoPE_comparison}
% \end{table}
\end{wraptable}

Moreover, we conduct a quantitative comparison in~\cref{tab:diff_RoPE_comparison}, using our 1B model to perform joint training for 10k steps. 
% we compare it with two other RoPEs, \textit{i.e.}, M-RoPE~\cite{wang2024qwen2-vl} and VideoRoPE~\cite{wei2025VideoRoPE}.
Note that M-RoPE is our baseline, equal to eliminating the distributed design and the scaling design from MM-RoPE.
% \textcolor{Watermelon_Red}{
U-RoPE improves upon M-RoPE by reallocating high-frequency components for modeling local spatial details, which is useful.
IL-RoPE allocating partial channels for modeling spatial information with an interleaved design, while it only allocates partial channels for modeling texts and visual data, hurting the performance.
% }
The concurrent work, VideoRoPE, proposes several techniques for understanding tasks: low-frequency temporal allocation, diagonal layout and adjustable temporal spacing. 
We use VideoRoPE's default hyperparameters.
We observe that VideoRoPE slightly trails MM-RoPE in our task. 
We attribute this to the difference in frequency allocation and adjustable temporal spacing. 
Low-frequency temporal allocation enables more local modeling for spatial features and global modeling for temporal features, while it is important to maintain both local and global modeling for spatiotemporal features, which MM-RoPE performs better at. 
% Moreover, adjustable temporal spacing increases the resolution, similar to the scaling factors in MM-RoPE. 
% We believe that adjusting the hyperparameter in adjustable temporal spacing can match the performance of the scaling factors in MM-RoPE, which is a future work for VideoRoPE.
Moreover, adjustable temporal spacing increases the resolution for better modeling, and we believe that adjusting the hyperparameter in adjustable temporal spacing can increase its performance, which is a future work for VideoRoPE.
% \textcolor{Watermelon_Red}{
HoPE improves upon VideoRoPE with a zero-frequency strategy for temporal modeling, therefore slightly outperforming VideoRoPE.
It trails MM-RoPE because the frequency spectra that HoPE use for modeling spatiotemporal dependency are not as comprehensive as MM-RoPE.
% }

\begin{wraptable}{r}{0.5\textwidth} 
% \begin{table}[h!]
\centering
\vspace{-.3cm}
\caption{
\small
% \textcolor{Watermelon_Red}{
\textbf{Comparison of training efficiency} for text-to-image and text-to-video models.
% }
}
% \vspace{-.2cm}
\resizebox{1.\linewidth}{!}{
\begin{tabular}{@{}lccc@{}}
\toprule
\textbf{Method} & \textbf{Params} & \textbf{\shortstack{Time \\ (8$\times$A100 GPU-Days)}} & \textbf{GenEval} \\
\midrule
SD-1.5          & 0.9B + 0.1B & 781.2  & 0.43 \\
SD-2.1          & 0.9B + 0.3B & 1041.6 & 0.50 \\
Dall-E 2        & 4.2B + 1.0B & 5208.3 & 0.52 \\
PixArt-$\alpha$ & 0.6B + 4.7B & 94.1   & 0.48 \\
Lumina-mGPT     & 7B          & 4488.8 & 0.56 \\
\midrule
\method & 1.5B & \textbf{25.9} & \textbf{0.725} \\
\bottomrule
\toprule
\textbf{Method} & \textbf{Params} & \textbf{\shortstack{Time \\ (8$\times$A100 GPU-Days)}} & \textbf{VBench} \\
\midrule
Open-Sora Plan v1.2 & 2.7B + 13B & > 592 & 75.98 \\
\midrule
\method & 1.5B & \textbf{85.4} & \textbf{78.17} \\
\bottomrule
\end{tabular}}
\vspace{-.2cm}
\label{tab:training_resource}
% \end{table}
\end{wraptable}

% \textcolor{Watermelon_Red}{
\noindent\textbf{Comparison of training resources.}
We compiled a detailed comparison of training costs against popular text-to-image and text-to-video models. 
To ensure a fair comparison, we normalized all training times to 8xA100 GPU-Days based on their respective FP16 Tensor Core performance. 
The results are placed in~\cref{tab:training_resource}.
We can observe from the table that, compared with text-to-image diffusion models or larger-scale AR models, Lumos-1 is more compute-efficient in terms of training and obtains better performance on GenEval.
Compared with Open-Sora Plan v1.2, Lumos-1 achieves similar performance, while reducing the training cost significantly.
Unlike diffusion models, which rely on a powerful, external text encoder, Lumos-1 learns language understanding from scratch.
Its high efficiency, even with this added learning burden, highlights the effectiveness of our discrete tokenization and MM-RoPE alignment strategy.
% }

\subsection{More Visualizations} \label{app:more_visualizations}
\begin{figure}[t]
\centering
\includegraphics[width=1.0\linewidth]{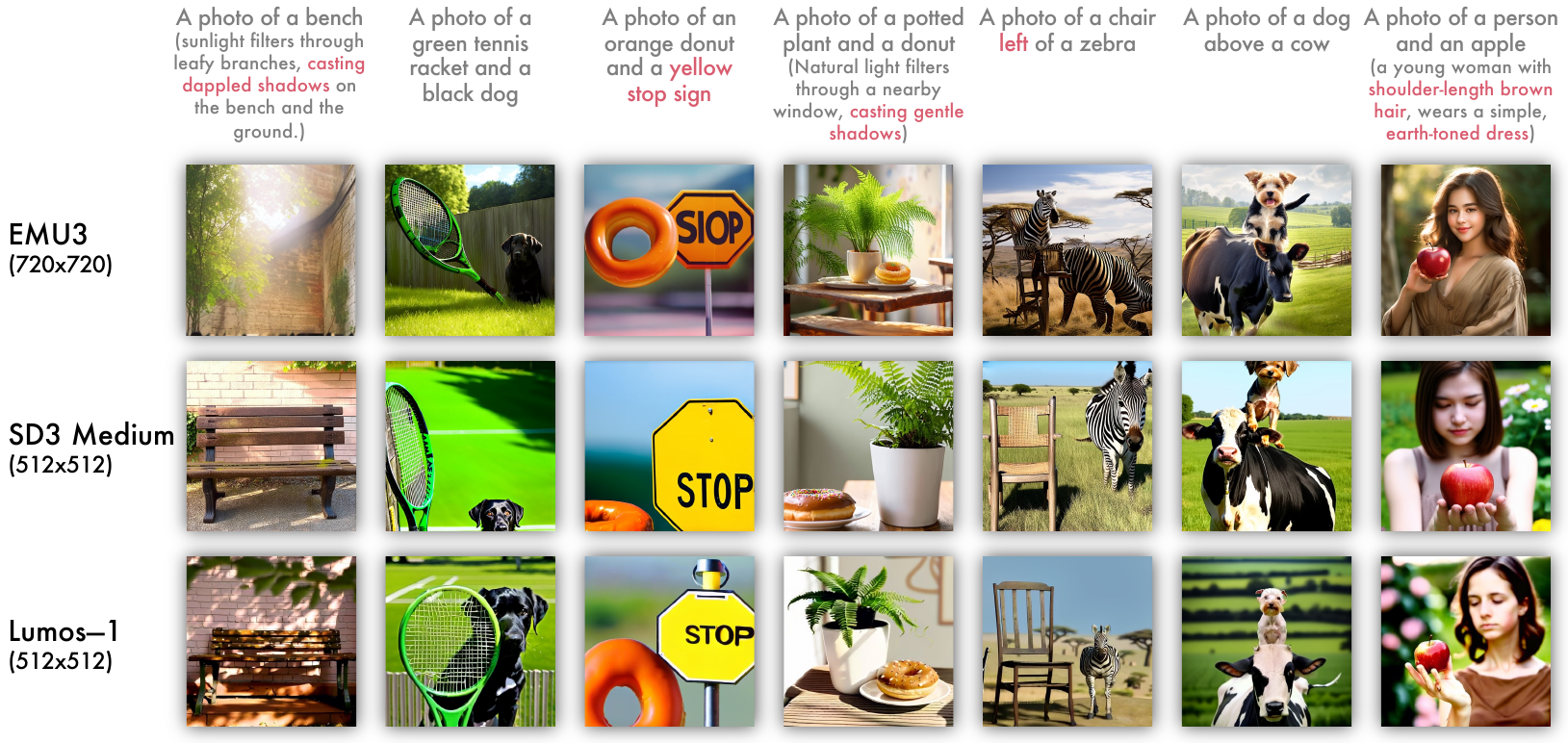}
\vspace{-.3cm}
\caption{
\small
\textbf{Visual comparison} of \method with EMU3 and Stabel Diffusion 3 (Medium) on the text-to-image task.
We place the simple version of the prompts above the images for reference.
Fragments of the detailed prompt are placed in parentheses, with the critical attributes highlighted in \textcolor{Watermelon_Red}{\textbf{red}}.
All images are generated using the detailed prompts for all models.
}
% \vspace{-.33cm}
\label{fig:t2i_comparison}
\end{figure}

\textbf{Qualitative visual comparison on text-to-image generation.}
We compare \method with popular text-to-image generation methods in~\cref{fig:t2i_comparison}.
All models are configured to their default inference settings and use the same version of detailed prompts to ensure quality generation.
We can observe that:
\textbf{1)} Although EMU3 is significantly larger than \method, \method generates images that reproduces fine prompt details more faithfully, such as the dappled shadows on the bench (example 1) and the correct “STOP” text and the yellow color on the road sign (example 3).
EMU-3 sometimes drops or warps these details and exhibits occasional anatomical distortions, such as the distorted zebra (example 5) and malformed fingers on the woman holding an apple (example 7).
\textbf{2)} Stable Diffusion 3 offers slightly crisper textures, which is expected from its continuous tokenizer, but it frequently ignores descriptive constraints, such as 
the dappled shadows (example 1) and gentle shadows (example 4), and the incorrect dress color (example 7). By contrast, \method meets these constraints while maintaining a convincing overall composition.
These examples illustrate that \method delivers strong vision–language alignment compared with both diffusion and autoregressive baselines, while remaining competitive in visual quality.
% Although EMU3 is significantly larger than \method, \method excels in better vision-language alignment by understanding details in descriptive prompts, \textit{e.g.}, bench and dappled shadows.
% \method even emerges the capability of some simple text rendering, \textit{e.g.}, a stop sign.
% EMU3 is a frontier unified model, but still suffer from deformation such as the zebra and fingers of the person who holds an apple.
% Compared with Stabel Diffusion 3 (Medium), \method can trail in fine-grained visual quality due to its discrete tokenizers.
% However, Stabel Diffusion 3 fails to render dappled shadows (example 1), gentle shadows (example 4) and earth-toned dress (example 7), demonstrating the effectiveness of \method.

\begin{figure}[t]
\centering
\includegraphics[width=1.0\linewidth]{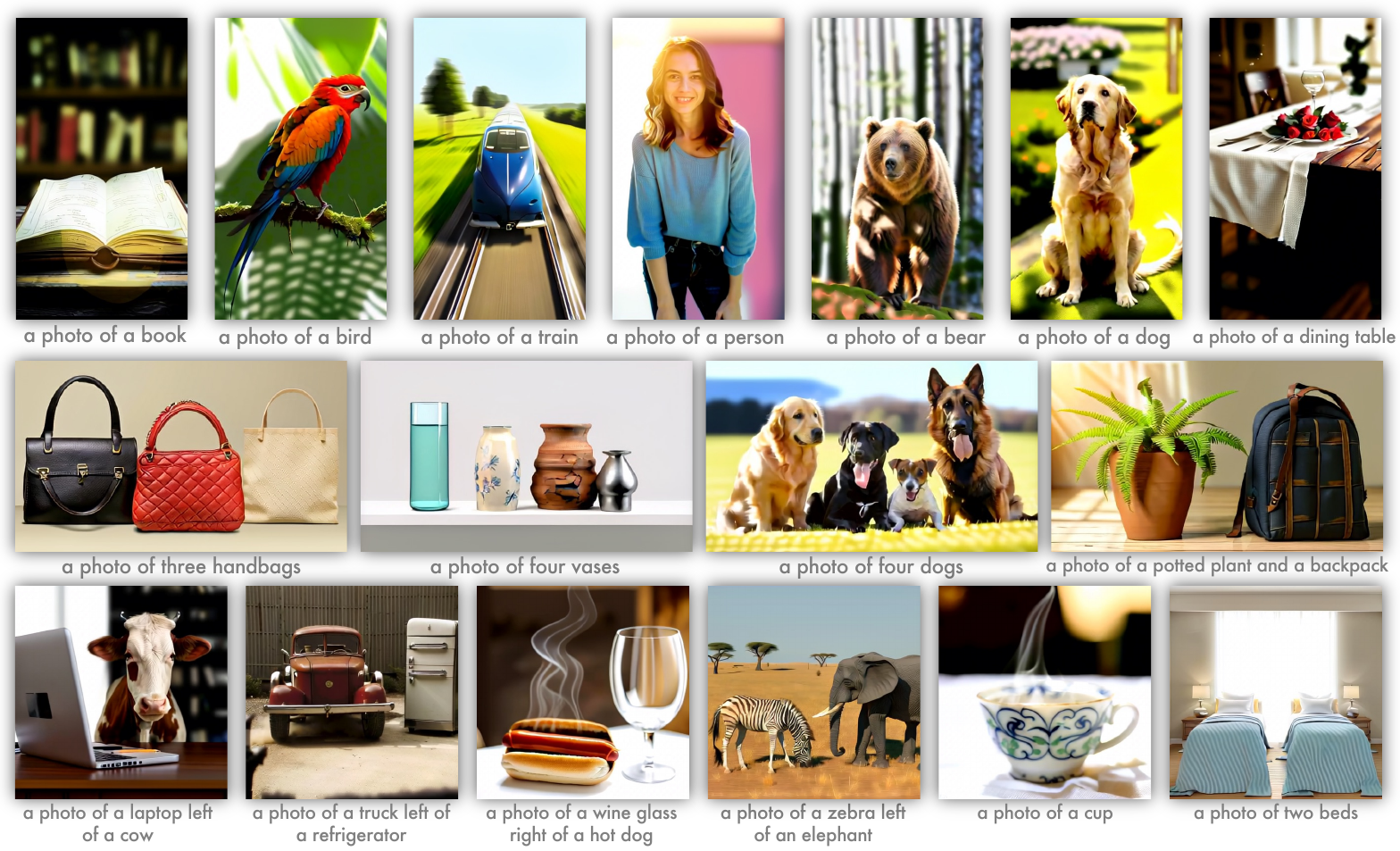}
\vspace{-.5cm}
\caption{
\small
\textbf{More text-to-image results} generated by \method.
All images are generated using \textit{detailed prompts}, but we only place short prompts here due to space limits.
% The following examples are 
The resolutions for three rows of images are $384\times 672$, $672\times 384$ and $512\times 512$.
}
% \vspace{-.33cm}
\label{fig:t2i_more_results}
\end{figure}

\textbf{More text-to-image visualizations.}
\cref{fig:t2i_more_results} presents more results generated by \method using descriptive captions.
Due to the unified codebook, we can train \method using the original resolution of the visual data, enabling generation with different aspect ratios such as $4:7$, $7:4$, and $1:1$, although we do not have a significant amount of data with aspect ratio $4:7$. 
% although there is not significantly large-scale data of resolution $7:4$.
% Even though \method is trained from scratch, the model processes the capability of generating images containing complex scenes that is aligned faithfully with long and detailed prompts.
% This can be exemplified by the second row of images, where the detailed prompts clearly describe the attributes of all objects, which \method all successfully generate.
Even though \method is trained from scratch, it transforms lengthy descriptions into complex scenes that are aligned faithfully with the prompts.
This can be exemplified by the second row of images, where these images clearly reflect the prompts:
\begin{itemize}
    \item ``\textit{a photo of three handbags}'' contains the description: \textit{To the left, a classic black leather handbag with a structured design and gold-tone hardware catches the eye. In the center, a vibrant red quilted handbag with a chain strap adds a pop of color and a touch of luxury. On the right, a casual beige canvas tote bag with simple, clean lines offers a more relaxed aesthetic.}
    
    \item ``\textit{a photo of four vases}'' contains the description: \textit{The first vase on the left is a tall, cylindrical glass vase with a slight teal tint, showcasing clear transparency and a modern design. Next to it is a vintage ceramic vase with an off-white base adorned with hand-painted blue floral patterns, adding a touch of traditional elegance. The third vase is a short, squat terracotta pot with a rustic, earthy texture and visible firing marks, evoking a natural, bohemian feel. Finally, a sleek, metallic silver vase with a geometric shape completes the set, providing a contemporary contrast.}
    
    \item ``\textit{a photo of four dogs}'' contains the description: \textit{To the left is a golden retriever, its fur glowing warmly in the sunlight, with a gentle, friendly gaze. Next to it is a sleek black Labrador, its coat shining with health, head tilted slightly as if listening intently. In the center, a small, energetic Jack Russell terrier bounces with excitement, its white and brown fur clearly defined. On the right, a large German Shepherd sits calmly, its regal posture and dark, expressive eyes adding a dignified presence to the group.}
\end{itemize}
These qualitative results underscore the strong vision–language alignment achieved by \method.

\begin{figure}[t]
\centering
\includegraphics[width=1.0\linewidth]{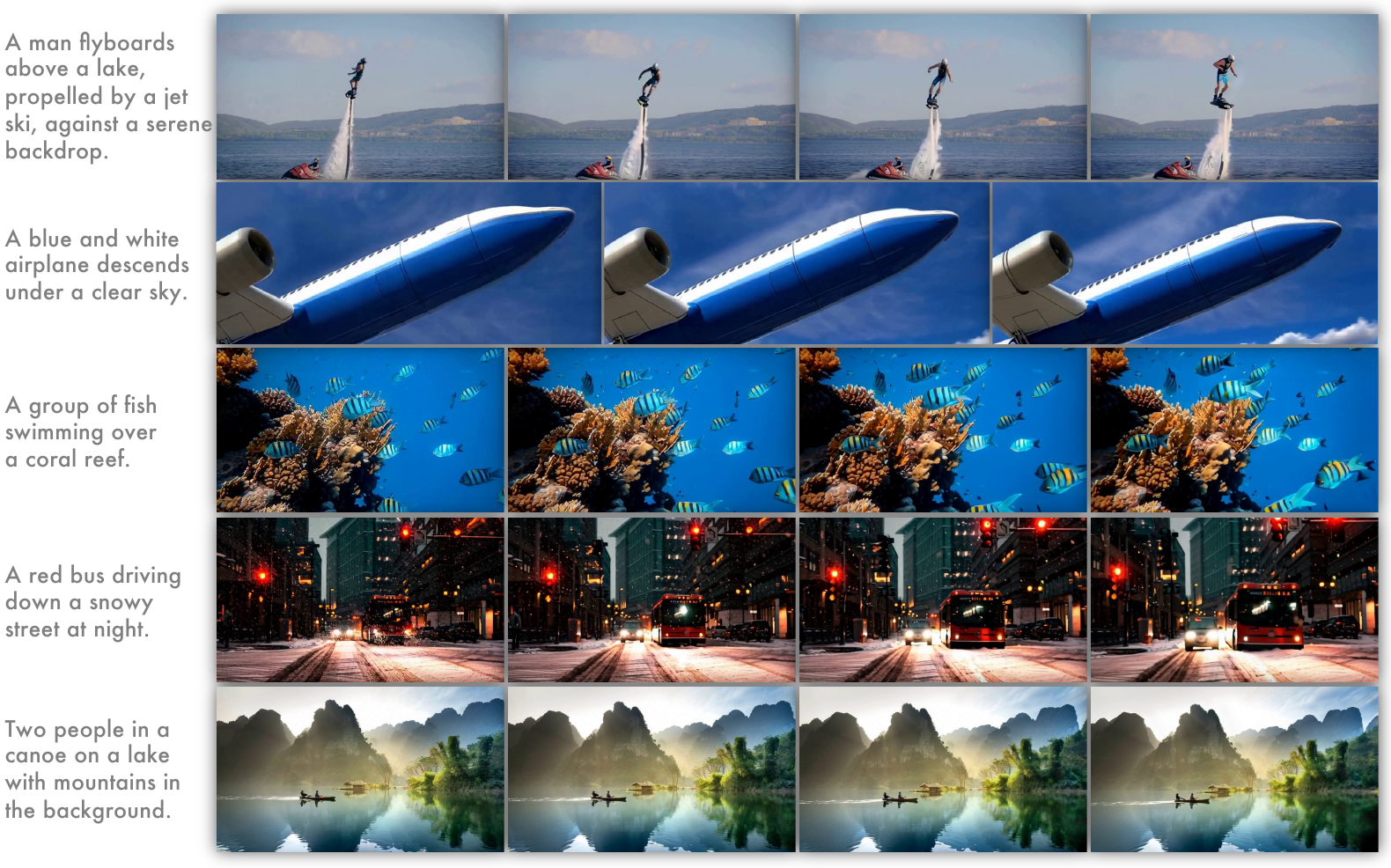}
\vspace{-.5cm}
\caption{
\small
\textbf{More image-to-video results} generated by \method.
All images are generated using \textit{detailed prompts}, but we only place short prompts here due to space limits.
The results encompass videos of resolution $672\times 384$ and $768\times 320$.
}
% \vspace{-.33cm}
\label{fig:i2v_more_results}
\end{figure}

\begin{figure}[t]
\centering
\includegraphics[width=1.0\linewidth]{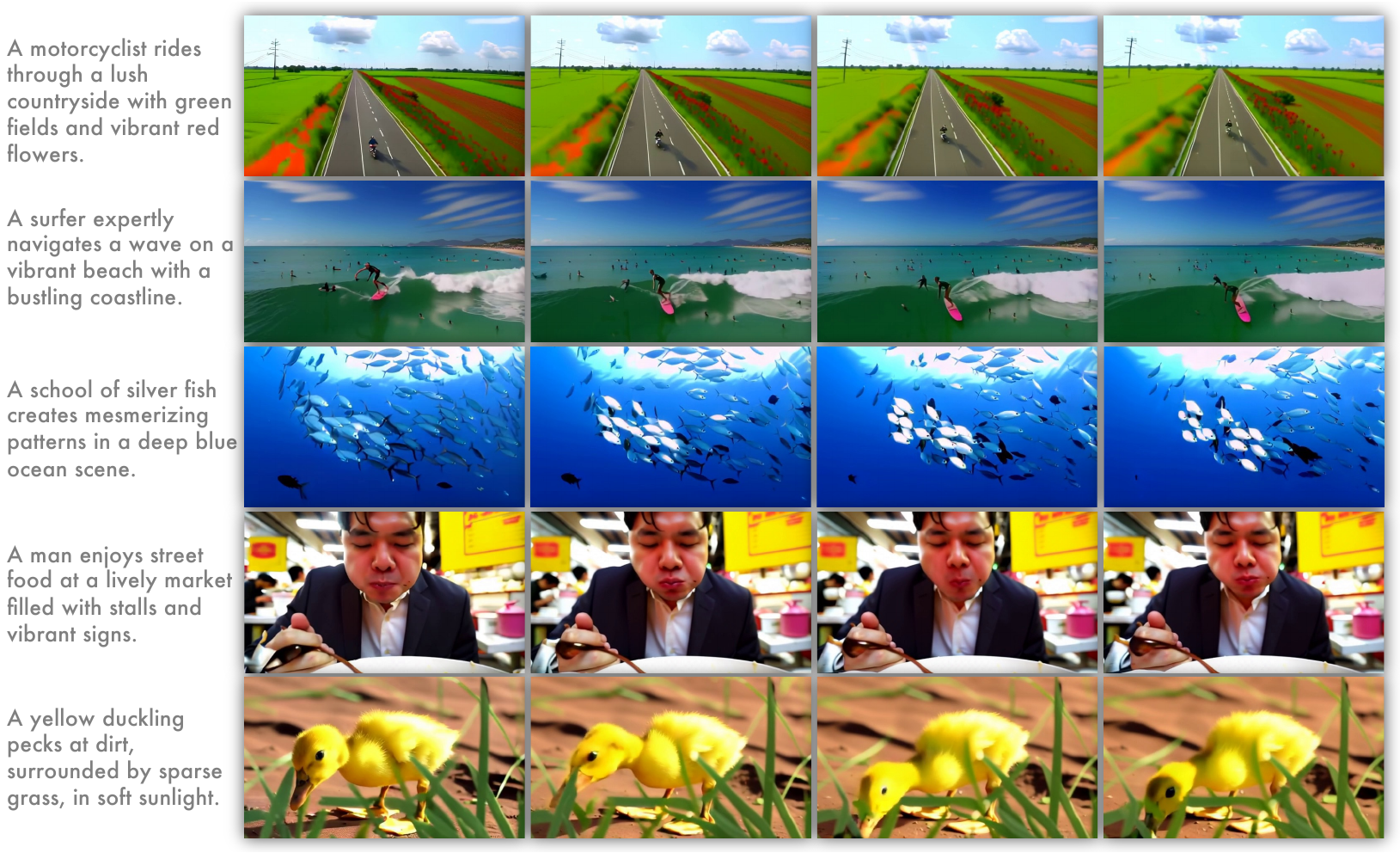}
\vspace{-.5cm}
\caption{
\small
\textbf{More text-to-video results} generated by \method.
All images are generated using \textit{detailed prompts}, but we only place short prompts here due to space limits.
The resolution is $672\times 384$.
}
% \vspace{-.33cm}
\label{fig:t2v_more_results}
\end{figure}

\textbf{More image-to-video and text-to-video visualizations.}
\cref{fig:i2v_more_results} presents more image-to-video results and \cref{fig:t2v_more_results} presents more text-to-video results generated by \method.
The prompts span a wide semantic and dynamical range, containing 
\textbf{1)} close-up animal/human scenes (example 4, 5 in \cref{fig:t2v_more_results}), 
\textbf{2)} fast human and vehicle motion (example 1, 4 in \cref{fig:i2v_more_results} and example 1, 2 in \cref{fig:t2v_more_results}),
\textbf{3)} fluid dynamics (example 1, 5 in \cref{fig:i2v_more_results} and example 2 in \cref{fig:t2v_more_results}), and
\textbf{4)} complex multi-object collective motion (example 3 in \cref{fig:i2v_more_results} and example 3 in \cref{fig:t2v_more_results}).
The results demonstrate that \method can
\textbf{1)} handle videos of different aspect ratios and 
\textbf{2)} generate single-subject and multi-subject motion that is temporally coherent and physically plausible.

% \section{Architecture pipeline: A Figure}

\section{Discussions} \label{app:discussions}

\subsection{A Road Map to Unified Models for Understanding and Generation}\label{app:road_map_to_unified_models}

% \textcolor{Watermelon_Red}{
Building upon our current work, a complete unified model can be achieved through a principled, staged approach:
% }

\begin{itemize}
    \item \textbf{Stage 1: Foundational Model Selection and Codebook Unification.} 
The starting point is a base model with strong language capabilities. There are two primary paths: 
    \textbf{1)} \textit{Path A (Pragmatic Start)}: Initialize with a pre-trained VLM that already possesses strong visual and textual understanding. The main task is then to integrate our generative capabilities by extending its vocabulary with a discrete visual codebook and fine-tuning with our proposed AR-DF and MM-RoPE on generation tasks.
    \textbf{2)} \textit{Path B (Purist, End-to-End)}: Start with a pure LLM. This requires developing a dual-purpose discrete visual tokenizer—one that produces tokens rich enough for both high-fidelity reconstruction (for generation) and high-level semantics (for understanding). This is a challenging but crucial research direction for true end-to-end unification.

    \item \textbf{Stage 2: Multi-Task Data Curation.}
A successful unified model requires a diverse, high-quality dataset covering all target modalities and tasks. This involves curating a mixture of: 
    \textbf{1)} \textit{Language-only Data}: To maintain and enhance the model's core reasoning and prevent catastrophic forgetting. 
    \textbf{2)} \textit{Image/Video-Text Pairs}: For training both understanding (e.g., captioning, VQA) and generation (T2I, T2V) tasks. \textbf{3)} \textit{Interleaved Multi-modal Documents}: To teach the model to seamlessly process and generate complex sequences of text and visuals.

    \item \textbf{Stage 3: Joint Multi-Task and Staged Training.}
The final step is to train the model to be a versatile multi-task model. This should be done in stages:
    \textbf{1)} \textit{Understanding Foundation (if using Path B)}: First, train the LLM on understanding tasks (e.g., VQA, captioning) to establish a strong vision-language alignment baseline.
    \textbf{2)} \textit{Joint Generation and Understanding Training}: In the main stage, train the model jointly on a mix of understanding and generation objectives. Our proposed AR-DF and MM-RoPE would serve as the core engine for all generative and understanding tasks. By sharing the same transformer backbone, the model is forced to develop a rich, shared representation space where concepts learned from understanding tasks can positively transfer to generation, and vice versa.
\end{itemize}

\subsection{Limitations and Future Work Discussions}\label{app:limitations_future_work}

We recognize that \method, as an initial endeavor in this area, comes with its limitations.
% Compared with models trained on billions of images and videos~\cite{wan2.1,flux2024}, we kindly acknowledge that they can have better generalization compared with \method.
% Because of this, \method might trail in certain cases where the videos feature sophisticated human action and complex scenes.
Most prominently, its training corpus (60 million images and 10 million videos) is modest compared with datasets used by recent foundation models \cite{wan2.1,flux2024}, which usually contain billions of samples.
Consequently, \method can under-generalize in scenarios that require fine-grained human actions or highly intricate scene dynamics.
% Considering this, one direct future work is to continue scaling the model and data scale.
% Moreover, we also consider the incorporation of dominant vision-language models, so that the obtained model could enjoy the benefits of the world knowledge.
Considering this, our immediate research plan therefore includes scaling along three axes:
\textbf{1)} \textit{Data volume and diversity} – expanding both image and video coverage to narrow the generalisation gap.
\textbf{2)} \textit{Model capacity} – training larger backbones while retaining the MM-RoPE and AR-DF designs.
\textbf{3)} \textit{Multimodal knowledge infusion} – initializing with strong vision–language models or co-training with visual understanding tasks so the generator can better ground its outputs in world knowledge.

% Moreover, since the most popular generative models (\textit{i.e.}, diffusion models) usually utilize .
% data and model scaling.
% editing.

\subsection{Potential Societal Impacts and Safeguards}\label{app:societal_impact_safeguards}

As the pioneering efforts in autoregressive visual generation, \method can be a significant step towards a large-scale unified model for general visual understanding and generation.
Through substantially large-scale training, the obtained model could stimulate intelligence that exceeds existing models trained on unimodal data.
The obtained model could serve as a foundation for various downstream applications, thereby reducing the carbon and economic cost that society spends on training various specialized models.

% However, as a general visual generative model, \method could be put into use in harmful applications or generate visually unappealing content.
At the same time, unrestricted deployment carries non-trivial risks: the model might be used to fabricate deceptive media, produce disallowed or disturbing visuals, or amplify harmful biases present in the training data.
% Therefore, necessary actions must be taken before its deployment to secure safe usage.
% First of all, necessary alignment techniques should be applied to ensure that the output aligns with human aesthetic and moral preferences.
% Moreover, filtering techniques should be designed and applied to the input stage for prompt filtering and to the output stage for quality assessment and XX [replace XX with proper words].
% \method is a research-oriented work, and its deployment to any circumstance beyond research should be approached with thorough oversight and evaluation to ensure responsible and ethical use.
To mitigate these concerns, we recommend the following safeguards before any real-world release:
\textbf{1)} \textit{Alignment tuning} - apply preference- or instruction-tuning so that outputs respect human aesthetic and moral preferences~\cite{yuan2024instructvideo}.
\textbf{2)} \textit{Dual-stage filtering} – i) prompt filters that refuse generation requests involving illicit content, and ii) post-generation classifiers that block or watermark unsafe outputs.
\textbf{3)} \textit{Red-team evaluation} – continuous adversarial testing to uncover failure modes in new domains or under distribution shift.
\method is a research-oriented work, and any broader deployment should proceed only with thorough oversight and evaluation to ensure responsible and ethical use.

\section{Disclosure of LLM Usage} \label{app:LLM_usage}
We utilized a LLM to assist in the preparation of this manuscript. 
The primary role of the LLM was to improve the language, clarity, and readability of our writing. 
% Specific tasks performed by the LLM include proofreading for grammatical errors, correcting spelling, refining sentence structure, and ensuring stylistic consistency.
All intellectual contributions, including the research ideation, experimental design, data analysis, and the core arguments presented in this paper, are entirely our own. 
The LLM was used solely as a writing-enhancement tool and did not contribute to the scientific aspects of the work. 
% The model used was OpenAI's GPT-4.

% \input{LlamaVGen/content/tables/Appendix_geneval_for_exp_record}

% \input{content/01-introduction}
% \input{content/02-related-work}
% \input{content/03-methodology}
% \input{content/04-experiments}
% \input{content/05-conclusion}

% \clearpage
% \input{content/07-appendix}

% \bibliographystyle{plain}
% \bibliography{reference}

\end{document}